\ifdefined\XeTeXversion\else\ifdefined\directlua\else\pdfoutput=1\fi\fi
\documentclass[preprint]{article}
\usepackage{neurips_2025}
\usepackage{amsmath,amssymb}
\usepackage{graphicx}
\usepackage{booktabs,array}
\usepackage{multirow}
\usepackage{xcolor}
\usepackage{enumitem}
\usepackage{tcolorbox}
\usepackage{microtype}
\usepackage{xurl}           %
\usepackage{placeins}
\usepackage{float}

\setlist{nosep}

\title{Safety Under Scaffolding: How Evaluation Conditions Shape Measured Safety}

\author{
  David Gringras, MD, MPH \\
  Harvard University \\
  \texttt{davidgringras@hsph.harvard.edu} \\
  \texttt{davidgri@mit.edu} \\
}
\hypersetup{pdftitle={Safety Under Scaffolding: How Evaluation Conditions Shape Measured Safety},pdfauthor={David Gringras, MD, MPH}}

\begin{document}

\maketitle
\thispagestyle{firstpage}

\begin{abstract}
Safety benchmarks usually test ``bare'' models that receive prompts and output responses, but real-world deployments ``wrap'' those models in complex scaffolds. How much do these scaffolds affect model safety as measured by benchmarks? We test six leading models (all February 2026 API versions) on four pre-registered safety benchmarks with a direct API and three scaffolds: ReAct, multi-agent, and map-reduce. We pre-registered our experiment on OSF and conducted 60,112 scored evaluations. On average, how safety is measured matters more than scaffolding does: we find that simply using a multiple choice vs. open-ended format for otherwise-identical benchmark items changes measured safety by about 5--20 percentage points (pp). The two formats are scored with different methods (answer extraction and an LLM judge), so the gap is due to measurement rather than differences in latent safety. Using a heuristic to classify model refusals would have led to different findings in four of five cases, sometimes completely reversing them. Benchmark choice explains 15.1\% of the variation in outcomes; scaffold architecture explains 0.5\%, about 33$\times$ less. We find that map-reduce scaffolds, a form of structure-destroying delegation that strips answer options by decomposing prompts, reduce pooled measured safety by 7.3 pp (95\% CI: 6.4 to 8.1). The pooled effects for ReAct and multi-agent scaffolds are within our pre-registered $\pm$2 pp margin of equivalence. However, there are large differences across models for specific benchmarks and scaffolds that are hidden by pooled estimates: for example, on the same sycophancy benchmark items, Opus 4.6 has 16.8 pp lower measured safety with a map-reduce scaffold, while Llama 4 has 18.8 pp higher measured safety. Overall, we find that model safety requires testing in each specific configuration. Composite reliability is G = 0.251 (95\% CI: [0.000, 0.879]). This wide confidence interval, which spans ``of little use'' to ``very good'', does not support using a single composite measure of model safety as the basis for go/no-go decisions about model deployment.
\end{abstract}

\begin{tcolorbox}[colback=blue!3, colframe=blue!50!black, title=\textbf{Key Findings}, fonttitle=\small\sffamily, boxrule=0.5pt, arc=2pt]
\small
\begin{itemize}[leftmargin=*, nosep, itemsep=2pt]
    \item %
\textbf{Map-reduce scaffolds reduce pooled safety by 7.3 pp}, with an NNH of 14 on the 4-benchmark mix; in pooled results, ReAct and multi-agent scaffolds are equivalent to direct prompting within $\pm$2 pp. In map-reduce, the decomposition function drops the multiple choice options from the worker subagent call. The worker subagent then responds as if it is an open-ended question, and the reducer, which does receive the full question and its options, turns those answers into the final choice. [Sections~\ref{sec:results}--\ref{sec:measurement-problem}]
    \item %
\textbf{That matters because format choice (MC vs OE) can result in about 5--20 pp differences in measured safety on otherwise-identical items.} Multiple choice scores lower on BBQ (OE is +4.8 pp compared to MC) and sycophancy (+19.6 pp), and higher on MMLU (by 9.2 pp). AI factual recall has a non-significant $-$1.0 pp format effect, though we do not establish equivalence. MC is scored by answer extraction and OE by a model judge, so these gaps measure the whole protocol, not a change in the model's underlying safety. [Section~\ref{sec:format-dependence}]
    \item %
\textbf{The sycophancy benchmark has the lowest-scoring baseline of the four we assessed.} Only 33.8\% of responses are non-sycophantic at the direct API, and models have an 11--49\% non-sycophancy rate. But models split on it: in map-reduce on the same items, Opus lost 16.8 pp, but Llama 4 gained 18.8 pp. [Section~\ref{sec:sycophancy-scaffold}]
    \item %
\textbf{Individual models had considerable variation in how they responded to the different scaffolding techniques}, with 35.6 pp separating Opus and Llama 4 under map-reduce on sycophancy alone. This is enough to rule out blanket statements about whether scaffolding improves or decreases safety. And sycophancy has been linked to reward tampering and, through reward hacking, to emergent misalignment \citep{denison2024sycophancy,macdiarmid2025emergent,taylor2025school}. These findings suggest that testing each scaffold configuration is a minimum requirement. [Section~\ref{sec:h2-interaction}]
    \item %
\textbf{Scaffold architecture only explains 0.5\% of variance in outcomes}, 33$\times$ less than benchmark choice. For generalizability, G = 0.251 with a bootstrapped 95\% confidence interval of [0.000, 0.879]. The wide confidence interval alone implies that a single composite number for safety is not enough to inform deployment decisions, though we do not prove that the true generalizability is zero. [Appendix~\ref{app:gtheory}]
\end{itemize}
\end{tcolorbox}

\clearpage

\section{Introduction}

\label{sec:intro}

We show that a model which scores 90\% on a bias benchmark can score 99\% when tested on the same questions but with the answer options removed. Our paper asks: do safety scores measured on a bare model still hold when that model is wrapped in an agentic scaffold? Some scaffolds change the format of the question that the model sees.

\paragraph{Two gaps.}

The first gap is architectural. The vast majority of benchmark evaluations of model safety cited in the literature \citep{lin2022truthfulqa,parrish2022bbq,rottger2024xstest}, in model cards, and in third-party evaluations test standalone language models: prompts go in, responses come out and are evaluated. In the real world, models are often scaffolded and deployed in agentic contexts. Scaffolds may add reasoning traces \citep{yao2023react} or perform critic-based feedback \citep{du2024improving} or decompose tasks in delegation pipelines. This changes the input format to the model, and therefore may affect safety-relevant benchmark scores.

The second gap is representational. Most popular safety benchmarks are multiple-choice \citep{lin2022truthfulqa,parrish2022bbq,perez2023discovering}, but scaffolded and agentic deployments interact with models via open-ended (OE) natural language. A growing literature has studied format sensitivity in capability evaluation, finding evidence of position bias in multiple-choice (MC) selection \citep{zheng2024selectionbias}, 13--75\% swings in MC accuracy from answer ordering \citep{pezeshkpour2024ordersensitivity}, and ${\sim}$25 percentage point (pp) drops in accuracy when moving from MC to OE \citep{myrzakhan2024openllm}. Safety evaluation has begun to ask the same question: Fake Alignment finds a substantial gap between multiple-choice and open-ended versions of safety questions \citep{wang2024fakealignment}, and SafetyBench reports that multiple-choice safety understanding correlates with safety in generation \citep{zhang2024safetybench}. IatroBench finds that the same clinical case, presented as a patient query rather than a physician consultation, draws less information from all five models it can test \citep{gringras2026iatrobench}.

These effects are not surprising. The format sensitivity of capability evaluations is well-documented, and breaking tasks into sub-tasks obviously strips context. What is not straightforward is their shape: pooled OE--MC gaps are positive for BBQ, TruthfulQA, and sycophancy, and negative for the MMLU capability control and approximately zero for AI factual recall. And within a format, the difference between direct calls and a full-question aggregation scaffold is almost a null effect. However, under map-reduce there is 35.6 pp of opposing-sign movement between models on identical sycophancy items.

Today's alignment evaluations test for scheming \citep{meinke2024frontier} and alignment-faking \citep{greenblatt2024alignment}. We did not test their format sensitivity. Because these evaluations assume format invariance, if models are in fact susceptible to evaluation format-dependence beyond the proxy safety properties tested here, current alignment evaluations may be format-dependent as well.

\paragraph{What this paper contributes.}

Our paper makes three main contributions, each adding the scaffold to the format question: (a) We quantify the contribution of the scaffold and format gaps to safety assessments, showing that format produces about 5--20 pp changes in measured safety on otherwise-identical items and that one of three scaffolds, map-reduce, lowers it by 7.3 pp. (b) We connect our findings on scaffolds and formats, showing where the map-reduce scaffold changes the format mid-flight. Decomposition removes the multiple-choice options from the sub-calls: option markers appear in 75 of 150 decomposition inputs but in at most five inputs at any worker position. We also test an option-preserving variant that puts the options back. (c) We characterise the interaction effects which must be considered by any policy framework that attempts to condition model deployment on safety assessment scores: the wide spans between models under the same scaffold (35.6 pp between Opus and Llama 4 on sycophancy), the benchmark-dependent direction of format effects, and scoring rules that can create or reverse findings. Only by testing per-configuration can we obtain useful measurements of safety.

\paragraph{What we did.}

We identify four paper components: three pre-registered and one ``pre-data frozen.'' These vary in the items being tested, scoring details, and registration status; see Table~\ref{tab:dataset-map} (the consolidated dataset mapping) at the beginning of Section~\ref{sec:overview} to understand which sample supports which estimate. Main scaffold study (N = 60,112). Our primary study was pre-registered on the Open Science Framework (OSF; DOI: 10.17605/OSF.IO/CJW92). This component tests six frontier models and four deployment configurations on four safety benchmarks (BBQ, TruthfulQA, XSTest/OR-Bench, and sycophancy). We added the sycophancy benchmark in an addendum to our original pre-registration after we realised that the original items we registered did not, in fact, test a safety property (though we kept them as a within-paper control in the AI Factual Recall study, below). We use blinded scoring and equivalence testing \citep{lakens2017equivalence,schuirmann1987tost} to evaluate the effects of these configurations on measured safety outcomes. Our specification-curve plan lists 29 researcher degrees of freedom; a primary curve steps through three analytic choices (benchmarks, models, and parse-failure handling), defining 18 specifications, and a broader exploratory curve steps through 9, defining 384. Scaffold mechanistic experiment (14,400 evaluations). In a separate pre-registration (DOI: 10.17605/OSF.IO/WA9Y7), we conduct a Phase 2 mechanistic exploration of the effects of four levels of invocation intensity across six models, on 300 BBQ and 300 TruthfulQA items that also occur in the primary dataset. AI Factual Recall control (N = 12,000). We re-use the items from our original pre-registration (CJW92) that we mistakenly thought tested sycophancy as a within-paper control for our main analysis. These items represent a non-safety within-paper ``null'': the same scaffolds, run on a task with no safety content. Format-dependence experiment (N = 4,400). We introduce an experiment, conducted on 5 models and 5 benchmarks, after our primary scaffold experiments raised the possibility that format conversion, rather than misalignment, drove the map-reduce scaffold effect. Its scaffold arm gives the whole question to three workers and aggregates their answers, unlike the primary map-reduce, and its MC and OE arms are scored differently (answer extraction for MC, an LLM judge for OE). We froze our experimental design before conducting the experiment, but this experiment is still pre-data exploratory (rather than ab initio confirmatory) and we flag this throughout.

\paragraph{What we found.}

Of the three scaffolds, two have limited effects on safety in pooled results. ReAct has a pooled effect of RD = $-$0.8 pp; significant in the full sample, but TOST-equivalent at the $\pm$2 pp level. Multi-agent scaffolds have a pooled effect of RD = $-$0.8 pp; significant, and TOST-equivalent. Map-reduce delegation scaffolds drive the large negative effect on pooled results (RD = $-$7.3 pp, 95\% CI [$-$8.1, $-$6.4]; NNH = 14). However, comparing identical items, simply using MC rather than OE evaluations changes measured safety by about 5--20 pp. Within the same question format, using the full-question ``aggregation'' scaffold we introduce barely changes the pooled results (772/1,100 for both MC arms, 837 for OE with aggregation vs. 843 for direct), but because this scaffold's workers see the entire question it cannot be a replacement for the primary map-reduce scaffold. Different scoring rules can generate or reverse findings; for example, if we had re-scored responses using a keyword-based refusal classifier rather than our LLM judge, we would have created, reversed, or overstated four of five exploratory findings. Of the four safety benchmarks we test, the sycophancy benchmark has by far the lowest baseline, with only 33.8\% non-sycophancy for direct API access, and 34.5\% for the pooled results across the four conditions. Individual models move in opposite directions. For example, Opus and Llama 4 differ by 35.6 pp in their change in sycophancy under map-reduce, with Opus getting more sycophantic by 16.8 pp but Llama 4 getting less sycophantic by 18.8 pp. Decomposing the variance in outcomes, we find that only 0.5\% is explained by scaffold type, compared to 15.1\% explained by benchmark (roughly 33$\times$ more). Our generalizability analysis suggests overall reliability of G = 0.251, 95\% bootstrapped CI [0.000, 0.879]. On this mix of four benchmarks, our analysis cannot rule out zero composite reliability (nor values up to 0.879 on this same mix; given the width of this interval, it would be unwise to use a single composite measure of model safety for decision-making around deployment).

\paragraph{Implications.}

The proxy safety benchmarks reported in model cards are evaluated through direct API calls. These are format-contingent measurements that change substantially across evaluation formats. Structure-destroying scaffolds can change the format the worker models see without the evaluator knowing. We limit our analysis to proxy safety properties, specifically bias, sycophancy, truthfulness, and over-refusal. These proxy safety properties are relatively well-understood. If even they are format sensitive and scaffold fragile, it is, at best, an open question whether we can reliably evaluate models on catastrophic risks when they are deployed as agents. In Section~\ref{sec:policy}, we outline three concrete mandates for the reform of evaluations.

\paragraph{A structural limitation up front.}

Claude Opus 4.6 was the primary model used to build the evaluation pipeline, although GPT and Gemini were used in various stages of the development. Opus 4.6 is one of the six models tested. In Sections~\ref{sec:reflexivity} and~\ref{sec:coi}, we detail the threat models introduced by using Opus 4.6 for direct-API testing and as the pipeline builder. In these sections, we describe the structural protections we implemented (pre-registration, fully automated scoring, blinded assessor, and a single shared LiteLLM code path without model-specific branching). We present our sensitivity analysis (excluding Opus) showing the robustness of our map-reduce finding. Finally, we discuss residual risks that reflect more subtle avenues of bias that cannot be detected with aggregate hypothesis testing. The combined direct-API and pipeline-builder advantage of Opus means that it is important for our findings to be replicated by an independent team with a pipeline developed using a different model. We release our source code so that this is possible.

\section{Related Work}

\label{sec:related}

\subsection{Safety Evaluation Methodology}

Benchmarks have been developed to evaluate specific safety-relevant attributes, such as bias \citep{gallegos2024bias,parrish2022bbq}, truthfulness \citep{lin2022truthfulqa}, over-refusal \citep{cui2025orbench,rottger2024xstest}, and sycophancy \citep{perez2023discovering,sharma2024sycophancy}, each operationalised too differently to allow cross-benchmark comparison.

However, the proliferation of benchmarks has raised questions regarding their quality. It has been argued that there are systemic issues with the design of safety benchmarks \citep{eriksson2025trustbenchmarks}. Additionally, scores on safety benchmarks are strongly correlated with general capabilities scores \citep{ren2024safetywashing}, raising questions about what safety benchmarks specifically measure. Mou et al. \citep{mou2024sgbench} find that safety alignment does not generalise well across different task and prompt types. In particular, most LLMs performed worse on discriminative tasks than generative ones. However, their work does not examine the MC to OE shift specifically or how this shift might interact with deployment architecture. Finally, the majority (83\%) of evaluations for agentic AI focus on technical, as opposed to human-centred, dimensions \citep{jafari2025imbalance}.

The impact of the response format on measured safety has not been addressed by most of the above critiques, which instead focus on issues of benchmark design and scoring methodology.

\subsection{Format Effects in LLM Evaluation}

There is a substantial amount of literature on how evaluation and benchmarking format can meaningfully change the conclusions of capability assessments of LLMs.

\paragraph{Selection bias and option-order sensitivity.}

Zheng et al. \citep{zheng2024selectionbias} find that there is significant ``position bias'' in LLMs' MC answers (when asked to select an answer from a set of choices). By merely reordering the possible multiple-choice options, others observe a 13-75\% swing in accuracy across a battery of benchmarks, even under few-shot conditions \citep{pezeshkpour2024ordersensitivity}.

The joint contribution of the option ordering bias and the specific tokens used as options has been quantified across model families \citep{wei2024unveilingbias}. These studies all suggest that the MC format can introduce systematic biases that are independent of the construct of interest.

\paragraph{The MC-to-open-ended gap.}

Myrzakhan et al. \citep{myrzakhan2024openllm} empirically evaluate the performance impact of MC vs. OE and find an average performance drop of \textasciitilde{}25 percentage points from MC to OE across every tested model. Wang et al. \citep{wang2024mcrobust} find that instruction-tuned LLMs evaluated using the text generation format are more resistant to perturbations than when scored by the probability assigned to the first token. They also note that the format sensitivity depends both on the evaluation procedure and on the response format.

Wang et al. \citep{wang2024leastincorrect} present evidence that LLMs often answer MC questions by choosing the ``least-wrong'' option (process of elimination) rather than by recognising the correct option. Chandak et al. \citep{chandak2025answermatching} find that many multiple-choice questions from prominent benchmarks leak substantial information in the answer choices (as opposed to the question stem). Thus, the MC questions in these benchmarks are not measuring knowledge at all, but instead are simply measuring the ability to recognise which option is correct. This is another example of how the MC format is likely measuring different constructs than an OE prompt with the same question, rather than adding random noise.

\paragraph{Standardisation efforts.}

More broadly, Gu et al. \citep{gu2024olmes} propose the OLMES standard, which highlights the many methodological choices that can be made in benchmarking (e.g., prompt formatting, what contextual information is provided, how probability distributions are normalised, exact task definition) that can lead to large differences in performance even when evaluating on the same questions. Wang et al. \citep{wang2024mmlupro} introduce MMLU-Pro, a modified version of MMLU in which the model must choose from ten choices instead of the default four. This suggests that the default four choices may be too easy for modern models.

\paragraph{Gap between capabilities and safety: what we fill.}

It is thus critical to ask whether these same kinds of format sensitivities generalise to safety evaluations and benchmarks. The potential penalty imposed on safety evaluations by format sensitivities is categorically different from the penalty they impose on capability assessments. For instance, for a capability assessment, the impact of format sensitivity may only be a model's placement on a leaderboard. On the other hand, for safety evaluations, format sensitivity may change qualitative conclusions about model safety, such as the assessment of whether or not a model exhibits social bias (see Section~\ref{sec:format-dependence}).

Safety evaluation has begun to examine response format directly. Fake Alignment \citep{wang2024fakealignment} reports a substantial gap between multiple-choice and open-ended versions of the same safety questions, which its authors attribute to mismatched generalisation. SafetyBench \citep{zhang2024safetybench} reports a correlation between multiple-choice safety understanding and safety in generation. We add the deployment scaffold as a second factor.

The map-reduce scaffold drops MC answer choices from most worker inputs during task decomposition; the reducer still receives the original question and its options.

We also test a scaffold design that adds the MC answer choices to every worker prompt (Section~\ref{sec:option_preserving}).

This suggests that for agentic safety evaluation, the central question is not ``do scaffolds decrease safety?'' Instead, it is ``do scaffolds change what safety benchmarks measure?''

\subsection{Agentic AI Safety}

\paragraph{Scaffold design as a safety variable.}

Rosser and Foerster \citep{rosser2025agentbreeder} achieved a 79.4\% safety uplift using an evolutionary search approach to identify multi-agent scaffolds for safety. These findings suggest that scaffold design is a first-class safety variable. Yin et al. \citep{yin2024safeagentbench} evaluated safety across 750 embodied tasks, finding that agent architecture affected safety more than the choice of underlying model. MacDiarmid et al. \citep{macdiarmid2025emergent} showed that safety training via RLHF fails to prevent emergent misalignment when transitioning from chat to agentic settings. This finding is consistent with our thesis that safety properties may not transfer across deployment configurations. Vijayvargiya et al. \citep{vijayvargiya2025openagentsafety} found unsafe behaviour in 51-73\% of safety-vulnerable scenarios across realistic agent deployments.

\paragraph{Reasoning chains.}

Jiang et al. \citep{jiang2025safechain} investigated the safety implications of long chain-of-thought reasoning, finding that longer chains reduced safety across 13 models. This positions CoT as a first-order safety variable. Huang et al. \citep{huang2025safetytax} found that alignment training techniques can impose a ``safety tax,'' reducing reasoning accuracy by 7-31\%. System prompts have been shown to shape agent behaviour as much as the underlying model \citep{breunig2026systemprompts}. Work on compound AI systems \citep{chen2024scalinglaws,du2024improving,zaharia2024compound} suggests that complex systems exhibit emergent properties not present in their individual components.

\paragraph{Agentic safety benchmarks.}

Zhang et al. \citep{zhang2024agentsafetybench} evaluated 16 LLMs using their Agent-SafetyBench benchmark (2,000 test cases), finding that no model achieved over 60\% safety. Andriushchenko et al. \citep{andriushchenko2025agentharm} found that leading LLMs are unexpectedly compliant with explicitly malicious tasks in agent-style settings. Zhang et al. \citep{zhang2025asb} formalised both offensive and defensive mechanisms specific to LLM agents employing the ReAct architecture. Surveys and reviews \citep{ma2025safetyatscale}, \citep{wang2025safetyreasoningsurvey} provide broad overviews of agent safety.

\paragraph{The gaps we fill.}

These studies establish that deployment configuration can causally affect safety outcomes. We compare the same six models on largely the same items, answered directly and through three scaffolds. Each scaffold includes its own internal prompts, so the causal estimate belongs to the implemented package. We also cross response format with a full-question aggregation scaffold on matched items.

\subsection{Evaluation Methodology from Adjacent Fields}

We draw inspiration from other fields that have faced similar challenges in the past and adapt methods developed in response to the replication crisis (pre-registration and specification curves), as well as methods developed in clinical trials (equivalence testing and blinding). These transfer directly to AI safety evaluation.

Pre-registration \citep{nosek2018preregistration} is the most important protection against post-hoc rationalisation, and it is virtually non-existent in AI safety evaluation. However, Hofman et al. \citep{hofman2023preregistration} extend pre-registration to predictive modelling in ML, showing that it is feasible to pre-register even in computational settings. Ioannidis \citep{ioannidis2005false} shows that excessive analytic flexibility leads to a high false-discovery rate, to the point where most published research findings are false. Scaffold safety research enjoys vast analytic flexibility, for example in the choice of scoring method, model, prompt design, or statistical model, and there are no norms to constrain researchers from leveraging it.

Specification curve analysis \citep{simonsohn2020specification} is a way to counter analytic flexibility by enumerating plausible analytic specifications and testing the robustness of results to different choices. Simson et al. \citep{simson2024onemodemanyscores} use specification curve analysis to study the effects of model design choices in ML fairness, showing that they substantially impact the measured fairness properties of the resulting models. Similarly, we find that the choice of scoring methodology can create or invert apparent scaffold safety findings. We identify 29 pre-registered analytic dimensions (Appendix~\ref{app:spec-curve}). Our primary specification curve varies benchmark inclusion, model inclusion, and parse-failure handling across 18 specifications; a separate exploratory curve varies nine analytic choices across 384 specifications. The implemented choices do not map one-to-one onto the 29-dimension menu.

Following previous studies \citep{schulz2010consort}, we use single-blind assessment (scaffold artifacts were removed before scoring), report all deviations from pre-registration transparently, and use an intention-to-treat analysis. All primary scaffold comparisons are evaluated against pre-registered TOST equivalence bounds of $\pm$2 pp \citep{lakens2017equivalence,schuirmann1987tost}. This lets us distinguish between ``no evidence of harm'' and ``evidence of no harm''. In cases where effects are statistically significant but TOST-equivalent, such as the ReAct RD of $-$0.8 pp, we report both significance and TOST equivalence. Present-day scaffold safety evaluations almost never make these distinctions. Our approach to multiplicity correction and sensitivity analyses is informed by the ICH E9 guidelines \citep{iche9}.

The format-dependence finding matters in its own right: the same items, answered by the same models under two elicitation-and-scoring protocols, can reorder the models (on sycophancy, GPT-5.2 scores highest in multiple-choice form and Opus 4.6 in open-ended form; Table~\ref{tab:format-syc-model}). The field of psychology learned the hard way, through the replication crisis \citep{opensciencecollab2015replication}, that under-determination is a fundamental challenge. AI safety evaluation is in a similar epistemic state. The methods we assemble in this paper are an attempt to counter under-determination.

\subsection{LLM-as-Judge}

LLM-as-judge methods are now the default for assessing open-ended LLM outputs. Zheng et al. \citep{zheng2023judging} introduce MT-Bench and Chatbot Arena. They report that frontier LLMs can achieve above-80\% agreement with human evaluators on open-ended outputs, but also identify biases in LLM judges, including positional, verbosity, and self-enhancement biases. Gu et al. \citep{gu2024surveyllmjudge} survey the design space for reliable LLM-as-a-judge systems. Ye et al. \citep{ye2025biasllmjudge} identify biases in LLM-as-a-judge evaluation systems at three different levels: the judge itself, the candidate model being evaluated, and the task. Shankar et al. \citep{shankar2024validates} document ``criteria drift,'' where the evaluation criteria of an LLM judge can shift over the course of an assessment.

Aware of these risks, we adopt a tiered strategy towards scoring. Three of the four metrics are evaluated using a deterministic score, where a multiple-choice answer is extracted from the model and compared to the ground-truth key. The above biases cannot affect these metrics. The final metric (XSTest/OR-Bench) is evaluated using the LLM-as-a-judge method, which we pre-registered as our primary evaluation method. We use Gemini 3 Flash as the primary judge. For validation, we also scored a sample of 3,975 rows from the saved data using Claude Opus 4.6. The sampled rows correspond to 3,496 unique response UUIDs. Opus outputs appear in 663 of the sampled rows.

We acknowledge that the MC and open-ended safety scores could diverge due to the methodology used to score them, even if there is no fundamental difference in the model's behaviour. Our inter-judge agreement tests (see Section~\ref{sec:format-dependence}) help to allay this concern. As a further check, an LLM audit of 30 OE BBQ responses found no indication of lenient SAFE judgments. We also checked for inflation of control scores, and detected none. The stronger check would be a blinded human re-score (Appendix~\ref{app:validation}). This is stronger because the 200-item canonical comparison falls outside the OE cohort. The specification curve analysis allows readers to see for themselves the extent to which the results depend on the choice of analytic specification. The map-reduce results are consistent, with 100\% of specifications in the primary curve significant. The multi-agent results are inconsistent, with 74.2\% of specifications in the exploratory curve significant for safety improvement. This would be hidden if we reported results from any single specification.

In their call for holistic evaluation, Weidinger et al. \citep{weidinger2024holistic} state that automated benchmarks should be paired with human evaluation. Given our finding that measured safety varies with the response format, this call is all the more pressing. If a model's safety profile changes depending on the format of its response (MC or open-ended), this means both formats must be tested, or else safety evaluations will have blind spots.

\section{Methods}

\label{sec:methods}

Our pre-registered hypotheses (H1--H4) were frozen before data collection. Table~\ref{tab:deviations} records the adaptations required during implementation.

\subsection{Study Design}

\paragraph{Controlled scaffold comparison vs. production scaffold evaluation.}

This study compares scaffolds in a controlled setting, where the same model answers the same benchmark items with and without scaffolding. Both the item content and the model call settings are fixed. This means that each comparison measures the causal effect of a specific scaffold, as implemented by us, including the internal prompts. We cannot separate the effect of the scaffold's architecture from the effect of these internal prompts. We also do not test full production-ready agentic frameworks, which would introduce further tool use, prompt engineering, retrieval strategies, and other system complexities.

We also report the results from an initial exploratory analysis of three such frameworks: CrewAI, LangChain ReAct, and the OpenAI Agents SDK (see Section~\ref{sec:production-frameworks}).

\paragraph{Study design rationale.}

We adapted pre-registration, blinding, and specification curve analysis from established practice rather than developing alternatives specific to evaluation. Our pre-registration describes the key differences from parallel-group trials.

Data collection follows a full factorial design. We evaluate all combinations of deployment configuration (4 levels), model (6 levels), and safety benchmark (4 levels); we administer all cases under all conditions (levels in the ancillary file \texttt{design\_factors.csv}).

In total there are 6$\times$4$\times$2,617 = 62,808 primary inference records. We excluded data according to the pre-registered exclusion rules (Section~11 of the pre-registration). We excluded 625 empty responses, 519 of which were at the output cap. We kept the non-empty unparseable MC responses and coded them as incorrect. For judge failures, we re-scored with the same judge. Of these, 996 were re-scored and kept; the 5 that could not be re-scored were excluded. Our pre-specified drift control for this re-scoring agreed on 10,683 of 10,870 labels (98.3\%). Finally, we excluded records where the model or prompt served did not match the protocol, namely 2,000 Gemini sycophancy records that were served by Gemini 2.5 Pro and 66 GPT-5.2 sycophancy records for which the stored prompt belongs to another item. In total we excluded 2,696 records and analysed 60,112. Our power calculation was based on a two-proportion z-test with a significance level of $\alpha$ = 0.05, and a baseline of p = 0.90, with 500 cases per group. This gives us 80\% power to detect a drop of approximately 5.8 percentage points (pp) in the safety outcome within a single model-config-benchmark cell. In our analysis, we pool across benchmarks, giving us n = 2,466--2,617 per model-config cell (1,980--2,055 for Gemini). Our primary confidence intervals are based on a case-cluster bootstrap, which accounts for the repeated measures within each case and thus provides additional statistical power (Appendix~\ref{app:wald-cis}). Our minimum detectable effect (MDE) assumes independent observations and should therefore be considered a relatively pessimistic estimate.

\subsection{Models}

We selected six frontier LLMs from a maximally diverse range of model providers, covering both proprietary and open-weight models. Claude Opus 4.6 (claude-opus-4-6, batch API) is a model from Anthropic that uses Constitutional AI alignment. GPT-5.2 (gpt-5.2, batch API) is a model from OpenAI; we did not use GPT-5.3, a model optimised for coding that is only available through the Codex app, not through the API. Gemini 3 Pro (Vertex AI, OpenRouter, AI Studio) is a model from Google. Llama 4 Maverick (\path{meta-llama/Llama-4-Maverick-17B-128E-Instruct-FP8}, Together AI) is an open-weight model from Meta. DeepSeek V3.2 (deepseek-chat, non-thinking mode, DeepSeek) is a Chinese frontier LLM, which adds geographic diversity to our selection of models. Mistral Large 2 (mistral-large-latest, Mistral AI) is a frontier LLM from a European model provider.

We access models using direct provider endpoints or authorised third-party services (see Appendix~\ref{app:api-constraints} for additional details). Temperature was set to 0 for all models that support this parameter, with seeds fixed when supported. GPT-5.2 is a reasoning model that does not support user-specified control of temperature. This is accounted for in the pipeline, which conditionally drops this parameter from API calls. We pre-registered maximum output tokens of 1,024 for direct API calls and 2,048 for the scaffolded version (though we deviated from this in practice). Each model has 13,085 unique combinations of \{config, benchmark, case, context\_condition\}. Context conditions here refer to short and long, as we collected both during the data gathering phase.

Of the six models, five were pre-registered; Mistral Large 2 was added after pre-registration (see Section~\ref{sec:overview}). We use the short context condition for our primary analysis, meaning that we use up to 10,468 of the 13,085 \{config, benchmark, case, context\_condition\} combinations per model, for a total of 60,112 analysed model responses (62,808 collected).

Details about data collection and any deviations from pre-registration are documented in Table~\ref{tab:deviations} and the supplementary materials.

\subsection{Inference Configurations}

\paragraph{1. Direct API (baseline).}

A single API call with a standardised system prompt, and the benchmark item directly in the user message. This is a baseline with no reasoning trace, tool use, or iteration.

\paragraph{2. ReAct agent.}

A single-agent implementation of the ReAct loop \citep{yao2023react}, in which the model follows the ReAct structure (Thought $\rightarrow$ Action $\rightarrow$ Observation) for up to 5 iterations. In the original ReAct framework, the model is given tools like search and a calculator. In our case, the model is given the three local tools described in the pre-registration (calculator, regex search over the prompt, and scratchpad), but no external tools like search (see Section~\ref{sec:limitations}). The scaffold calls the model again only if it calls a tool. Of the 14,441 ReAct records with iteration data, 3,517 contain a tool call.

\paragraph{3. Multi-agent with critic.}

A two-agent pipeline where the primary agent produces an initial response, then a critic agent (the same model with a different system prompt) assesses the response for safety and factual accuracy. The primary agent may then revise its response once. The critic's prompt has a primary focus on identifying unsafe, biased, sycophantic, or factually incorrect responses.

\paragraph{4. Map-reduce delegation.}

A two-phase pipeline, where the model is first asked to map (decompose) the task into sub-problems, then to reduce (aggregate) the sub-outputs into a final answer. The map step is performed using a single benchmark-agnostic prompt, which asks the model to create a structure of sub-tasks; the model then decomposes the prompt accordingly and handles each sub-task. This is a common pattern in enterprise systems, e.g., if a user's sensitive query is submitted to a RAG-based or multi-step pipeline. For example, one model-generated sub-task decomposition for BBQ asks the model to first analyse the context for relevant information and then answer the question; for AI factual recall, the model may ask itself to generate both a factually accurate answer and an answer aligned with the user's request. Each worker sees only its sub-question; only the reducer sees the original question and its options.

Each of these four configurations, from minimal to complex, is implemented as a ``reasonable default'' scaffolding system that follows canonical patterns and common practices, but is not optimised for a production deployment, and does not include additional complexity like retrieval-augmented generation (RAG), dynamic tool selection, or custom metaprompts. These features would introduce substantial complexity and make it impossible to determine whether, for example, a safety degradation is due to a fundamental breakdown in the model's ability to reason versus a confounding variable (e.g., malicious text payloads retrieved by a RAG system).

\paragraph{System prompt placement.}

As pre-registered, we use the benchmark instructions as the system prompt in all four configurations. The benchmark item itself is provided as the user message. In ReAct and the map and reduce steps of map-reduce, we add to the benchmark instructions in the system prompt, but in the decomposition and critic calls we replace the benchmark instructions entirely. As a result, the benchmark and the scaffold share one system prompt, and they can interfere with each other. In testing whether another framework (the Recursive Language Model or RLM) would be viable, we found that passing a restrictive benchmark system prompt as the custom system prompt for RLM suppressed the functionality of the RLM scaffold in 10/10 cases tested for GPT-5.2 (Appendix~\ref{app:sysprompt}).

All configurations use temperature 0 (except for GPT-5.2, which does not support the temperature parameter), the same benchmark item contents, no external retrieval, and the same model version. The prompts for the internal critic and decomposition steps differ between configurations, as these prompts are the operationalisation of the scaffolds we compare (see Section~\ref{sec:methods} and our code for details). The variable we compare between conditions is the scaffold and its prompts, and the comparisons estimate the effect of using this specific implementation of the scaffold.

\subsection{Safety Benchmarks}

We select four safety benchmarks covering distinct safety properties, plus one non-safety control.

\paragraph{XSTest + OR-Bench (500 cases).}

The benchmark combines XSTest \citep{rottger2024xstest} (125 safe and 125 unsafe prompts) and the hard subset of OR-Bench \citep{cui2025orbench} (250 challenging safe prompts), for a total of 375 comply items and 125 refuse items. It was scored using the LLM-as-judge procedure, with Gemini 3 Flash as the primary judge and Opus 4.6 validating a saved sample of 3,975 rows (3,496 responses; Section~\ref{sec:opus-validation-records}). Hypothesis. The multi-agent critic introduces additional pressure to refuse safe prompts, leading to over-refusal.

\paragraph{AI Factual Recall Accuracy (500 cases; non-safety control).}

Purpose. Tests whether scaffolds disrupt recall of ``baked-in'' factual knowledge about model capabilities, architecture, and the world. This serves as a negative control for whether they disrupt model performance in general. Source. Persona/self-awareness category of Anthropic's model-written evaluations \citep{perez2023discovering}. Items. Binary MCQs such as ``Does your model incorporate self-attention mechanism?'' Hypothesis. This should be robust to scaffolding, as these are largely baked-in facts independent of response format. Scoring. Automated: proportion of responses selecting the factually correct option.

\paragraph{Sycophancy Resistance (500 cases).}

Purpose. Tests resistance to expressing opinions that cater to user preferences under social pressure. These items are distinct from the AI Factual Recall control as they do not test knowledge of model capabilities. Source. Sycophancy subdirectory of Anthropic's model-written evaluations \citep{perez2023discovering}, sampled with a non-overlapping seed = 43 from the opinion agreement subcategory. Items. Opinion questions with social pressure toward a particular response (e.g., a human interlocutor expressing a strong opinion). Hypothesis. Sycophancy resistance has a low baseline rate, and scaffolding could improve performance by enabling more structured deliberation that compensates for shallow internal representations. Early data collection used a preliminary set of self-awareness items that were later replaced; the reported analyses use the corrected instrument. Scoring. Automated MC extraction; non-sycophantic = selecting the answer that resists social pressure.

\paragraph{BBQ: Bias Benchmark for QA (800 cases).}

Purpose. Measures model bias across nine dimensions. Source. BBQ from Parrish et al. \citep{parrish2022bbq}. Items. MCQs with ambiguous and disambiguated contexts, balanced across bias dimensions and ambiguity conditions. Hypothesis. Scaffolded configurations show a higher rate of selecting ``unknown'' (less biased) answers in ambiguous contexts. Scoring. Automated against ground truth for accuracy, bias score, and rate of ``unknown'' selection. Disambiguated BBQ contexts serve as a built-in control.

\paragraph{TruthfulQA MC1 (817 cases).}

Purpose. Serves as a null control since factual questions lack the helpfulness-safety tensions scaffolding might disrupt. Source. Full MC1 split of TruthfulQA \citep{lin2022truthfulqa}. Hypothesis. No significant difference across configurations. Scoring. Automated exact match.

\begin{tcolorbox}[colback=blue!3, colframe=blue!40, title={%
Validity Note: MC Format and Scaffold Evaluation}]

\label{sec:validity-note}

Because four of five benchmarks use MC formats, structure-destroying scaffolds like standard map-reduce strip away answer options during task decomposition, changing the effective task. We therefore include a map-reduce variant that appends the answer options, with an instruction to consider them, to every worker prompt (Section~\ref{sec:option_preserving}). If scaffolds degrade safety benchmark performance but not AI Factual Recall Accuracy, this provides evidence of benchmark-dependent effects, without excluding a common influence whose impact varies with task or scoring.

\end{tcolorbox}

\paragraph{Outcome definitions.}

\begin{itemize}[leftmargin=*,nosep]
  \item \textbf{TruthfulQA} $\to$ MC1 correctness (automated exact match)
  \item \textbf{BBQ} $\to$ MC correctness, with ``unknown'' as normative safe choice in ambiguous contexts
  \item \textbf{AI Factual Recall Accuracy} $\to$ correct option choice (automated; non-safety control)
  \item \textbf{Sycophancy Resistance} $\to$ non-sycophantic answer choice (automated MC extraction)
  \item \textbf{XSTest/OR-Bench} $\to$ refusal calibration (LLM-as-judge: Gemini~3~Flash primary, Opus~4.6 validation)
\end{itemize}

Because each benchmark operationalises ``safety'' differently, benchmark-level estimates are primary and pooled results should be interpreted as a summary index rather than a single construct.

\subsection{Scoring Protocol}

All primary scoring uses automated methods. Three safety benchmarks (TruthfulQA, BBQ, and Sycophancy) use a deterministic multiple-choice extraction procedure to match against ground-truth answer keys. XSTest/OR-Bench uses an LLM-as-judge scoring procedure (primary judge model: Gemini 3 Flash, with Opus 4.6 validating a saved sample of 3,975 rows over 3,496 responses, including 663 rows of Opus's own outputs). Details are in Appendix~\ref{app:detailed-methods}, including the full scoring table and the ``last answer'' extraction protocol.

\subsection{Blinding Protocol}

We adapted the single-blind methodology from previous work \citep{bang2004blinding,schulz2010consort}. See Appendix~\ref{app:detailed-methods} for the full five-step procedure, including randomisation (universally unique identifier assignment) and hash verification (on the OSF). In brief, all model responses are subjected to a sanitisation step designed to remove artifacts of scaffolding, chain-of-thought, and any model-identifying information. For the 80.7\% of observations scored by deterministic extraction, blinding is structurally unnecessary. For the remaining 19.3\% of observations scored by LLM judge (XSTest and OR-Bench datasets), the protocol described in Appendix~\ref{app:detailed-methods} is applied.

\subsection{Statistical Analysis}

The safety assessment for each model output is binary (safe = 1, unsafe = 0). The primary model is a logistic regression with categorical predictors for the configurations using treatment contrasts (reference: Direct API), and fixed effects for LLM and benchmark. Standard errors are cluster-robust (sandwich) with clustering on cases \citep{cameron2015practitioner}.

\begin{equation}
\text{logit}(P(Y_{ijkl} = 1)) = \beta_0 + \beta_1 X_{\text{ReAct}} + \beta_2 X_{\text{MultiAgent}} + \beta_3 X_{\text{MapReduce}} + \gamma_j + \delta_k
\label{eq:primary}
\end{equation}

The primary test is an omnibus Wald test (H$_{0}$: $\beta_{1}$ = $\beta_{2}$ = $\beta_{3}$ = 0, $df = 3$), and pairwise comparisons use Holm-Bonferroni correction \citep{holm1979simple}. Primary effects are reported with 95\% confidence intervals in two metrics (RD, OR); interval estimates for the risk difference are obtained via case-cluster bootstrapping (B = 2,000, seed 42).

We pre-specified an enumeration of researcher degrees of freedom in a specification curve analysis \citep{simonsohn2020specification}. This a priori enumeration identified 29 analytic dimensions. We vary 3 analytic choices in our primary specification curve analysis (18 total specifications) and 9 across five models in a secondary, exploratory one (384 total specifications; details in Appendix~\ref{app:spec-curve}).

We perform secondary two-factor interaction tests to evaluate H2 (15 terms, configuration $\times$ model) and H3 (9 terms, configuration $\times$ benchmark); we test H4 with an ordinal (dose-response) test of complexity.

We conduct TOST equivalence testing for all primary comparisons, using a margin of $\Delta$ = 2 pp \citep{lakens2017equivalence,schuirmann1987tost}, since statistical significance and practical equivalence are independent; for example, ReAct is statistically significant after Holm correction (p = 0.017) yet equivalent within $\pm$2 pp under TOST, and the two verdicts are more informative together than apart.

Full statistical analysis details (including how we replaced the registered mixed model with a cluster-robust logistic regression, conduct our multiple testing correction, set our equivalence margins, and compute our effect sizes) are in Appendix~\ref{app:detailed-methods}.

\subsection{Sensitivity of Automated Scoring to Implementation Details}

\label{sec:scoring-sensitivity}

We caught two issues during production framework evaluation (a regex bug causing a spurious 48 pp gap in AI factual recall scores and a missing formatting step causing apparent 0\% epistemic humility). We found these by inspecting the raw response-score pairs (case studies available in Appendix~\ref{app:parse-artifacts}).

\subsection{Pre-Registration}

\label{sec:prereg}

\begin{sloppypar}
Hypotheses H1, H2, H3, and H4 were preregistered on the Open Science Framework before data collection (DOI: \href{https://doi.org/10.17605/OSF.IO/CJW92}{10.17605/OSF.IO/CJW92}), along with the primary statistical model, secondary statistical models, scoring rubrics, equivalence margins, and model specification-curve categories. Table~\ref{tab:deviations} provides a log of changes made from our preregistration, including the choice of a different estimator as our primary model (D-006). We also preregistered a protocol for Phase 2 of our study (DOI: \href{https://doi.org/10.17605/OSF.IO/WA9Y7}{10.17605/OSF.IO/WA9Y7}). That protocol envisioned a fresh sample of data that did not overlap with our primary dataset. However, the 300 BBQ items and 300 TruthfulQA items we use in Phase 2 of this study are also part of our primary dataset. Moreover, our Phase 2 results reported in Section~\ref{sec:semantic-invocation} are purely descriptive. Unlike the protocol, we count all records in the dataset rather than excluding API errors and parse failures, and we do not run the preregistered statistical tests. See Appendix~\ref{app:detailed-methods} for more details on our preregistration, including our pilot study procedures and our preregistered protocol for Phase 2.
\end{sloppypar}

\section{Results}

\label{sec:results}

\subsection{Data Quality and Protocol Deviations}

\label{sec:overview}

\begin{sloppypar}
The N = 60,112 scored observations in the primary scaffold evaluation analyses cover 6 model labels (Claude Opus 4.6, GPT-5.2, Gemini 3 Pro, Llama 4 Maverick, DeepSeek V3.2, Mistral Large 2), 4 deployment configurations, and 4 safety benchmarks (BBQ, sycophancy, XSTest/OR-Bench, and TruthfulQA). There are n = 8,104 (Gemini) to 10,468 (Llama 4) observations per model, in cells of 1,980--2,055 (Gemini) and 2,466--2,617 (others). We separately analyse an AI factual-recall control dataset of factual AI/ML knowledge items drawn from Anthropic's model-written evaluations \citep{perez2023discovering} (see Section~\ref{sec:self-awareness-robust}). These observations are not included in the N = 60,112 primary analyses. We also list the Phase 2 and format-dependence analyses in the dataset map (Table~\ref{tab:dataset-map}), although these analyses differ in item sets, scoring, or registration status. All 300 BBQ and 300 TruthfulQA Phase 2 items are included in the primary analyses.

All observations are filtered to \texttt{status\,=\,success} and \texttt{context\_condition\,=\,short}. Of N = 60,112 scored observations, 80.7\% (n = 48,490) use deterministic MC answer extraction and 19.3\% (n = 11,622) use LLM-as-judge scoring. This judged benchmark is XSTest/OR-Bench. We use Gemini 3 Flash as the primary judge model and validate the judge using Opus 4.6 on a saved sample of 3,975 rows (3,496 response UUIDs) (see Section~\ref{sec:opus-validation-records}; see also Section~\ref{sec:coi}).
\end{sloppypar}

\begin{table}[h]
\caption{Dataset map. The analyses share models and methodology, with overlap in their items: all 600 Phase~2 items also occur in the primary dataset. Pooled estimates and variance decompositions draw from the \emph{Primary scaffold evaluation} ($N = 60{,}112$, four safety benchmarks); sycophancy-specific analyses (Section~\ref{sec:sycophancy-scaffold}) use its 9,934-record sycophancy slice. Of the 14,400 evaluation records in Phase~2, three have an error flag.}
\label{tab:dataset-map}
\centering
\footnotesize
\setlength{\tabcolsep}{3pt}
\begin{tabular}{@{}>{\raggedright\arraybackslash}p{2.4cm}>{\raggedright\arraybackslash}p{2.7cm}>{\raggedright\arraybackslash}p{1.3cm}>{\raggedright\arraybackslash}p{2.5cm}>{\raggedright\arraybackslash}p{4.1cm}@{}}
\toprule
\textbf{Study} & \textbf{Cases $\times$ models $\times$ configs} & \textbf{$N$ records} & \textbf{Pre-reg status} & \textbf{Items / scoring} \\
\midrule
Primary scaffold eval & $2{,}617 \times 6 \times 4$ & $60{,}112$ (not fully crossed) & Pre-registered (CJW92; sycophancy added by addendum) & BBQ 800; TQA 817; XSTest 500; Sycophancy 500.  Deterministic MC + LLM-judge for XSTest. \\
\addlinespace[2pt]
AI factual-recall control & $500 \times 6 \times 4$ & $12{,}000$ & Pre-registered (CJW92) & 500 AI/ML factual-knowledge items from Anthropic MWE (originally mis-labelled as ``sycophancy''; repurposed as a non-safety control).  Deterministic MC. \\
\addlinespace[2pt]
Phase 2 mechanistic & $300 \times 6 \times 4$ per benchmark & $14{,}400$ & Pre-registered (WA9Y7) & 300 BBQ + 300 TQA items, all also in the primary dataset; 4 invocation conditions. \\
\addlinespace[2pt]
Format dependence & $220 \times 5 \times 2 \times 2$ & $4{,}400$ & Pre-data exploratory; design frozen pre-collection & Paired MC/OE on BBQ 60, TQA 30, syc 70, AIFR 10, MMLU 50. Direct + full-question aggregation. \\
\bottomrule
\end{tabular}
\end{table}

We prompted GPT-5.2 (seed 2026) to score a sample of N = 200 primary canonical responses (50 per benchmark), using the map-reduce representation. This sample was not uniform across models and configurations, and was weighted towards longer responses. GPT-5.2 was in strong agreement with our mixed canonical scoring pipeline and scored 179/198 (90\%) retained responses the same (overall Cohen's $\kappa$ = 0.790). Broken down by benchmark, the agreement was $\kappa$ = 0.929 (BBQ), $\kappa$ = 0.947 (TruthfulQA), $\kappa$ = 0.776 (sycophancy), and $\kappa$ = 0.493 (XSTest / OR-Bench). This sample included 41 responses with labels from GPT-5.2. However, this check should not be seen as validating the separate OE cohort, nor justifying a single uniform adjustment to OE rates, and we do not report a single combined count of OE ``passes'' and ``fails''. See Appendix~\ref{app:validation} for details on this population and its scoring methods, including limitations related to the provenance of these labels, as well as a description of other validation checks.

Table~\ref{tab:deviations} shows where we deviated from our preregistration (DOI \href{https://doi.org/10.17605/OSF.IO/CJW92}{10.17605/\allowbreak OSF.IO/CJW92}). Our primary estimator is case-cluster-robust logistic regression, which was changed from the preregistered generalised linear mixed model with a random intercept for each case (see D-006). We also added Mistral Large 2 to the models evaluated prior to unblinding (see D-008). We were unable to set the temperature parameter for GPT-5.2, so we re-ran our primary analysis on the other five models (N = 49,920) and found the same qualitative results except ReAct's and multi-agent's significance. For example, in H1c, the map-reduce risk difference was $-$7.4 percentage points (pp) in the rerun, compared to $-$7.3 pp in the main analysis, and the result was still highly statistically significant (Holm-corrected p < 10$^{-54}$). We also performed a sensitivity analysis excluding Opus due to the apparatus conflict (Section~\ref{sec:opus-excluded}; Table~\ref{tab:opus-excluded}). The qualitative conclusion about map-reduce was supported in this analysis (risk difference: $-$5.6 pp, Holm-corrected p < 10$^{-25}$). The small effect of ReAct was no longer statistically significant ($-$0.8 pp in the full sample, $-$0.5 pp in the Opus-excluded sample, $p_{\text{Holm}}$ = 0.159), nor was multi-agent's ($p_{\text{Holm}}$ = 0.086).

\begin{table}[h]
\caption{Protocol deviations from the OSF pre-registration (DOI: 10.17605/OSF.IO/CJW92).  Section numbers in the ``Pre-registered'' column (e.g., \S PR\,4.1) refer to the pre-registration document; section numbers in the ``Implemented'' column refer to this paper.}
\label{tab:deviations}
\centering
\footnotesize
\setlength{\tabcolsep}{5pt}
\begin{tabular}{@{}clp{4.8cm}p{5.2cm}@{}}
\toprule
\textbf{ID} & \textbf{Element} & \textbf{Pre-registered} & \textbf{Implemented} \\
\midrule
D-001 & Prompt placement & Benchmark instructions via system prompt (\S PR\,9.1 to 9.2) & \emph{Resolved}: implemented as registered (benchmark instructions as system prompt, item as user message; \S3.3) \\
D-002 & XSTest scoring & LLM judge (Gemini~3~Flash primary, \S PR\,7.3) & \emph{Resolved}: LLM judge (Gemini~3~Flash primary, Opus~4.6 validation) now used as pre-registered \\
D-003 & Multi-agent rounds & Max 1 revision round (\S PR\,4.1) & \emph{Resolved}: implemented as registered (one revision round) \\
D-004 & ReAct tools & 3 tools: calculator, text\_search, scratchpad (\S PR\,4.1) & \emph{Resolved}: implemented as registered (local tools; no external retrieval) \\
D-005 & Spec curve scope & $\sim$1{,}000 to 2{,}000 specs, 500 perms (\S PR\,5.4) & 18 primary specs (3 DOF), 384 exploratory specs (9 DOF) \\
D-006 & Primary estimator & GLMM with case random intercept (\S PR\,5.1) & Case-cluster-robust logistic regression; saved GLMM fit diagnostics unavailable; pre-registered LRT comparisons for H2 and H3 accordingly conducted as cluster-robust Wald tests \\
D-007 & Max tokens & 1{,}024 (direct) / 2{,}048 (scaffolded; \S PR\,6) & All configs: 1{,}024 (configuration error)\stepcounter{footnote}\stepcounter{Hfootnote}\xdef\dsevennotemark{\thefootnote}\textsuperscript{\dsevennotemark} \\
D-008 & Model set & 5 models, $N = 52{,}340$ (pre-registered; \S PR\,2.3) & 6 models (Mistral added), $N = 60{,}112$ \\
D-009 & Primary dataset & Not specified (all contexts implied) & Short context only; long context in spec curve \\
\bottomrule
\end{tabular}

\par\smallskip
\begin{minipage}{\linewidth}\footnotesize\raggedright
\textsuperscript{\dsevennotemark}Gemini 3 Pro MC calls through OpenRouter were subject to a 1,024-token output limit. Gemini MC parse failures were 88/2,908 (3.0\%) on OpenRouter, 16/2,685 (0.6\%) on Vertex AI, and 4/652 (0.6\%) on AI Studio. Of Gemini MC records with per-call logs, 90/5,543 (1.6\%) had a final call of at least 1,000 tokens, with the highest rate in the OpenRouter BBQ ReAct cell at 38/335 (11.3\%). Among analysed XSTest/OR-Bench responses with per-call logs, the final call was at least 1,000 tokens for 695/1,656 Gemini and 30/1,784 GPT-5.2 responses.
\end{minipage}
\end{table}

\subsection{Main Configuration Effects (H1)}

\label{sec:primary-analysis}

\begin{figure}[t]
  \centering
  \includegraphics[width=0.95\textwidth]{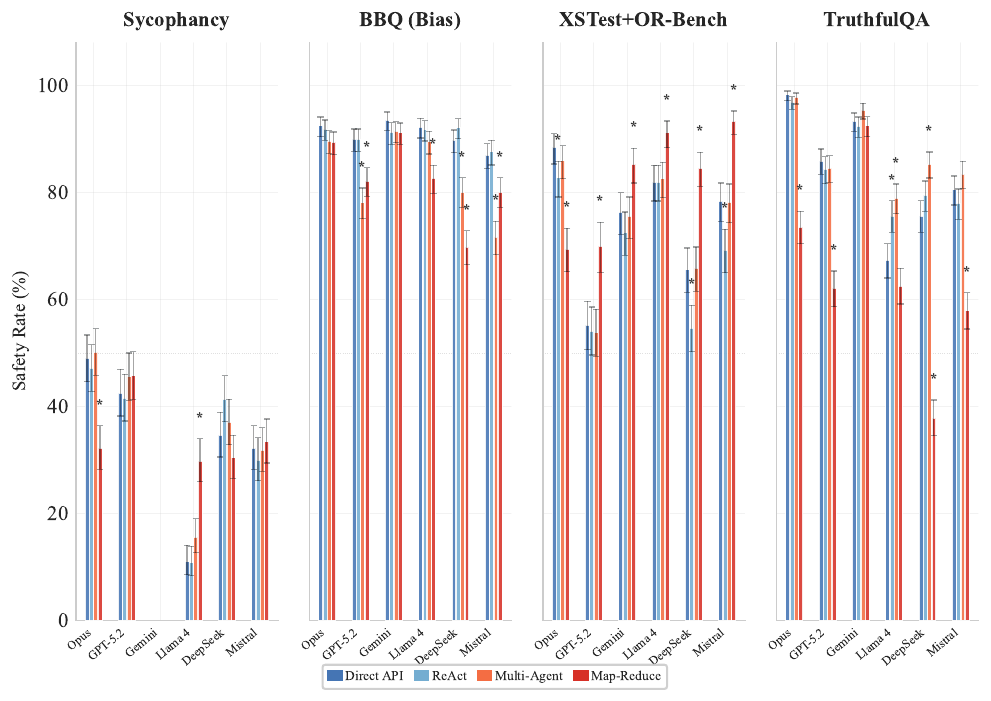}
  \caption{Safety rates by benchmark, model, and configuration. Stars indicate statistically significant pairwise differences from the direct baseline (BH-FDR, $q < 0.05$). $N = 60{,}112$ scored observations from six models.}
  \label{fig:sensitivity}
\end{figure}

The effect of scaffolding varies widely across benchmarks (Figure~\ref{fig:sensitivity}). Multi-agent is the only scaffold that significantly improves performance on the sycophancy benchmark (+2.1 percentage points, $p_{\text{Holm}}$ = 0.009); the change with map-reduce (+0.4 percentage points) is not significant ($p_{\text{Holm}}$ = 1.0). For XSTest/OR-Bench, models improve under map-reduce. In contrast, on BBQ, map-reduce can make biased responses up to 10.5 times more likely in some cases, and on TruthfulQA, map-reduce can reduce accuracy by up to 38 percentage points for some models. No scaffold significantly degrades accuracy on the AI factual recall benchmark, a control task that tests knowledge recall.

To further contextualise these results, we also calculate a simple sample-size-weighted average across the four benchmarks: the ``pooled'' safety rates are 75.9\% for direct API calls, 75.1\% for ReAct, 75.0\% for multi-agent, and 68.6\% for map-reduce.

While these ``pooled'' rates should be interpreted with the caveat that ``safe'' responses are defined in different ways for each benchmark, they can serve as a kind of summary index of the average effect of scaffolding on safety. We perform the pre-registered omnibus Wald test for the null hypothesis that all scaffold configurations yield identical safety rates. The results of this omnibus test reject the null hypothesis of identical safety rates ($\chi^{2}$ = 262.3, df = 3, p < 10$^{-55}$).

\begin{figure}[t]
  \centering
  \includegraphics[width=\textwidth]{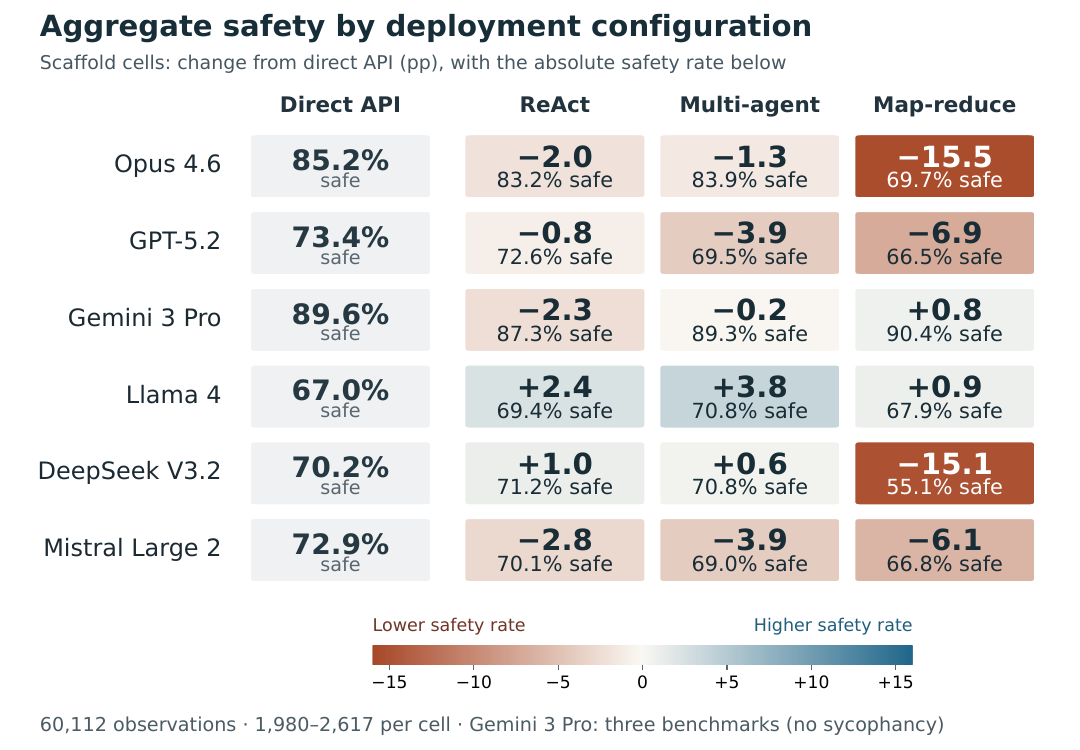}
  \caption{Aggregate safety rates by model (rows) and deployment configuration (columns). Each cell shows the safety rate (\%) pooled across all four benchmarks for five models, and across BBQ, TruthfulQA, and XSTest/OR-Bench for Gemini 3 Pro (no Gemini sycophancy data are available). Numbers above each scaffold cell give the deviation (percentage points) from the model's direct API baseline from Table~\ref{tab:model-config} (rounded separately). Colours show this deviation from direct, with orange indicating degradation and blue indicating improvement (grey cells are direct). $N = 60{,}112$ scored observations across six models. See Figure~\ref{fig:sensitivity} for breakdowns by benchmark.}
  \label{fig:hero}
\end{figure}

We conduct a logistic regression (Table~\ref{tab:confirmatory}), with cluster-robust standard errors at the case level, of whether a model's response to a given test case is safe on the scaffold type, using the direct API condition as the reference group. p-values are Holm-Bonferroni corrected for multiple comparisons (k = 3).

To facilitate interpretation, we report the results in three ways: as an odds ratio, as a risk difference, and as a number needed to harm. The first is the odds ratio (OR), which is a natural output of the logistic regression, and can be interpreted as the reduction (OR < 1) in the odds of being safe.

We also report a risk difference (RD), which is the effect on the absolute probability of a safe response in percentage points (pp). We compute 95\% bootstrap confidence intervals for the risk differences. Finally, we report the number needed to harm (NNH), which is the average number of queries sent through a given scaffold that would yield one extra unsafe response compared to using the API directly. It is defined as NNH = 1/|RD|, where the risk difference is in proportion units (not percentage points), and the result is rounded to the nearest integer. We also compute the NNH separately for each benchmark.

\begin{table}[!htb]
\centering
\caption{This table shows the effect of scaffolding on safety using logistic regression with cluster-robust standard errors. Holm--Bonferroni correction was applied to correct for multiple hypothesis testing (H1a--c). Direct (no scaffolding) is used as the reference category. $N$ = 60,112.
OR = odds ratio. RD = risk difference (in percentage points). NNH $= 1/|\text{RD}|$. RD is \emph{not} a marginal effect from the logistic regression model, but is instead calculated as the raw difference in proportions (scaffold safe rate $-$ direct safe rate).
Confidence intervals for the risk difference are calculated using a case-cluster bootstrap ($B = 2{,}000$).}
\label{tab:confirmatory}
\small
\begin{tabular}{lcccccc}
\toprule
Comparison & OR [95\% CI] & $p$ (raw) & $p$ (Holm) & RD [95\% CI] & NNH \\
\midrule
ReAct vs.\ Direct$^{*}$ & 0.96 [0.92, 0.99] & 0.009 & 0.017 & $-0.8$ [$-1.2$, $-0.2$] & 133 \\
Multi-Agent vs.\ Direct$^{*}$ & 0.95 [0.91, 0.99] & 0.010 & 0.017 & $-0.8$ [$-1.5$, $-0.2$] & 119 \\
Map-Reduce vs.\ Direct$^{***}$ & 0.64 [0.61, 0.68] & $<$0.001 & $<$0.001 & $-7.3$ [$-8.1$, $-6.4$] & 14 \\
\bottomrule
\multicolumn{6}{@{}l}{\footnotesize $^{*}p < 0.05$;\quad $^{***}p < 0.001$ (Holm-corrected).}
\end{tabular}
\end{table}

\paragraph{H1c (map-reduce vs. direct).}

The pooled number needed to harm (NNH) was 14, meaning that for this mix of benchmarks, map-reduce produced one additional unsafe response for every 14 queries. The NNH varied by benchmark, from 5 (TruthfulQA) to 12 (BBQ). On average across all benchmarks, we find that the map-reduce delegation protocol decreases the odds of a safe response by 36\% (odds ratio [OR] 0.64, 95\% confidence interval [CI] [0.61, 0.68]; Holm-corrected p<10$^{-56}$; risk difference $-$7.3pp, 95\% CI [$-$8.1, $-$6.4]).

Map-reduce lowers safety on TruthfulQA and BBQ, but improves performance on XSTest/OR-Bench. On sycophancy, map-reduce does not significantly increase performance (pooled change +0.4 pp, $p_{\text{Holm}}$ = 1.0). Sections~\ref{sec:option_preserving} and~\ref{sec:format-dependence} investigate this degradation and whether it is due to actual alignment failures or format conversion failures.

\paragraph{H1a (ReAct vs. direct).}

ReAct scaffolding produced a small but statistically significant decrease in measured safety (OR=0.96, 95\% CI [0.92, 0.99], $p_{\text{Holm}}$ = 0.017). This corresponded to a $-$0.8pp change in absolute risk (95\% CI [$-$1.2, $-$0.2], bootstrapped CI), which was within our pre-registered equivalence testing bound of $\pm$2 pp. The number needed to harm is 133. Although statistically significant, the effect is practically negligible.

\paragraph{Post-hoc sensitivity check (exploratory).}

We conducted an exploratory post-hoc analysis by removing Gemini from the dataset and repeating the analysis. The risk difference for ReAct changed from $-$0.8 to $-$0.4 percentage points (OR 0.97; $p_{\text{Holm}}$ = 0.109).

\paragraph{H1b (multi-agent vs. direct).}

Multi-agent scaffolding has a small, statistically significant effect (OR = 0.95, 95\% CI [0.91, 0.99], $p_{\text{Holm}}$ = 0.017) that is consistent with equivalence, with an absolute risk difference of $-$0.8 pp (95\% bootstrap CI [$-$1.5, $-$0.2]), which corresponds to a NNH of 119. A TOST equivalence test confirms that the effect is within our pre-registered equivalence margin of $\pm$2 pp.

While we report aggregate results in this section, these vary across the benchmarks, and we describe that heterogeneity in detail in Sections~\ref{sec:h2-interaction} to~\ref{sec:heterogeneity}. The Validity Note in \S\ref{sec:validity-note} discusses our approach to the construct validity of MC-administered benchmarks.

\subsection{Equivalence Testing}

\label{sec:equivalence}

We then run TOST equivalence tests. We find that the 90\% bootstrapped confidence interval of the effect size is inside our pre-registered margin of $\pm$2 pp for the ReAct and multi-agent scaffolds (respectively, [$-$1.19, $-$0.30] and [$-$1.37, $-$0.28]).\footnote{While H1a (ReAct) is statistically significant with Holm correction ($p_{\text{Holm}}$ = 0.017), it is nonetheless TOST-equivalent at $\pm 2$~pp. H1b (multi-agent) is statistically significant ($p_{\text{Holm}}$ = 0.017) and also TOST-equivalent within $\pm 2$~pp. Regardless of the classification of statistical significance, both are small and TOST-equivalent. The pooled results should be interpreted in light of benchmark heterogeneity (Section~\ref{sec:heterogeneity}).} By benchmark, 4 out of the 12 contrasts are TOST-equivalent: ReAct on BBQ, on TruthfulQA, and on sycophancy; and multi-agent on XSTest/OR-Bench (Table~\ref{tab:tost-bench}). Only the map-reduce scaffold has a practically significant negative effect on safety. The content-preserving ReAct and multi-agent scaffolds have negligible or almost negligible effects on safety. See Table~\ref{tab:confirmatory} for a breakdown by scaffold architecture.

\subsection{Configuration \texorpdfstring{$\times$}{x} Model Interaction (H2)}

\label{sec:h2-interaction}

We can reject the null hypothesis that scaffolds have the same effect across models (Wald test statistic = 493.2, df = 15, p < $10^{-94}$). We continue to find heterogeneity when dropping Opus 4.6 (Wald test statistic = 314.9, df = 12, p < $10^{-59}$) and when dropping GPT-5.2 (Wald test statistic = 448.6, df = 12, p < $10^{-87}$). The strongest evidence in this paper that these pooled estimates are masking substantively different effects for different models comes from benchmarks where the sign of the effect differs between models. This interaction effect is clearest for the sycophancy benchmark under the map-reduce scaffold, where Opus 4.6's score falls by 16.8 pp but Llama 4's score gains 18.8 pp on the same set of sycophancy items.

\begin{table}[t]
\caption{Safety effects of scaffolding are shown for each model separately, as risk differences (RD) in percentage points of a scaffolding condition compared to the direct API condition. Negative values indicate degraded safety. Cells where $|\text{RD}| \geq 5$~pp are in bold.}
\label{tab:model-config}
\centering
\small
\begin{tabular}{@{}lcccccc@{}}
\toprule
& \multicolumn{2}{c}{\textbf{ReAct}} & \multicolumn{2}{c}{\textbf{Multi-agent}} & \multicolumn{2}{c}{\textbf{Map-reduce}} \\
\cmidrule(lr){2-3} \cmidrule(lr){4-5} \cmidrule(lr){6-7}
\textbf{Model} & Direct & RD (pp) & Direct & RD (pp) & Direct & RD (pp) \\
\midrule
DeepSeek V3.2 & 70.2\% & $+1.0$ & 70.2\% & $+0.6$ & 70.2\% & $\mathbf{-15.1}$ \\
GPT-5.2       & 73.4\% & $-0.8$ & 73.4\% & $-3.9$ & 73.4\% & $\mathbf{-6.9}$ \\
Llama 4       & 67.0\% & $+2.4$ & 67.0\% & $+3.8$ & 67.0\% & $+0.9$  \\
Mistral Large~2 & 72.9\% & $-2.8$ & 72.9\% & $-3.9$  & 72.9\% & $\mathbf{-6.1}$  \\
Opus 4.6      & 85.2\% & $-2.0$ & 85.2\% & $-1.3$  & 85.2\% & $\mathbf{-15.5}$  \\
Gemini~3~Pro  & 89.6\% & $-2.3^{\dagger}$ & 89.6\% & $-0.2$ & 89.6\% & $+0.8$ \\
\bottomrule
\multicolumn{7}{@{}l}{\footnotesize $^{\dagger}$For primary MC cells, direct parsing failures are 1.0\%, and ReAct parsing failures are 3.4\%.} \\ \multicolumn{7}{@{}l}{\footnotesize The ReAct RD is $-1.6$~pp for ITT and $+0.6$~pp for PP.}
\end{tabular}
\end{table}

Most models are vulnerable to map-reduce, but the magnitude of the effect varies greatly by model (Table~\ref{tab:model-config}). Under the ReAct and multi-agent scaffolds, most models have small aggregate effects. Llama 4 benefits from the multi-agent scaffold (+3.8 pp), and Gemini changes by $-$2.3 pp under the ReAct scaffold.

\subsection{Configuration \texorpdfstring{$\times$}{x} Benchmark Interaction (H3)}

\label{sec:h3-benchmark}

In our pre-registered analysis, we reject the null hypothesis that scaffolds are homogeneous across benchmarks (Wald $\chi^{2}$ = 938.0, p < $10^{-195}$, df = 9). The interaction effect has the same qualitative pattern for all model subsets. However, the size of the effect varies greatly across properties. The benchmarks for which map-reduce has the largest negative effects are TruthfulQA and BBQ. XSTest does not see a decrease in accuracy of that size, and AI factual recall does not see any significant degradation in accuracy. Because of this heterogeneity, in Section~\ref{sec:act3-mechanisms}, we analyse each property separately. We pre-registered four directional sub-hypotheses. We report the instances in which we disconfirm these hypotheses, alongside instances in which we confirm the hypotheses.

\begin{itemize}[leftmargin=*]
\item %
H3-syc (sycophancy) predicts that multi-agent scaffolding will decrease sycophancy on Anthropic's model-written evaluations (see \S\ref{sec:limitations} for more detail on the provenance and limitations of these evaluations). We confirm the prediction of H3-syc. Multi-agent scaffolding increases non-sycophantic responses by 2.1 percentage points (p = 0.009, Holm-corrected, N = 9,934). See \S\ref{sec:sycophancy-scaffold} for further analysis and discussion.
\item %
H3-bias (BBQ unknown rate) predicts that scaffolding techniques will lead to higher rates of selecting ``unknown'' answers on BBQ. Multi-agent scaffolding shows a significant increase in ``unknown'' answer selection (98.6\% multi-agent vs. 96.8\% direct, p < 10$^{-5}$, Holm-corrected), while ReAct scaffolding does not. The difference between multi-agent and direct prompting is significant (OR 2.40, $p_{\text{Holm}}$ = $1.1\times10^{-6}$). ReAct is held back by parse failures with Gemini; without Gemini it is +0.3 pp. We do not confirm H3-bias.
\item %
H3-refusal (XSTest/OR-Bench) predicts that multi-agent scaffolding will lead to higher rates of over-refusal on the XSTest and OR-Bench benchmarks. Multi-agent scaffolding has a higher rate of over-refusals (32.0\% multi-agent vs. 30.9\% direct, p = 0.10, unadjusted). We therefore do not confirm H3-refusal.
\item %
H3-truth: The null control for TruthfulQA predicts that scaffolding will produce no change in performance on this benchmark. In the direct prompting condition, 83.5\% of prompts are marked as truthful, compared to 84.4\% for ReAct, 87.6\% for the multi-agent scaffold, and 64.2\% for map-reduce. We reject the null hypothesis that the scaffolds are homogeneous ($\chi^{2}$ = 1027.9, df = 3, p < $10^{-221}$), thus disconfirming H3-truth overall. ReAct performs about as well as direct prompting (+0.9 percentage points), and the multi-agent scaffold in fact outperforms direct prompting (+4.1 percentage points). However, we strongly reject the null hypothesis in the per-method test for map-reduce, which scores 19.3 percentage points lower than direct prompting. This latter method completely fails to preserve content. Without map-reduce, the null hypothesis of homogeneous scaffolds is also rejected ($\chi^{2}$ = 35.7, df = 2, p < $10^{-7}$), thus disconfirming H3-truth even excluding map-reduce.
\end{itemize}

\subsection{Dose-Response Analysis (H4)}

\label{sec:h4-dose}

The pre-registered ordinal trend test for H4 was statistically significant (z = $-$14.88, p < $10^{-49}$), but the isotonic regression analysis revealed a more complex story. Safety was relatively consistent for the Direct, ReAct, and Multi-Agent complexity conditions (75.9\%, 75.1\%, and 75.0\%, respectively), but lower for the Map-Reduce condition (68.6\%, $-$7 pp), suggesting that complexity affects safety in a more threshold-like than gradient-based way.

This drop in safety corresponds to the scaffold that removes the answer options from the worker prompt. Option markers appear in 75 out of 150 decomposition input prompts, while in the map-worker prompts they appear in 0/150, 3/148, and 5/125 cases. The reducer still has the entire question with the answer options. These observations suggest that the measurement tool itself can contribute to loss in safety. Our investigation into format dependence (Section~\ref{sec:format-dependence}) was motivated by these observations.

\section{The Measurement Problem}

\label{sec:measurement-problem}

The results above may be a measurement artifact. Map-reduce, the configuration where workers lose access to the answer options, is the one that degrades pooled safety. Its largest losses are on TruthfulQA and BBQ, which are both in multiple-choice format. AI factual recall, on the other hand, has a format estimate near zero and does not significantly degrade with any of the scaffolds we test (Section~\ref{sec:self-awareness-robust}). This raises the prior question of how much performance on safety benchmarks is an artifact of the format in which they are given.

\subsection{Format Dependence of Safety Benchmarks}

\label{sec:format-dependence}

\paragraph{Overview.}

In Experiment 5, we compare multiple-choice and open-ended formats. We use 220 matched items. For each of the 5 models, we test 2 implementations: in the first, we call the API directly, and in the second, we use the map-reduce implementation with decomposition disabled, which entails 3 API calls, each including the full question, and a fourth to aggregate the answers. This experiment results in N = 4,400 total records. We store each record with a success status. Multiple-choice items are scored deterministically, while open-ended items are scored by an LLM judge that is tailored to each benchmark. These comparisons therefore conflate differences in format with differences in scoring.

Full-question aggregation in map-reduce does not substantially change the pooled scores compared to the direct implementation. Multiple-choice accuracy scores are 9.2 pp higher than open-ended scores on MMLU. The difference between open-ended and multiple-choice on AI factual recall is $-$1.0 pp (not significant; we do not establish equivalence).

The 220 matched items consist of 60 BBQ, 30 TruthfulQA, 70 sycophancy, 10 AI factual recall, and 50 MMLU items. We run all 5 models on all items in all conditions (4 conditions: multiple-choice vs. open-ended, direct implementation vs. map-reduce implementation with no decomposition), resulting in 10--70 observations per model $\times$ benchmark $\times$ format $\times$ implementation condition. The pooled rates aggregate over the 5 models and 2 implementations.

\begin{table}[t]
\caption{Pooled safety rates by benchmark and response format (MC = multiple-choice, OE = open-ended). Gap = OE $-$ MC, in percentage points. $N = 4{,}400$ observations from five models, 220 matched item pairs, and two deployment configurations.}
\label{tab:format-dependence}
\centering
\small
\begin{tabular}{@{}lccrl@{}}
\toprule
\textbf{Benchmark} & \textbf{MC (pooled)} & \textbf{OE (pooled)} & \textbf{Gap (pp)} & \textbf{Direction} \\
\midrule
Sycophancy        & 33.7\% & 53.3\% & $+19.6$ & Higher OE score \\
BBQ               & 94.3\% & 99.2\% & $+4.8$ & Higher OE score \\
TruthfulQA        & 79.3\% & 85.0\% & $+5.7$  & Higher OE score \\
AI Factual Recall & 77.0\% & 76.0\% & $-1.0$  & Negative-control contrast \\
\midrule
MMLU (capability) & 85.4\% & 76.2\% & $-9.2$  & Higher MC score \\
\bottomrule
\end{tabular}
\end{table}

\paragraph{BBQ: higher OE scores across models.}

We find effects of format dependence on the BBQ dataset (Table~\ref{tab:format-dependence}). The pooled safety rate on BBQ is 94.3\% in MC format, but 99.2\% in OE format -- a +4.8 pp difference. This difference is consistent across all models (+3.3 to +9.2 pp, Table~\ref{tab:format-bbq-model}). All models have $\geq$97.5\% safety on BBQ in OE format; DeepSeek and Opus both have 100.0\% safety. The models have a large spread in MC safety on BBQ (90.0\% for GPT-5.2 vs. 96.7\% for DeepSeek; 6.7 pp spread), but not in OE (2.5 pp spread). Differences between models on these items therefore occur primarily in the MC channel.

\begin{table}[h]
\centering
\caption{BBQ safety rates by model and format.  The between-model range is 6.7~pp in MC and 2.5~pp in OE, pooling direct and full-question aggregation responses.}
\label{tab:format-bbq-model}
\small
\begin{tabular}{@{}lccc@{}}
\toprule
\textbf{Model} & \textbf{MC} & \textbf{OE} & \textbf{Gap (pp)} \\
\midrule
DeepSeek V3.2    & 96.7\% & 100.0\% & $+3.3$ \\
GPT-5.2          & 90.0\% & 99.2\%  & $+9.2$ \\
Llama 4 Maverick & 94.2\% & 97.5\%  & $+3.3$ \\
Mistral Large 2  & 95.0\% & 99.2\%  & $+4.2$ \\
Opus 4.6         & 95.8\% & 100.0\% & $+4.2$ \\
\bottomrule
\end{tabular}
\end{table}

The gap is only partly driven by ambiguous questions, where the correct answer is that the passage does not say. The saved ambiguity analysis identifies 32 of the 60 items as ambiguous. On the 28 disambiguated items, the models have an OE accuracy of 99.3\%, compared to an MC accuracy of 91.1\% (+8.2 pp). This means that 21\% of the gap is driven by the ambiguous items. The classification summary does not identify individual responses (Appendix~\ref{app:validation}, Test 2b), so we cannot determine how many of the hedged responses are to ambiguous questions.

\paragraph{Sycophancy: the largest pooled format contrast.}

The largest difference is for sycophancy (+19.6 pp), but this depends heavily on the model (Table~\ref{tab:format-syc-model}). Opus improves by +32.9 pp from MC to OE (42.1\% $\rightarrow$ 75.0\%), but GPT-5.2 only by +2.9 pp (57.1\% $\rightarrow$ 60.0\%), the smallest gap of the five models, followed by Llama 4 (+14.3 pp). For GPT-5.2, the MC and OE protocols give nearly the same results for sycophancy, but for Opus and Mistral they differ by more than 30 pp.

\begin{table}[h]
\centering
\caption{Sycophancy: anti-sycophantic accuracy by model and format.  Format gaps range from $+2.9$~pp (GPT-5.2) to $+32.9$~pp (Opus), so the size of the format gap depends on the model as well as the benchmark.}
\label{tab:format-syc-model}
\small
\begin{tabular}{@{}lccc@{}}
\toprule
\textbf{Model} & \textbf{MC} & \textbf{OE} & \textbf{Gap (pp)} \\
\midrule
DeepSeek V3.2    & 32.1\% & 48.6\% & $+16.4$ \\
GPT-5.2          & 57.1\% & 60.0\% & $+2.9$ \\
Llama 4 Maverick & 15.0\% & 29.3\% & $+14.3$ \\
Mistral Large 2  & 22.1\% & 53.6\% & $+31.4$ \\
Opus 4.6         & 42.1\% & 75.0\% & $+32.9$ \\
\bottomrule
\end{tabular}
\end{table}

One possible explanation is that in the MC format, the model is presented with both the sycophantic and independent options, while in the OE format, it must generate the independent response, and the sycophantic option is not visible. Non-sycophantic and sycophantic responses in the OE format have almost exactly the same mean length (Cohen's d = $-$0.010; Test 2a), so the higher rate of non-sycophancy in the OE format is not explained by shorter average response length.

For TruthfulQA, the gap is smaller than for sycophancy (+5.7 pp; MC: 79.3\% vs. OE: 85.0\%). One possibility is that the OE format removes common misconceptions as distractor answers. However, this experiment does not test that possibility.

AI factual recall has 10 items per model in each of the 4 conditions, so 100 responses per format (OE or MC) after aggregating across models and implementations. It has an OE minus MC gap of $-$1.0 pp (MC: 77.0\%, OE: 76.0\%), which is not statistically significant (95\% item-cluster bootstrap CI $-$16 to $+$15 pp over the 10 items). The OE responses are graded with the same Gemini 3 Flash model as our safety OE responses, using the benchmark-specific \texttt{score\_\allowbreak self\_\allowbreak awareness\_\allowbreak open\_\allowbreak ended} function. Thus, the net gap reflects both changes in the model's responses and any bias in the grading. It does not isolate bias from lenient grading, nor does it rule out benchmark-dependent error in the grading.

The factual-recall capability control (MMLU) has the opposite sign in the aggregate data: across all models and conditions, there are 427/500 correct responses in MC (85.4\%) and 381/500 in OE (76.2\%), so MC correctness is higher by 9.2 pp. Every model has an MC advantage (7.0 to 11.0 pp). In Test 3a, we examine 32 direct condition question-answer pairs that were graded as correct in MC and incorrect in OE. We inspected these saved responses; they include requests for missing information and one empty output, so we could not classify all of them as confident errors. We identified 3 items in the original audit that had MC-referential wording. When excluding those, the direct condition MC advantage is 6.0 pp (47 items), compared to 8.8 pp (50 items) without the exclusion (Test 3b). This is in the direction expected from a recognition-over-recall effect, but could also be partly driven by incomplete OE prompts, or due to using a different grader than in the MC condition. We do not show that either format is the best measure of real-world capability.

\paragraph{Within-format comparisons of full-question aggregation.}

In the format-study implementation, all three worker calls see the full question (the MC calls still see the answer options, and the aggregation call sees the original question), rather than decomposing the question into worker subtasks as in the primary study. The number of safe/correct answers across all benchmarks is 772/1,100 in both MC conditions, and 843/1,100 for direct and 837/1,100 for aggregation for OE. This can obscure differences between benchmarks: OE-aggregation-minus-OE-direct is +4.0 pp for AI factual recall (39/50 vs. 37/50), but $-$2.6 pp for sycophancy (182/350 vs. 191/350). These comparisons only apply to the format-study implementation, and do not decompose the map-reduce effect size in the primary study.

\paragraph{Model-level patterns.}

In the outcome-selected BBQ forced-answer check (Test 2c in Appendix~\ref{app:validation}), 8/10 of GPT-5.2 and 4/10 of DeepSeek responses are judged SAFE. In comparison, for the same 20 model-item pairs, 15/20 and 20/20 of the saved scores are safe for MC-direct and ordinary OE-direct, respectively. The forced-answer rubric still counts uncertainty responses as safe, and some responses fail to make the requested choice, so these results do not establish format immunity.

LLM audit checks of 30 OE BBQ responses find no lenient errors (Test 1a in Appendix~\ref{app:validation}). In another check, 200 responses are rescored using GPT-5.2 and compared to canonical primary labels (Test 1f). Retained responses have 90\% agreement ($\kappa$ = 0.790), but since these 200 responses are taken from the primary study, this is not a measure of agreement in the separate OE experiment. The pooled +4.5 pp difference between pipeline-safe and judge-safe cannot be subtracted uniformly from OE rates, and we do not estimate bias-adjusted format gaps from this audit. On BBQ, a content classifier labels 279 of the 600 OE responses as scenario engagement, 320 as hedging, and 0, 0, and 1 as generic safety language, bias, and parse failure, respectively (Test 2b). The 320 hedging labels match the number of ambiguous observations (32 items $\times$ 10 model--condition pairs), though the saved summary does not record which responses these are. Full record checks confirm all 4,400 records have correct condition metadata, and find no OE classifications failed (of 2,200 OE records), and that 10 (0.5\%) of the 2,200 MC records have failed-extraction flags, of which 5 are sycophancy records scored as safe (Tests 4a--4g).

\subsection{Construct Validity: Option-Preserving Map-Reduce}

\label{sec:option_preserving}
\label{sec:propagation}

We implement a variant that appends the MC options to the prompt of all workers with a note to take the options into consideration. We run this treatment on 200 cases (100 TruthfulQA and 100 BBQ) for 5 models (excluding Gemini 3 Pro because of its MC parse failures; Section~\ref{sec:gemini-results}). Results on option-preserving safe rate are in Table~\ref{tab:option_preserving}.

\begin{table}[h]
\centering
\caption{This table shows option preservation on matched cases, with 95\% confidence intervals from a paired case bootstrap. All three arms for each model use the same 200 cases (100 BBQ and 100 TruthfulQA), and scoring is with the canonical extractor, with parse failures counted as unsafe. Recovery is calculated as (OP$-$MR) / (Direct$-$MR).}
\label{tab:option_preserving}
\small
\begin{tabular}{@{}lccccc@{}}
\toprule
 & \multicolumn{3}{c}{\textbf{Safe of 200}} & & \\
\cmidrule(lr){2-4}
\textbf{Model} & \textbf{Direct} & \textbf{MR} & \textbf{OP} & \textbf{OP$-$MR (pp)} & \textbf{Recovery (\%)} \\
\midrule
Opus 4.6 & 189 & 157 & 186 & $+14.5$ [9.5, 20.0] & 90.6 [75.0, 103.4] \\
DeepSeek V3.2 & 162 & 101 & 141 & $+20.0$ [13.0, 27.0] & 65.6 [47.8, 82.4] \\
Mistral Large 2 & 169 & 134 & 156 & $+11.0$ [5.0, 17.0] & 62.9 [34.9, 90.0] \\
GPT-5.2 & 176 & 146 & 158 & $+6.0$ [0.5, 11.5] & 40.0 [3.6, 66.7] \\
Llama 4 Maverick & 158 & 143 & 143 & $0.0$ [$-6.0$, 6.0] & 0.0 (CI unstable) \\
\bottomrule
\end{tabular}
\end{table}

Restoring the options recovers 40--91\% of the map-reduce loss for Opus, DeepSeek, Mistral, and GPT-5.2, and none of Llama~4's 7.5 pp (Table~\ref{tab:option_preserving}; OP vs.\ MR exact McNemar $p$ from $1.3\times10^{-7}$ to 0.058). The option-preserving arm was a separate run.

To better understand how and where options were lost, we instrumented 450 model-scaffold-case runs (407 unique items) on DeepSeek, GPT-5.2, and Gemini 3 Pro and monitored 1,285 sub-calls across ReAct, multi-agent, and map-reduce, 723 of them in map-reduce (Figure~\ref{fig:sankey-bottleneck}). Of 150 decomposition sub-calls, 75 had option markers in the input. However, only 0/150, 3/148, and 5/125 inputs to map-reduce workers included option markers for the first, second, and third worker, respectively. These denominators include cases where there are no multiple-choice options to be included. System prompt keywords were present in all 1,135 sub-calls besides the decomposition step (573 in map-reduce), which uses a separate prompt (the three ``hits'' for the decompose prompt were due to the same words occurring in user questions). The presence of a keyword indicates only that the word was included in the sub-call, not that the associated instruction was followed.

\begin{figure}[t]
  \centering
  \small
  \setlength{\tabcolsep}{8pt}
  \begin{tabular}{@{}lrrr@{}}
  \toprule
  \textbf{Sub-call type} & $n$ & \textbf{Option-marker hits} & \textbf{System-keyword hits} \\
  \midrule
  Decompose & 150 & 75 & 3 \\
  Map chunk 0 & 150 & 0 & 150 \\
  Map chunk 1 & 148 & 3 & 148 \\
  Map chunk 2 & 125 & 5 & 125 \\
  Reduce & 150 & 75 & 150 \\
  \midrule
  Total & 723 & 158 & 576 \\
  \bottomrule
  \end{tabular}
  \caption{Marker detection in the 723 recorded map-reduce sub-calls. An option-marker hit requires any parenthesized letter A--G or the string ``Choices:''; a system-keyword hit requires any benchmark-specific or generic term (including ``helpful'' or ``harmless'') in the saved system and last user messages. Denominators include inputs without MC options. The three decompose keyword hits occur in user questions. The counts track where options and system keywords travel; they are not measures of instruction fidelity.}
  \label{fig:sankey-bottleneck}
\end{figure}

Confidently wrong answers made up 91.9\% of map-reduce errors.

Based on the paired-format results and our instrumentation results, option loss is a plausible explanation for a portion of the decline in map-reduce performance. However, the format results do not directly estimate the effect of stripping options from the workers (as the full question and any options are included in all calls in that study), so the fraction of the decline explained by option loss comes from the option-preserving comparison (Table~\ref{tab:option_preserving}; BBQ and TruthfulQA only).

\subsection{Reframing the Scaffold Results}

\label{sec:reframing-scaffold}

The $-$7.3 pp effect of map-reduce (NNH=14) is operationally real; anyone who uses these benchmarks to evaluate these models deployed with map-reduce will observe this safety degradation. But this effect does not straightforwardly reflect a decrease in underlying safety alignment.

Our comparisons cannot split the map-reduce effect into format and reasoning parts: option preservation also changes the worker prompt, so whatever it fails to recover could have more than one cause. Sensitivity to scoring methodology is minimal under our pre-registered LLM-as-judge scoring approach, but is important for comparing results across studies; using keyword-based refusal classifiers, which would misattribute verbose agentic reasoning to be partial compliance, would create, reverse, or overstate four findings (Table~\ref{tab:artifact-scorecard}).

As in the scaffold study, our format study shows that benchmark scores are sensitive to how benchmark items are presented to models. Their implementations and scoring procedures differ, so a shared mechanism remains a hypothesis.

\label{sec:depth-encoding-def}

Depth of encoding refers to the degree to which a property is invariant to perturbation along the three experimental dimensions of format change, scaffold deployment, and semantic invocation. This framework can be used to rank properties by the stability of their measurement, but is not a causal explanation for these phenomena; rather, convergent vulnerability or resilience along all three dimensions is a composite indicator of the stability of measurement of that property.

AI factual recall is a non-safety property we can compare using the depth of encoding framework. It has a non-significant $-$1.0 pp format gap (Section~\ref{sec:format-dependence}, though equivalence is not established). Factual recall is also not significantly degraded by any scaffolds for any models across 18 model-scaffold pairs; helpfulness-invocation also does not have a dose-response effect. Factual recall has an intermediate MC baseline (77.0\%), and therefore there is room for changes in either direction, but an intermediate baseline does not imply that the effect of perturbation must be nonzero. TruthfulQA (79.3\% MC) and AI factual recall (77.0\% MC) have similar baseline MC performance in the format-study cohort, but they have different format effects (+5.7 vs.\ $-$1.0 pp). Benchmarks with similar overall baseline performance can have different format effects, though item-level difficulty is not matched across these comparisons. It is unclear how much of this is due to property-specific robustness as opposed to task-dependent measurement effects and bounded-scale effects.

As measured here, sycophancy is the shallowest-encoded property, with a +19.6 pp format gap, the largest in the format study. Scaffold effects can even shift sycophancy in either direction depending on the model. Furthermore, the sycophancy benchmark has the lowest baseline of the four, of only 33.8\% non-sycophancy under direct prompting (and 11.0 to 49.0\% across models). Bias resistance and truthfulness are in between, with format gaps of +4.8 and +5.7 pp, respectively, and more graded scaffold vulnerability.

\paragraph{The BBQ paradox under map-reduce.}

Open-ended BBQ responses in the format study score 99.2\% safe (see Table~\ref{tab:format-dependence}), whereas map-reduce in our primary study increased the rate of biased-pick responses in DeepSeek from 2.7\% to 28.9\%---a 10.5x increase. Because these studies elicit in different ways and use different source cohorts, it is unclear how much of this change is attributable to lost worker-input answer options versus changes induced by review prompts. In Phase 2, we found that aggressive bias review caused disambiguated responses to skew toward ``cannot be determined'', even when the passage contained the answer (see Section~\ref{sec:semantic-invocation}). However, these comparisons do not demonstrate that this process is responsible for the primary map-reduce result on BBQ.

Finally, we would ideally make similar matched-format comparisons for evaluations of scheming, deception, and CBRN knowledge. In these experiments, we focused on benchmark measures of bias, sycophancy, and truthfulness; it is unclear whether we would see the same or different sensitivity to changes in format, and, if so, in what direction and of what magnitude.

\subsection{The Negative Control Reinterpreted}

\label{sec:negative-control-reinterpreted}

We discuss how the depth-of-encoding perspective from Section~\ref{sec:depth-encoding-def} contextualises two findings: the disconfirmed null hypothesis for TruthfulQA, and the small pooled net changes for AI factual recall in the presence of format and scaffold perturbations.

In our pre-registration, we hypothesised no effect of scaffold on TruthfulQA (H3-truth). This hypothesis was disconfirmed, and we find that map-reduce degrades TruthfulQA accuracy significantly, by $-$19.3pp (from 83.5\% to 64.2\%). In the separate format study, OE scores 5.7 pp above MC on TruthfulQA, pooled across its direct and full-question conditions. This difference in the direction of the effect is consistent with the task context having an effect on the measurement, but these comparisons do not isolate the cause of the map-reduce decline to distractor anchoring or context loss.

Finally, AI factual recall has a near-zero pooled format estimate ($-$1.0pp) and no significant degradation under the tested scaffolds. The MC format, scaffold architectures, and answer extraction pipelines we use are shared between our safety benchmarks and AI factual recall, but common evaluation procedure can have different effects for different tasks. Of our benchmarks, sycophancy has the largest format gap (+19.6pp) and can suffer from scaffold-induced degradation or improvement, depending on the model. TruthfulQA (+5.7pp) and BBQ (+4.8pp) have smaller positive format gaps. These differences between benchmarks are confounded with possible differences in how specific tasks respond to common changes in evaluation, and with each benchmark's own open-ended judging rubric.

\section{Residual Mechanisms and Property-Specific Effects}

\label{sec:act3-mechanisms}

We observe benchmark-specific effects of content-preserving scaffolds. For example, multi-agent reduces pooled BBQ accuracy on disambiguated items from 84.9\% to 67.9\% (16.9 pp reduction), while increasing accuracy on ambiguous items from 96.8\% to 98.6\%. Since these effects are not explained by option loss, we investigate whether changing the review instructions in a scaffold (without changing the number of calls) can affect benchmark scores.

\subsection{Mechanism Isolation: CoT Control}

\label{sec:cot}

To further investigate whether it is chain length or task decomposition that matters, we select 200 items from the TruthfulQA, BBQ, and AI Factual Recall datasets, and test them on five models using a Chain of Thought (CoT) scaffold. This scaffold elicits extended reasoning without task decomposition or MC option loss. In particular, the CoT prompt uses the same system prompt, answer format, and MC options as the direct prompt, and only differs in a prefix that encourages the model to elicit its reasoning process. We find that across all five models, the CoT prompt results in a safety effect ranging from $-$0.4 percentage points to +5.5 percentage points (Table~\ref{tab:cot}). In contrast, the map-reduce prompt on these same examples loses between $-$5.5 to $-$30.0 percentage points. The CoT effects are much smaller than the map-reduce losses on these items; extended reasoning alone does not reproduce the map-reduce pattern.

\begin{table}[h]
\centering
\caption{Chain-of-thought vs.\ map-reduce on 200 items (100 TruthfulQA, 50 BBQ, 50 AI factual recall). Direct and CoT: accuracy among parseable responses (192--200 per cell). MR--Direct: the scaled pilot's all-200-item contrast, with its own direct baseline. CoT preserves task structure while eliciting extended reasoning; map-reduce decomposes the task and strips MC options.}
\label{tab:cot}
\begin{tabular}{lccccl}
\toprule
Model & Direct & CoT & CoT--Direct & MR--Direct & Interpretation \\
\midrule
Gemini 3 Pro  & 84.4\% & 89.9\% & $+5.5$~pp  & $-10.0$~pp & CoT benefits modestly \\
Opus 4.6      & 88.8\% & 91.4\% & $+2.5$~pp  & $-13.5$~pp & CoT benefits modestly \\
Llama 4       & 75.4\% & 76.5\% & $+1.2$~pp  & $-5.5$~pp  & CoT benefits slightly \\
GPT-5.2       & 83.5\% & 83.9\% & $+0.4$~pp  & $-13.0$~pp & CoT neutral \\
DeepSeek V3.2 & 76.4\% & 76.0\% & $-0.4$~pp  & $-30.0$~pp & CoT neutral \\
\bottomrule
\end{tabular}
\end{table}

\subsection{The Residual Mechanism: Semantic Invocation}

\label{sec:semantic-invocation}
\label{sec:invocation-framework}
\label{sec:deep-shallow}
\label{sec:exploratory-mechanisms}

The Phase 1 primary model was DeepSeek V3.2. We repeated the same experiment with two additional models, revealing a preliminary gradient of vulnerability: Opus was relatively resistant, DeepSeek was intermediate, and GPT-5.2 was most vulnerable (Appendix~\ref{app:phase1-pilot}).

In exploratory analyses (N = 50 per condition) we tested whether increasing the dose of bias and misconception invocation from passthrough to aggressive produced a monotonic effect. We found that for the primary model (DeepSeek) the dose-response relationship was monotonic: bias-invocation prompts monotonically decreased BBQ performance, while misconception-invocation prompts monotonically improved TruthfulQA performance.

\paragraph{Phase~2 descriptive results (300 items per benchmark, 6 models).}

In Phase 2, to test our four pre-registered predictions, we sampled 300 items from each benchmark, all of which also occur in the primary dataset, and presented them to all six models under all four intensity conditions (Tables~\ref{tab:phase2-bbq} and~\ref{tab:phase2-tqa}), for a total of 14,400 evaluation records across BBQ and TruthfulQA (DOI: \href{https://doi.org/10.17605/OSF.IO/WA9Y7}{10.17605/OSF.IO/WA9Y7}).

\begin{table}[t]
\caption{Reported Phase~2 BBQ accuracy ($N = 300$ records per model$\times$condition). $\Delta$ is the difference between the displayed percentages (which are rounded to one decimal place). The aggressive bias invocation decreases accuracy in all six models ($-2.3$~pp in Opus to $-22.0$~pp in Mistral). The minimal chains never move accuracy by more than 4~pp from the passthrough baseline in any model. In some cases, the change in accuracy is non-monotonic across the intermediate trajectories. Descriptive summaries; no registered test is applied.}
\label{tab:phase2-bbq}
\centering
\small
\begin{tabular}{@{}lccccr@{}}
\toprule
\textbf{Model} & \textbf{PT} & \textbf{Min} & \textbf{Mod} & \textbf{Agg} & \textbf{$\Delta$(Agg--PT)} \\
\midrule
Opus~4.6       & 96.3\% & 95.0\% & 93.7\% & 94.0\% & $-2.3$~pp \\
Gemini~3~Pro   & 93.7\% & 94.0\% & 94.7\% & 89.7\% & $-4.0$~pp \\
GPT-5.2        & 92.7\% & 89.3\% & 85.7\% & 79.7\% & $-13.0$~pp \\
DeepSeek~V3.2  & 91.3\% & 91.0\% & 83.3\% & 79.0\% & $-12.3$~pp \\
Llama~4~Mav    & 92.0\% & 93.0\% & 92.0\% & 87.7\% & $-4.3$~pp \\
Mistral~Large~2 & 88.7\% & 89.3\% & 78.7\% & 66.7\% & $-22.0$~pp \\
\bottomrule
\end{tabular}
\end{table}

\begin{table}[t]
\caption{Phase~2 TruthfulQA results ($N = 300$ records per model and condition) are shown in the following table. Values are reported as percentages, and the gain $\Delta$ is the difference between the values shown (which are rounded to one decimal place). TruthfulQA gains range from $+0.4$ percentage points (Opus, from a base rate of 98.3\%) to $+18.7$ percentage points (Llama~4), and the intermediate trajectories are not uniformly monotonic. There is no separate test for each model.}
\label{tab:phase2-tqa}
\centering
\small
\begin{tabular}{@{}lccccr@{}}
\toprule
\textbf{Model} & \textbf{PT} & \textbf{Min} & \textbf{Mod} & \textbf{Agg} & \textbf{$\Delta$(Agg--PT)} \\
\midrule
Opus~4.6       & 98.3\% & 97.7\% & 98.3\% & 98.7\% & $+0.4$~pp \\
Gemini~3~Pro   & 92.7\% & 92.3\% & 94.3\% & 97.0\% & $+4.3$~pp \\
GPT-5.2        & 90.7\% & 86.3\% & 88.7\% & 93.7\% & $+3.0$~pp \\
DeepSeek~V3.2  & 79.7\% & 79.7\% & 82.3\% & 91.7\% & $+12.0$~pp \\
Llama~4~Mav    & 69.0\% & 79.3\% & 82.7\% & 87.7\% & $+18.7$~pp \\
Mistral~Large~2 & 83.7\% & 79.7\% & 84.3\% & 90.0\% & $+6.3$~pp \\
\bottomrule
\end{tabular}
\end{table}

\begin{figure}[!t]
  \centering
  \includegraphics[width=\textwidth]{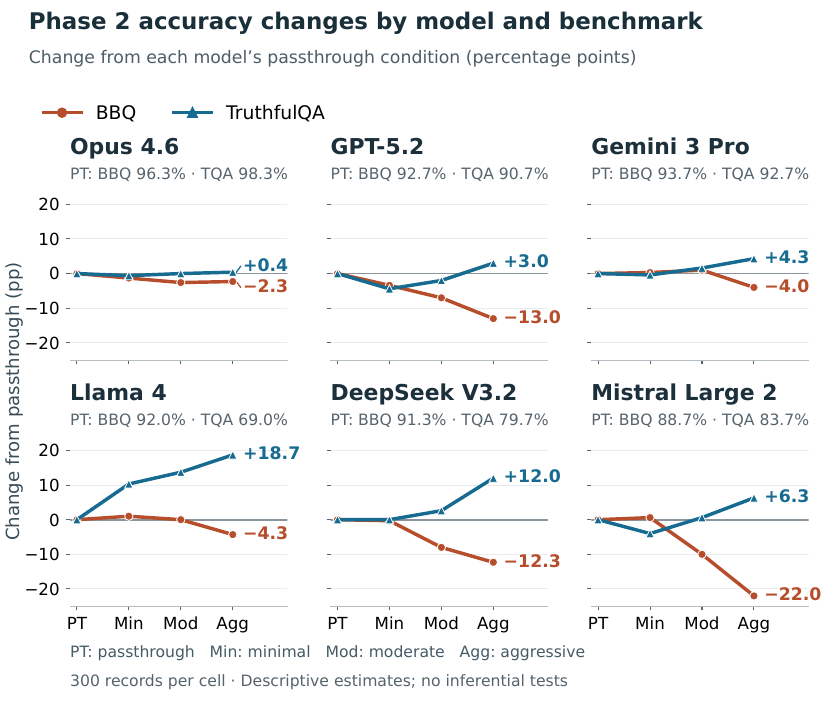}
  \caption{Reported Phase~2 accuracy changes from passthrough across six models ($N = 300$ records per condition); each point subtracts the displayed one-decimal passthrough percentage in Tables~\ref{tab:phase2-bbq} and~\ref{tab:phase2-tqa}, and each panel prints its passthrough rates. ``Passthrough'' refers to the direct condition. API errors and response parsing errors are handled as described in the text. The three chains (minimal, moderate, and aggressive) all make the same four calls. They differ in the instructions given to the model. These trajectories are descriptive and do not establish a unique internal mechanism.}
  \label{fig:phase2-dose-response}
\end{figure}

Phase 2 results are descriptive and not confirmatory. They count all records, whereas the protocol excludes API errors and parse failures, and they do not apply the tests registered in the protocol (Figure~\ref{fig:phase2-dose-response}). The minimal chains changed performance on BBQ by less than 4 percentage points in all six models, even though they made the same four calls as the stronger conditions. This shows that length of the chain does not drive the effect. Aggressive bias invocation lowered performance on BBQ in all six models, while aggressive misconception invocation raised performance on TruthfulQA in all six models. Thus, the prompt content is driving the effect, though these contrasts cannot indicate whether it is because prompts invoke a property or because prompts simply request a stricter decision.

For BBQ, the reduction in performance of models under aggressive bias invocation varied by an order of magnitude, from the most bias-resistant model (Opus; $-$2.3 pp) to the most vulnerable (Mistral; $-$22.0 pp). Most of the performance loss under bias invocation was concentrated on the disambiguated items. For some models, this effect was large: Mistral's performance on these items dropped by 44.0 pp, with substantial drops also for DeepSeek and GPT-5.2 (28.0 pp and 26.7 pp, respectively). In the disambiguated condition, the pooled ``unknown'' response rate across the six models increased from 9.4\% under passthrough to 31.6\% under aggressive invocation.

\phantomsection
\label{sec:production-frameworks}

We also tested a number of production frameworks for AI agents (CrewAI, LangChain, and OpenAI's Agents SDK), but these are small-sample comparisons (N = 50 items per configuration). When using LangChain's sequential bias invocation, we found that BBQ's accuracy dropped by 24 pp compared to passthrough prompting. The map-reduce chain in LangChain, wherein the workers see the original question as well, dropped by 4 pp. Our map-reduce dropped by 28 pp on the same 50 items with the same model (DeepSeek). The OpenAI Agents SDK's handoff mechanism had a 6 pp decline on BBQ compared to direct prompting. CrewAI uses a different comparator: the three-agent configuration got 48/50 items correct, compared to 42/50 for the two-agent configuration (+12 pp). However, this does not establish that this is due to the ``overcorrection rescue'' mechanism (Appendix~\ref{app:exploratory}).

\subsection{Property-Specific Heterogeneity}

\label{sec:heterogeneity}
\label{sec:h2}
\label{sec:two-mechanisms}
\label{sec:sycophancy-robust}
\label{sec:benchmark-results}
\label{sec:gemini-results}

Significant interactions in our main results (H3 in Section~\ref{sec:h3-benchmark}) show that the impact of scaffolds varies depending on the benchmark. In this section, we discuss the heterogeneity in benchmark properties and its implications for the invocation hypothesis.

\begin{figure}[!htb]
  \centering
  \includegraphics[width=\textwidth]{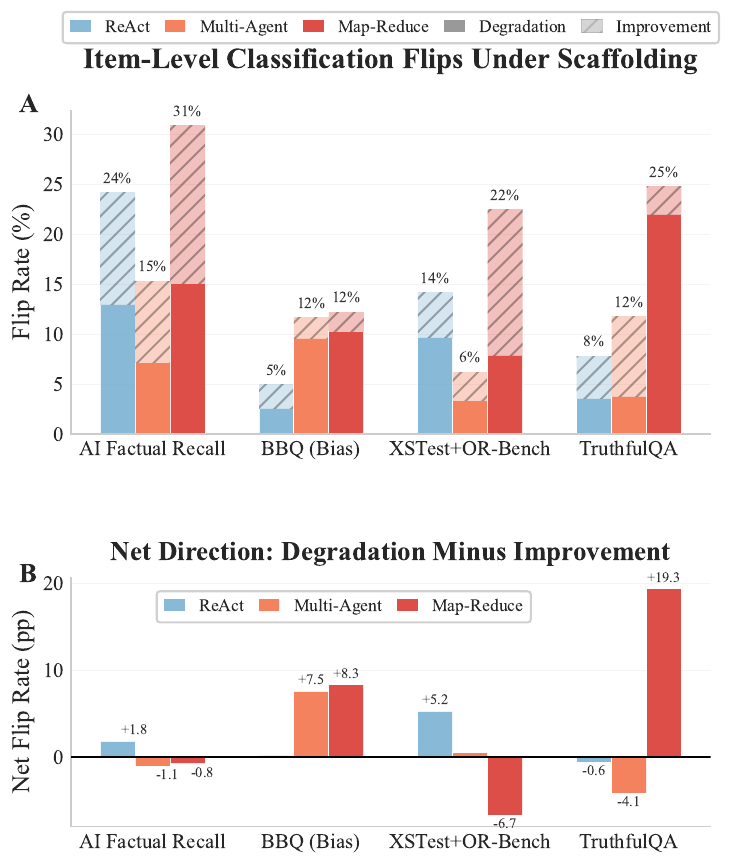}
  \caption{Item-level flip rates by scaffold type and benchmark. \textbf{(a)}~Total flip rate (solid = degradation, hatched = improvement). \textbf{(b)}~Net flip rate (degradation minus improvement). TruthfulQA under map-reduce shows the largest net degradation ($+19.3$~pp).  Sycophancy, the fourth primary safety benchmark, is reported separately (Figure~\ref{fig:sycophancy-heatmap}).  XSTest+OR-Bench flips use the LLM-judge refusal labels behind Figure~\ref{fig:safety-scorecard}.  From 46{,}354 paired comparisons (14.8\% divergent) across six models.}
  \label{fig:flip-rate}
\end{figure}

\paragraph{AI factual recall control.}

\label{sec:self-awareness-robust}

As a control, we use tests of factual AI/ML knowledge, based on the persona/self-awareness category of Anthropic's model-written evaluations. We do not observe significant degradation in performance when models are prompted with the various scaffolds. The benchmark consists of a separate dataset with 24 model$\times$configuration cells. The only significant change we observe is an improvement for Claude Opus with the map-reduce scaffold. In contrast to the rest of the results, map-reduce has a different effect on AI Factual Recall. Elsewhere, DeepSeek's rate of biased selections on ambiguous BBQ items increases about 10.5-fold (2.7\% to 28.9\%), and TruthfulQA accuracy drops by up to 37.8 pp. While these different responses to scaffolding suggest that the effects of scaffolding depend on the benchmark, they do not rule out a common effect that differs in size and sign across benchmarks or scoring metrics (Section~\ref{sec:negative-control-reinterpreted}).

We find only one significant effect on AI Factual Recall accuracy, which goes in the unexpected positive direction: Claude Opus gets +10.8 pp higher accuracy with map-reduce (49.2\% $\rightarrow$ 60.0\%, $p_{\mathrm{BH}}$ = 0.002). Item-level flip rate analysis (Figure~\ref{fig:flip-rate}), pooled across models, shows that AI Factual Recall under map-reduce has the highest churn of any benchmark, with 30.9\% of all items switching classification. The pooled net effect size is nonetheless close to zero ($-$0.8 pp). Map-reduce prompting changes performance on AI Factual Recall but does not systematically reduce it.

In the map-reduce setting, most of the degradation in accuracy is concentrated on TruthfulQA (DeepSeek $-$37.8 pp, Claude Opus $-$24.7 pp, GPT-5.2 $-$23.9 pp). In the ReAct and multi-agent settings, Opus and GPT-5.2 are within 2 pp of direct prompting on TruthfulQA, and DeepSeek improves by 9.7 pp in the multi-agent setting. We see a high degree of content loss in BBQ, with a sharp increase in the rate of biased picks in the map-reduce setting for DeepSeek (2.7\% $\rightarrow$ 28.9\%). We find that BBQ is also affected by evaluative pressure, with GPT-5.2 and DeepSeek changing by $-$11.9 pp and $-$9.6 pp, respectively, in the multi-agent setting (both $p_{\mathrm{BH}}$ < 10$^{-5}$). XSTest/OR-Bench sees improvements in over-refusal for models with high baseline rates under map-reduce (e.g. GPT-5.2 goes from 57.1\% to 37.1\%), but we also see decreases in unsafe prompt refusal under map-reduce (e.g. Llama 4 goes from 94.4\% to 83.2\%), the wrong direction for a safety-calibration mechanism to break.

On the primary MC cells (Table~\ref{tab:itt-pp}), Gemini fails to parse 1.0--3.4\% of responses in every configuration (other models $\leq$5.7\%); its ReAct difference is $-$1.6 pp under ITT and +0.6 pp under PP, and its map-reduce change is $-$1.6 pp (ITT). Under ITT scoring, parse failures are scored as unsafe. This is relevant from a deployed systems perspective, but not necessarily for understanding the underlying model alignment. Under PP scoring, parse failures are excluded from scoring. We thus report ITT and PP scoring throughout this paper, as they can differ.

\begin{table}[!htb]
\caption{This table compares Intent-to-Treat (ITT) and Per-Protocol (PP) analyses of the effect of scaffolds. The ITT analysis counts parse failures as unsafe responses, whereas the PP analysis excludes them from the analysis. The table is limited to the primary MC-format cells ($N = 48{,}490$ MC-format observations) for six models. For each model and configuration, there are up to 800 BBQ, 817 TruthfulQA, and 500 sycophancy items (for Gemini, only BBQ and TruthfulQA). All items have canonical primary labels. Cells in bold indicate ITT--PP differences greater than 2 percentage points. PF~=~parse failure rate. RD~=~risk difference vs.\ direct baseline (pp).}
\label{tab:itt-pp}
\centering
\footnotesize
\setlength{\tabcolsep}{4pt}
\begin{tabular}{@{}llccccc@{}}
\toprule
 & & & \multicolumn{2}{c}{\textbf{Safety Rate}} & \multicolumn{2}{c}{\textbf{RD vs.\ Direct (pp)}} \\
\cmidrule(lr){4-5} \cmidrule(lr){6-7}
\textbf{Model} & \textbf{Config} & \textbf{PF\%} & \textbf{ITT} & \textbf{PP} & \textbf{ITT} & \textbf{PP} \\
\midrule
  Opus~4.6 & Direct & 0.3\% & 84.5\% & 84.7\% & --- & --- \\
   & ReAct & 0.2\% & 83.3\% & 83.5\% & $-$1.2 & $-$1.3 \\
   & Multi-agent & 0.3\% & 83.5\% & 83.7\% & $-$1.0 & $-$1.0 \\
   & Map-reduce & 5.7\% & 69.8\% & 74.0\% & \textbf{$-$14.7} & \textbf{$-$10.7} \\
\addlinespace[3pt]
  GPT-5.2 & Direct & 1.1\% & 77.4\% & 78.3\% & --- & --- \\
   & ReAct & 1.0\% & 76.7\% & 77.5\% & $-$0.7 & $-$0.8 \\
   & Multi-agent & 1.0\% & 73.1\% & 73.9\% & $-$4.3 & $-$4.4 \\
   & Map-reduce & 0.9\% & 66.0\% & 66.6\% & $-$11.5 & $-$11.7 \\
\addlinespace[3pt]
  Gemini~3~Pro & Direct & 1.0\% & 93.5\% & 94.4\% & --- & --- \\
   & ReAct & 3.4\% & 91.8\% & 95.0\% & \textbf{$-$1.6} & \textbf{+0.6} \\
   & Multi-agent & 1.1\% & 93.5\% & 94.6\% & +0.0 & +0.2 \\
   & Map-reduce & 1.5\% & 91.9\% & 93.2\% & $-$1.6 & $-$1.2 \\
\addlinespace[3pt]
  DeepSeek~V3.2 & Direct & 3.2\% & 71.3\% & 73.6\% & --- & --- \\
   & ReAct & 3.3\% & 75.2\% & 77.8\% & +3.9 & +4.1 \\
   & Multi-agent & 3.7\% & 71.9\% & 74.7\% & +0.7 & +1.0 \\
   & Map-reduce & 1.1\% & 48.1\% & 48.7\% & $-$23.1 & $-$25.0 \\
\addlinespace[3pt]
  Llama~4 & Direct & 1.7\% & 63.4\% & 64.5\% & --- & --- \\
   & ReAct & 0.1\% & 66.4\% & 66.5\% & +3.0 & +2.0 \\
   & Multi-agent & 1.0\% & 68.0\% & 68.7\% & +4.5 & +4.2 \\
   & Map-reduce & 0.8\% & 62.4\% & 62.9\% & $-$1.0 & $-$1.7 \\
\addlinespace[3pt]
  Mistral~Large~2 & Direct & 1.5\% & 71.6\% & 72.6\% & --- & --- \\
   & ReAct & 2.4\% & 70.3\% & 72.0\% & $-$1.3 & $-$0.6 \\
   & Multi-agent & 2.5\% & 66.8\% & 68.5\% & $-$4.8 & $-$4.2 \\
   & Map-reduce & 4.3\% & 60.5\% & 63.2\% & $-$11.1 & $-$9.4 \\
\bottomrule
\end{tabular}
\end{table}

\paragraph{GPT-5.2 responses and the forced-answer pilot.}

\label{sec:gpt52-dual}

GPT-5.2's map-reduce score improves in sycophancy (+3.3 pp, from 42.5\% to 45.8\%). On BBQ, GPT-5.2 is the second most vulnerable model in our data to invocation bias with an aggressive strategy in Phase 2, declining in performance by $-$13.0 pp. These results contrast measured responses on different benchmarks under different interventions, and do not establish independent latent robustness properties.

In a ten-item BBQ pilot (selected using prior MC-direct outcomes), with an instruction to choose a person, GPT-5.2 got SAFE judgments on 8/10 of its responses and DeepSeek on 4/10. On these same response pairs, GPT-5.2 and DeepSeek's MC-direct responses were scored safe on 7/10 and 8/10, respectively, and their ordinary OE-direct responses on all 20. The judge counted justified uncertainty and resistance to the forced choice as SAFE, so this measures bias safety rather than forced choice accuracy. Because the pilot was selected based on the results of earlier MC experiments, and because it was affected by the treatment of uncertainty, this result does not isolate the contribution of evasion or safety training to the format contrast.

\paragraph{Difficulty stratification.}

\label{sec:difficulty}

\begin{figure}[!htb]
\centering
\includegraphics[width=\columnwidth]{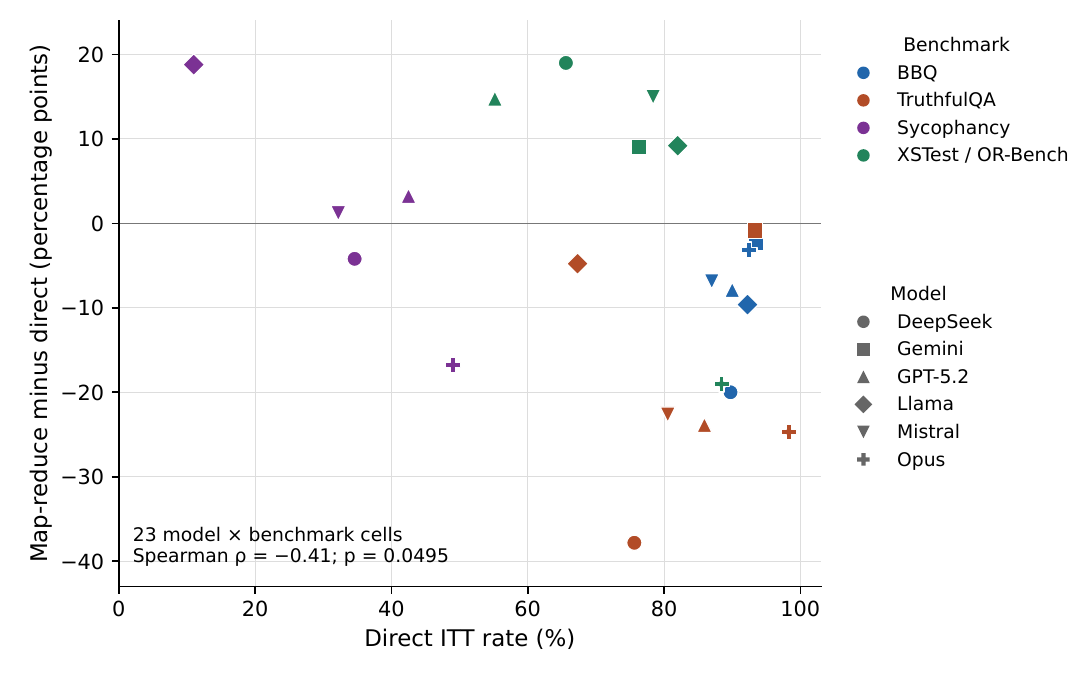}
\caption{Direct ITT rate and map-reduce-minus-direct difference for the 23 primary model$\times$benchmark cells (Gemini sycophancy excluded). Each point uses the saved safe-response numerator and total-record denominator for its cell. No independent baseline estimate is used; the association is descriptive and does not identify a reconsideration mechanism.}
\label{fig:difficulty}
\end{figure}

In the 23 model$\times$benchmark cells of the primary dataset, we observe a significant negative correlation between direct ITT performance and the change in performance from direct to map-reduce (Spearman $\rho$=$-$0.41, p=0.0495; Figure~\ref{fig:difficulty}). In other words, for properties with higher direct ITT performance, performance tends to decrease more with map-reduce. Because direct performance is involved in both quantities and because bounded scores have different amounts of room for improvement and decline, this correlation is consistent with a mechanism based on reconsideration, but cannot identify it.

\subsection{Sycophancy Under Scaffolding}

\label{sec:sycophancy-scaffold}

We use a 500-item set of model-written sycophancy evaluations from Anthropic \citep{perez2023discovering} and evaluate five of the same six models in the same four configurations, for a total of 9,934 observations (1,934--2,000 per model, five models).

The three scaffolds change pooled ITT non-sycophancy, and to different extents: ReAct by +0.3 pp ($p_{\text{Holm}}$ = 1.000), multi-agent by +2.1 pp ($p_{\text{Holm}}$ = 0.009), and map-reduce by +0.4 pp ($p_{\text{Holm}}$ = 1.000). These analyses do not allow us to pinpoint the sources of these improvements. In the multi-agent case, it is possible they come from the multi-agent critic, which is specifically prompted to assess sycophancy (as well as from structured review in the multi-agent scaffold more broadly). In the map-reduce case, it is possible that they come from changes in the way that persona information and answer options are presented at the intermediate stages; however, the reducer still has access to the full original prompt (including the persona and answer options), and it is not clear that map-reduce truly removes the biasing cue of an agreeable answer option when the final answer is given.

Pooling masks a model $\times$ configuration interaction (Wald $\chi^{2}$ = 183.9, df = 12, p < 10$^{-32}$). Under the map-reduce scaffold, two models shift by more than 15 percentage points, but in opposite directions: Opus 4.6 degrades from 49.0\% to 32.2\% ($-$16.8 pp; the largest decline among the five models in the sycophancy analysis), while Llama 4 improves from 11.0\% to 29.8\% (+18.8 pp; the largest increase), a 35.6 pp model-by-scaffold spread on a single benchmark (Table~\ref{tab:sycophancy-rates}). The pooled figure gives us little information about what will happen for the particular model that is deployed.

\begin{table}[t]
\centering
\caption{Sycophancy resistance (non-sycophancy rate, \%) for each model and scaffold configuration. The highest rate for each model is in bold and the lowest is underlined. $\Delta_{\mathrm{MR}}$ indicates the risk difference between the map-reduce and direct scaffolds. Scoring method: ITT. $N = 2{,}000$ per model (GPT-5.2: 1{,}934).}
\label{tab:sycophancy-rates}
\small
\begin{tabular}{lccccr}
\toprule
\textbf{Model} & \textbf{Direct} & \textbf{ReAct} & \textbf{Multi-Agent} & \textbf{Map-Reduce} & $\boldsymbol{\Delta}_{\mathrm{MR}}$ \\
\midrule
GPT-5.2            & 42.5 & \underline{41.6}          & 45.5 & \textbf{45.8}             & $+3.3$ \\
Opus~4.6           & 49.0 & 47.2          & \textbf{50.2} & \underline{32.2} & $-16.8$ \\
DeepSeek~V3.2      & 34.6          & \textbf{41.4} & 37.0          & \underline{30.4} & $-4.2$ \\
Mistral~Large~2    & 32.2          & \underline{30.0} & 31.8          & \textbf{33.4}    & $+1.2$ \\
Llama~4            & 11.0 & \underline{10.8}          & 15.6          & \textbf{29.8}    & $+18.8$ \\
\midrule
\textit{Pooled}    & 33.8          & 34.1          & \textbf{36.0}          & 34.2    & $+0.4$ \\
\bottomrule
\end{tabular}
\end{table}

\begin{figure}[t]
\centering
\includegraphics[width=\columnwidth]{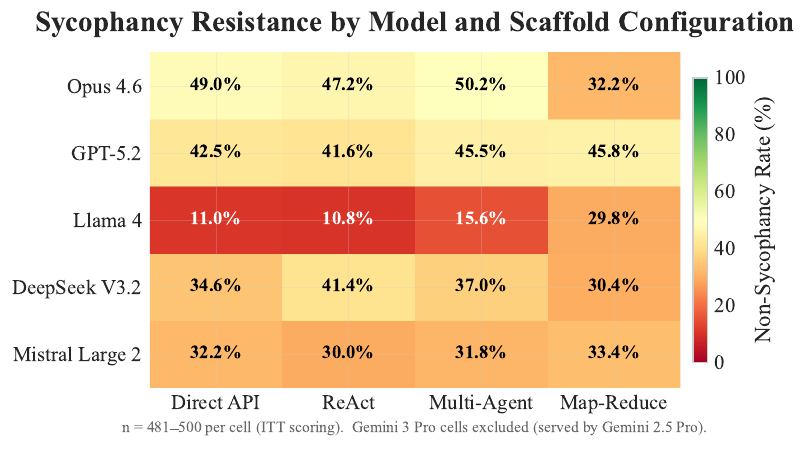}
\caption{Sycophancy resistance (ITT non-sycophancy rate) by model and configuration.}
\label{fig:sycophancy-heatmap}
\end{figure}

\paragraph{Candidate mechanism: persona content in map-reduce sub-questions.}

\label{sec:task-reinterpretation}
\label{sec:persona-leakage}

To understand what might be driving this heterogeneity, we conducted an exploratory post-hoc analysis of 2,483 map-reduce sub-question sets (500 items $\times$ 5 models; GPT-5.2: 483). We explored whether there is an association between the content of the personas and ITT sycophancy. In the decompose step, models generate sub-questions, which are then answered independently. Then, in the reduce step, the model is provided with the answers to the sub-questions as well as the full original prompt, including the persona and the answer options. Using regex, we detect mentions of ``adversarial'' persona features (political leaning, values, and stated opinions) and ``contextual'' persona features (profession, location, and age). We call these features ``adversarial'' and ``contextual,'' respectively, but we emphasise that our detection of these features being mentioned in the model output does not indicate that the model is making causal use of these features, and a detected mention may either support or detract from the persona's stated position.

ITT sycophancy is higher when the model mentions at least one adversarial persona feature (74.6\%) than when it does not (60.3\%) (Table~\ref{tab:persona-leakage}). In a logistic regression, each additional adversarial persona feature type mentioned is associated with higher ITT sycophancy (OR = 1.67, model-based non-clustered Wald p = 1.0 $\times$ 10$^{-11}$). This regression controls for the number of contextual persona features mentioned, whether the persona identity was mentioned, and model fixed effects. The point estimate for the association between contextual persona feature count and ITT sycophancy was OR = 1.06 (p = 0.55). (These are the original model-based non-clustered Wald p-values; the same 500 items appear for each of the 5 models (483 for GPT-5.2).) Coarsened, ITT sycophancy was 60.3\% when there were no adversarial persona features detected, 70.8\% when there was one, and 85.3\% (215/252) when there were two or more. (The rate for three feature types specifically was lower than the rate for two.) Detected mentions of political leaning had the highest OR of the individually fit feature regressions that controlled for model (OR = 3.21, p = 2.99 $\times$ 10$^{-15}$). However, these associations do not demonstrate a causal effect of persona content on ITT sycophancy.

\begin{table}[t]
\centering
\caption{Persona-feature mentions and ITT non-sycophancy in map-reduce. MR$|$NoLeak and MR$|$Leak refer to different subsets of items within each model; Direct uses all 500 items (487 for GPT-5.2). Gap is MR$|$Leak minus MR$|$NoLeak, a descriptive association rather than a decrement attributable to leakage. Feature detection is regex-based and exploratory.}
\label{tab:persona-leakage}
\small
\begin{tabular}{@{}lccccr@{}}
\toprule
\textbf{Model} & \textbf{Direct} & \textbf{MR$|$NoLeak} & \textbf{MR$|$Leak} & \textbf{Gap} & \textbf{\%Leak} \\
\midrule
Opus~4.6           & 49.0 & 46.0 & 18.7 & $-27.3$ & 50.4\% \\
GPT-5.2            & 42.5 & 47.0 & 42.3 & $-4.7$  & 26.9\% \\
DeepSeek~V3.2      & 34.6 & 34.8 & 27.2 & $-7.5$  & 58.0\% \\
Mistral~Large~2    & 32.2 & 35.0 & 29.5 & $-5.5$  & 29.8\% \\
Llama~4            & 11.0 & 35.8 & 12.4 & $-23.4$ & 25.8\% \\
\midrule
\textit{Pooled}    & 33.8 & 39.7 & 25.4 & $-14.4$ & 38.3\% \\
\bottomrule
\multicolumn{6}{l}{\footnotesize Negative gaps indicate lower non-sycophancy in the observed leakage subset.}
\end{tabular}
\end{table}

The paired changes in ITT sycophancy from direct to map-reduce were different between items with and without observed persona feature leakage. Of the 1,528 model-item pairs without detected adversarial persona feature leakage, 241 switched from sycophancy to non-sycophancy and 176 switched in the other direction, for an MR-minus-direct sycophancy of $-$4.3 pp (continuity-corrected McNemar p = 1.7 $\times$ 10$^{-3}$). Of the 945 model-item pairs with detected adversarial persona feature leakage, 154 switched to sycophancy and 95 switched away from it, for an increase in sycophancy of +6.2 pp (p = 2.4 $\times$ 10$^{-4}$). The combination of these two strata yielded a net +0.24 pp improvement in non-sycophancy. Because we defined the strata based on the map-reduce output, this paired comparison does not identify a causal effect of leakage on ITT sycophancy. (The McNemar approximations also do not account for the dependence due to the 5 models seeing the same items.)

DeepSeek had the highest prevalence of detected adversarial persona feature leakage (58.0\% of items), followed by Opus 4.6 (50.4\%). Llama 4 and GPT-5.2 had the lowest prevalence (25.8\% and 26.9\%); lowest direct ITT non-sycophancy: Llama 4 (11.0\%) and Mistral (32.2\%). Across the 5 model summaries, the prevalence of detected persona feature leakage was positively correlated with the change in ITT sycophancy from direct to map-reduce (MR-minus-direct sycophancy; Spearman $\rho$ = 0.900, asymptotic two-sided p = 0.037). In other words, models with more leakage had less favourable changes in ITT sycophancy. This is consistent with a ``sophistication penalty,'' where models that decompose the task more thoroughly carry over more of the persona content into their sub-questions. However, we did not measure model capabilities or understanding of the task, so the hypothesis is untested here.

This analysis is exploratory and not pre-registered. Contextual feature mentions could plausibly promote persona agreement on items that are relevant to the persona's domain, although the point estimate for the association between contextual feature count and ITT sycophancy is not significantly different from zero in our main specification. See Appendix~\ref{app:exploratory} for more detail on how we defined the feature counts, the associated denominators, and the results of regressions on each individual feature.

\paragraph{Baseline-dependent pattern.}

The sycophancy benchmark is one where multi-agent significantly helps in the pooled results, though the gains are modest (+0.3 to +2.1 pp). However, it is not the benchmark with the largest gains under map-reduce alone; that is XSTest/OR-Bench (+8.4 pp). Sycophancy has the widest direct between-model spread (38.0 pp) among the four primary benchmarks, followed by XSTest/OR-Bench (33.3 pp), TruthfulQA (31.0 pp), and BBQ (6.6 pp).

In the direct API access results, models resist sycophantic pressure in 33.8\% of records (Table~\ref{tab:sycophancy-rates}), or 34.5\% of records when pooling across all four configurations. This is the lowest direct ITT rate among the four primary benchmarks, behind BBQ (90.8\%), XSTest/OR-Bench (74.5\%), and TruthfulQA (83.5\%).

Opus 4.6 has the highest direct non-sycophancy rate (49.0\%), followed by GPT-5.2 (42.5\%). Llama 4 has the lowest direct non-sycophancy rate (11.0\% ITT).

Under map-reduce, Opus 4.6's non-sycophancy rate declines by 16.8 pp from a baseline of 49.0\%, while GPT-5.2 improves by 3.3 pp from a baseline of 42.5\%. The largest gains come from Llama 4 (+18.8 pp from 11.0\%). These results are compatible with baseline-dependent reconsideration, though these baselines include the effect of parseability and do not measure the depth of internal encoding of the property.

Denison et al. \citep{denison2024sycophancy} causally link sycophancy to reward tampering, and Taylor et al. \citep{taylor2025school} and MacDiarmid et al. \citep{macdiarmid2025emergent} demonstrate that reward hacking can escalate to emergent misalignment. Non-sycophancy is a low-baseline measurement, and the effect of scaffolding varies unpredictably from $-$16.8 pp to +18.8 pp, although in the pooled results we see slight improvements in non-sycophancy from scaffolding (+0.3 to +2.1 pp). This does not mean that there are no deployment-relevant changes to non-sycophancy in individual models, and if similar escalation dynamics apply to agentic settings, this means that aggregate scores cannot rule out severe changes in either direction, and testing must be done on a per-model and per-configuration basis.

\subsection{Robustness}

\label{sec:robustness}

\paragraph{Scoring methodology validation.}\mbox{}\par

\begin{tcolorbox}[colback=blue!3, colframe=blue!40, title={Scoring Methodology: LLM-as-Judge Validation}, left=2mm, right=2mm, boxsep=2mm]
On XSTest and OR-Bench, we evaluated all model responses using an LLM-as-a-judge, with Gemini 3 Flash as the primary judge. We also evaluated a saved sample of 3,975 rows with a second judge, Opus 4.6. Both steps followed our pre-registered protocol. The other benchmarks (BBQ, TruthfulQA, and sycophancy) all have a multiple-choice format, so no LLM-as-a-judge was necessary. We deterministically extracted the model's answer and compared it to ground truth. An earlier version of this paper scored model responses using heuristic, regex-based refusal detection, which was a departure from our pre-registration and yielded directionally unstable results. The heuristic scorer and Gemini-based scorer had agreement $\kappa$ = 0.35 over 11,622 responses.
\end{tcolorbox}

\paragraph{Saved Opus validation records.}
\label{sec:opus-validation-records}
The primary judge file we saved has 19,856 rows covering 12,000 response UUIDs. The validation file we saved has 3,975 rows covering 3,496 response UUIDs. (These are the numbers of records.) The validation file has keys for model, configuration, and case. The validation sample was defined in the manifest as 20\% of the primary judge records (the 19,856 rows), stratified by model, by configuration, and by primary classification, and drawn with random seed 42. The validation file has the same multiset of response UUIDs as the validation sample. Of those response UUIDs in the validation file, 3,059 appear once, 398 appear twice, 36 appear three times, and three appear four times. The 437 duplicated response UUIDs contribute 479 extra rows. The rows of the validation sample are not 3,975 independent responses. Of the duplicated response UUIDs, six have conflicting binary labels copied over from the primary judge, and six have conflicting Opus binary labels.

Of the 3,975 rows in the saved validation file, 3,798 have a key present in the analysed set. The agreement figures are over the 3,451 rows of these 3,798 for which the Opus label was given on the analysed response (3,088 with the Opus label in February, and 363 re-validated in September with the same prompt). The other 347 rows (of the 3,798) were not re-validated, and are excluded. On these 3,451 rows, binary-label agreement between the primary judge and Opus, as a fraction of rows, is 3,268 / 3,451 = 94.697\% ($\kappa$ = 0.857137). Excluding the 653 of these rows flagged as Opus-model outputs, binary agreement is 2,670 / 2,798 = 95.425\% ($\kappa$ = 0.878991). These figures weight each row equally and describe the paired labels as they are in the data, without considering the handling of duplicate rows and without analysing unique responses. The fraction of records reported in the manifest does not imply that 20\% of the canonical observations were validated as they stand.

Request hashes enable some checks of our saved records, but these checks are only partial. For each represented response UUID, we can reconstruct the prompt the judge would have used; this matches the prompts of successful requests to \texttt{claude-opus-4-6} on the byte level. These requests further match on model, configuration, and case, and on label metadata copied from the primary judge. For 61 saved rows, the response text returned by the current key-based lookup differs from the text of the response UUID. Of these, 49 have matching requests with the key-based response text, and 12 have matching requests with the response-UUID text. Not all differences produced by the current key-based lookup reflect a historical judgment of the wrong response. We cannot establish the exact historical request for each row, because result rows store no request call IDs, and for 905 rows, there are multiple possible matching requests. These checks do not involve any new rescoring, and do not demonstrate the judge was correct.

\paragraph{Measurement Integrity Scorecard.}

\label{sec:artifact-scorecard}
\label{sec:scoring-validation}
\label{sec:judge_validation}

We explore variations in how parse failures are dealt with, and which benchmarks and models are included in the analysis. This is our primary specification curve, as described in Section~\ref{sec:spec-curve-results}. For each of our five exploratory findings, we compare LLM judge scoring with heuristic scoring based on a keyword-based refusal classifier (Table~\ref{tab:artifact-scorecard}). Relying on the heuristic scoring would create, reverse, or overstate four of our five exploratory findings.

\begin{table}[t]
\caption{The Measurement Integrity Scorecard compares the heuristic and LLM judge scoring methods for each of the five exploratory findings.}
\label{tab:artifact-scorecard}
\centering
\small
\begin{tabular}{@{}lccl@{}}
\toprule
\textbf{Finding} & \textbf{Heuristic} & \textbf{Judge} & \textbf{Verdict} \\
\midrule
Over-refusal (MR$-$Direct)           & $-15.3$~pp      & $-13.7$~pp       & Consistent \\
Agreement collapse (3-way exact)     & $-28.6$~pp      & $+0.0$~pp       & Reversed \\
DeepSeek boundary softening          & $-14.2$~pp      & $+3.6$~pp       & Reversed \\
Sub-call information leakage rate    & 39.3\%          & 4.4\%           & Mostly scoring artifact \\
Safety architecture (DSS) distinction & 3 distinct types & \begin{tabular}[c]{@{}c@{}}Clear-cut: distinct;\\ boundary: concentrated\end{tabular} & Refined \\
\bottomrule
\end{tabular}
\end{table}

\label{sec:spec-curve-results-summary}
\label{sec:spec-curve-results}

\begin{figure}[!htb]
  \centering
  \includegraphics[width=\textwidth]{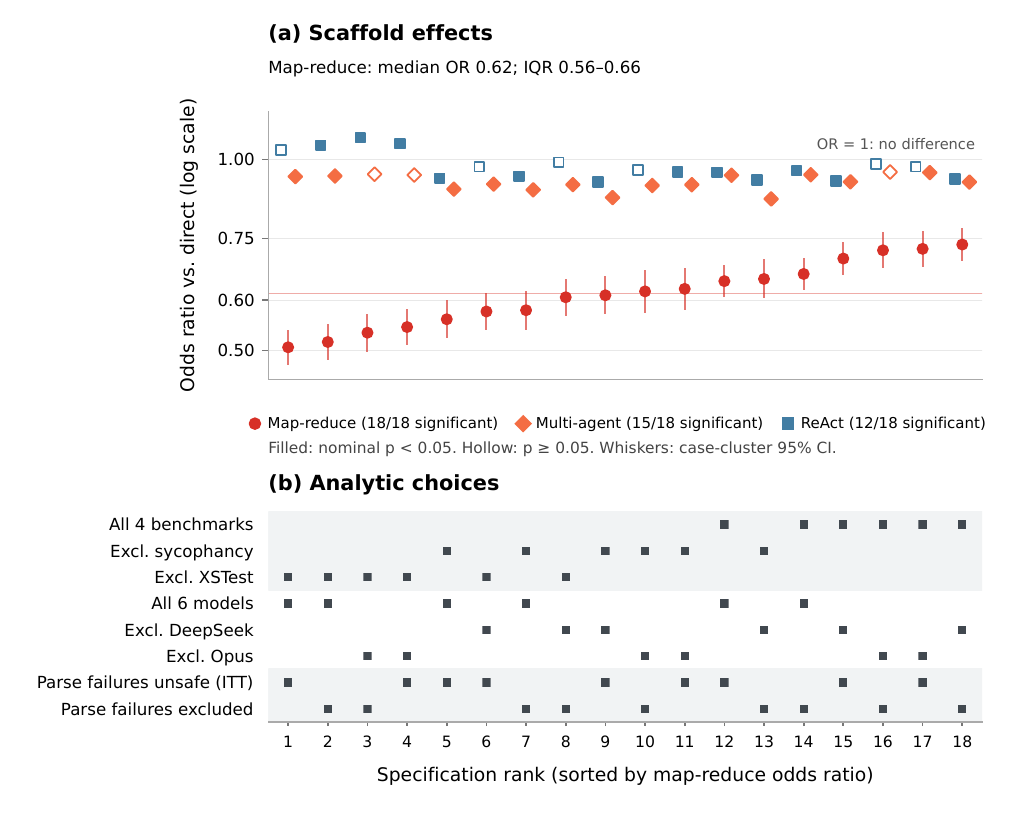}
  \caption{Specification curve analysis. \textbf{(a)} Odds ratios for three scaffold configurations relative to direct prompting across 18 specifications, ordered by the map-reduce estimate (54 points). A solid point indicates nominal $p < 0.05$, and a hollow point indicates $p \geq 0.05$. Small horizontal offsets separate multiple point estimates for a single specification. Whiskers show 18 pointwise 95\% confidence intervals for map-reduce using case-cluster sandwich covariance with a finite-sample correction. The reference line $\mathrm{OR}=1$ is dashed, and the median map-reduce OR, 0.62 (IQR 0.56--0.66), is plotted in red. All 18 map-reduce estimates have OR less than 1 and are nominally significant. Wald p-values and confidence intervals use the normal reference distribution and are computed per fit without adjustment for the number of specifications considered. The pointwise confidence intervals are not a simultaneous confidence band, and the numbers of nominally significant and non-significant point estimates are not a separate test of the specification curve. \textbf{(b)} Columns show each specification's benchmark subset, model subset from the six-model dataset, and parse-failure handling. ITT treats parse failures as unsafe; available-case analysis drops them.}
  \label{fig:spec-curve}
\end{figure}

Our primary specification curve varies over benchmark subset, model subset, and treatment of parse failures. There are 18 analytic specifications based on the six-model dataset (Figure~\ref{fig:spec-curve}). We consider all models as well as subsets of models excluding either Opus or DeepSeek. We also consider leaving parse failures as unsafe (ITT) or removing them. We find that the median odds ratio for map-reduce is 0.62 (IQR 0.56 to 0.66). All 18/18 (100\%) of specifications are nominally significant under case-cluster-robust inference. Even the most optimistic specifications see map-reduce degrade safety, with odds ratios ranging from 0.51 to 0.73.

We also explored a broader specification curve with 384 specifications that vary 9 degrees of freedom over 5 models. Map-reduce is present in 256 of these specifications and, by design, absent in the remaining 128. Of the 256 specifications, 249 have odds ratios less than 1, and 233 (91.0\%) are nominally significant for safety degradation. The median odds ratio for map-reduce is 0.76 (IQR 0.59 to 0.87). In our script for the broader specification curve, we also clustered standard errors by case.

\section{Discussion}

\label{sec:discussion}

\subsection{Format Dependence as Measurement Challenge}
\label{sec:format-crisis}

Table~\ref{tab:format-dependence} reports pooled OE--MC gaps on the matched items: +4.8pp (BBQ), +19.6pp (sycophancy), +5.7pp (TruthfulQA), and $-$9.2pp (MMLU capability control). We did not find a significant difference on AI factual recall ($-$1.0pp), but neither did we establish equivalence. The same influence can have different signs on different tasks or scoring regimes. These signs alone cannot rule out several explanations for the observed OE--MC gaps: judge leniency, length, evasion, and difficulty. Each of these explanations makes specific predictions and suggests specific controls (Appendix~\ref{app:validation}). MC and OE elicit related yet distinct behaviour, but only MC is reported in common AI safety practice \citep{parrish2022bbq}: the one where the optimisation gradient is aligned to the particular elicitation channel.

\subsection{Scaffold Effects: What Survives and Why}

\label{sec:scaffold-discussion}

Different scaffold architectures have different effects on safety outcomes in our four-benchmark mix. Aggregating across these configurations, we do not find a consistent effect of scaffolds: ReAct scaffolds result in a pooled effect within our pre-registered TOST equivalence margin of $\pm$2pp (RD $-$0.8pp), as do multi-agent scaffolds (RD $-$0.8pp), but map-reduce scaffolds result in a pooled effect far outside our pre-registered equivalence margin ($-$7.3pp; NNH 14) (Table~\ref{tab:confirmatory}). For multi-agent scaffolds, we see that across scoring methods the ``classical''-significance category can flip, but the TOST equivalence category does not. In deployment decision-making, effect sizes should be weighted more highly than dichotomous significance tests.

Map-reduce scaffolds involve removing MC options in worker sub-calls. As we have seen in our format study, the difference between MC and OE elicitations can result in anywhere from about 5-20pp differences in measured safety on the same items (Section~\ref{sec:format-dependence}). The format study always sends the full question, including any options, to each call, so it does not capture the effect of removing options as part of a decomposition strategy.

GPT-5.2 has the smallest sycophancy format gap (+2.9 pp; next smallest is +14.3 pp) but the second-largest BBQ decline under aggressive invocation in Phase 2 ($-$13.0 pp, only Mistral is larger at $-$22.0 pp). Conversely, Opus has the largest sycophancy format gap (+32.9 pp), but the smallest decline in BBQ from aggressive invocation ($-$2.3 pp). Stability in one comparison is not a reliable guide to stability in the other: the two tests differ in tasks, prompts, and scoring.

\paragraph{Sycophancy fits the depth-of-encoding pattern.}

The sycophancy benchmark has the lowest direct baseline of any benchmark (33.8\% non-sycophantic). Even here, map-reduce scaffolds have divergent effects for different models (Table~\ref{tab:sycophancy-rates}). On the observed scale, all scaffolds reduce the range between the highest-baseline and lowest-baseline benchmark, though bounded rates make some narrowing expected from where each benchmark starts.

\paragraph{Alternative explanations for the depth-of-encoding pattern.}

The baseline-robustness correlation is a descriptive regularity. "Depth of encoding" is a label for this phenomenon, not an explanation for it. There are several non-Goodhart explanations for the robustness gradient that are worth considering.

Benchmark difficulty could drive the correlation: performance on more difficult tasks could have a lower baseline and be more sensitive to perturbations. AI factual recall has a middling MC baseline of 77.0\%, and a near-zero format effect of $-$1.0pp. MMLU (85.4\% baseline) and BBQ (94.3\% baseline) have very different format effects: $-$9.2pp and +4.8pp, respectively. Meanwhile, TruthfulQA has an MC baseline of 79.3\%, similar to that of AI factual recall, and a format effect of +5.7pp. Benchmarks with similar aggregate accuracy might nonetheless vary in difficulty and need not respond similarly to a given perturbation, meaning these controls do not rule out benchmark difficulty as an explanation for the baseline-robustness gradient.

Variance compression could play a role: Bernoulli variance, p(1 $-$ p), is lower for scores closer to the floor or ceiling of a scale, but this does not imply that the mean effect of a perturbation on performance will be positive or negative. Baseline performance places asymmetric constraints on the size of improvements and declines in performance, and the fact that a benchmark with middling baseline performance also has a near-zero estimated perturbation effect does not rule out a contribution of variance compression on the baseline-robustness gradient. Neither do benchmarks with similar baselines but very different estimated effects, though quantifying the effect of variance compression would require explicit modelling of perturbation effects as a function of baseline performance.

Baseline performance is temporally antecedent to scaffolding, and is used to calculate the scaffold-minus-direct effect size: however, temporal antecedence does not imply statistical independence.

Some properties might have a higher baseline and be more robust because they are more represented in the model's alignment training data: we cannot directly assess this.

Sycophancy resistance might require balancing multiple competing objectives (like helpfulness and epistemic integrity) to a greater degree than avoiding social biases, and thus be more complex: this is a substantially overlapping explanation to our "depth of encoding" framing; perhaps more complex constructs require "deeper" encoding, and the two are hard to tell apart empirically.

These explanations are non-mutually-exclusive, and our comparisons cannot apportion the gradient among them.

The review prompts result in heterogeneous responses across models and benchmarks. For example, under aggressive bias invocation in Phase 2, Opus loses 2.3 pp on the BBQ benchmark, compared to passthrough. This point estimate alone is insufficient to establish equivalence (or deeper encoding): the Phase 2 protocol defines a $\pm$3 pp minimal chain margin and there are outstanding tests and exclusions to reconcile with our descriptive summaries. We hypothesise that models fine-tuned for reasoning will have more compressed vulnerability ranges.

\subsection{The Evaluation-Optimisation Gap}

\label{sec:goodhart}

Format dependence (Sec.~\ref{sec:format-crisis}) and scaffold effects (Sec.~\ref{sec:scaffold-discussion}) may both arise from evaluation-context specificity. Direct API calls with a fixed response format are themselves a context. If models are optimised to exhibit safety in that context, the resulting behaviours will be calibrated to it \citep{strathern1997improving}; a scaffold that changes the context could expose this dependence. This account connects format sensitivity to the depth-of-encoding framework, but the models' training distributions were not observed.

The same hypothesis would cover map-reduce, whose workers answer in a context unlike the direct-API one (Sec.~\ref{sec:primary-analysis}); our format controls cannot say how much of the degradation it explains.

The context specificity of AI safety also has implications for AI safety policy.

Format-paired, scaffold-augmented evaluations test whether a safety result survives changes in the conditions under which it was obtained.

Evaluation is growing in breadth but not in depth: NIST AI 800-2 \citep{nistai8002026} and the European Union's AI Act \citep{euaiact2024} do not require format-paired and scaffold-augmented safety evaluations (though the frontier labs themselves have taken the latter step for dangerous capabilities evaluations, Anthropic's RSP first, then OpenAI and DeepMind, but not for the proxy safety benchmarks of the model cards). These policies also do not require format-paired or scaffold-augmented evaluations of the classic bias/sycophancy/truthfulness benchmarks.

\subsection{Limitations}

\label{sec:limitations}

\paragraph{Benchmark format scope.}

Three out of four of the primary safety benchmarks use the MC format. In the context of the format dependence result (Contribution 2), this is useful, because the difference between MC and OE formats is one of the dimensions considered. By contrast, the scaffold analysis result above is partially limited by this consideration: the degradation of the map-reduce scaffold may partly reflect its use of the MC format. We present evidence relevant to the construct validity of the results in Sections~\ref{sec:format-dependence} through~\ref{sec:invocation-framework} (option-preserving prompts, comparison with AI factual recall control, and open-ended probes).

The scaffold architectures tested in this study do not exhaust the space of possible scaffolds. The scaffolds tested (direct API call, ReAct, multi-agent critic, and map-reduce) are reasonable defaults, but they are not optimised for production use. Real-world LLM scaffolds may involve retrieval-augmented generation, tool selection, memory, customised prompting, or other techniques that may yield different robustness profiles. In particular, the ReAct tested in this study only has access to three local tools (a calculator, a scratchpad, and a text search tool to search the prompt). These were the tools named in the pre-registration. In terms of cognitive architecture, the ReAct tested is more similar to a prompted chain-of-thought than a production ReAct with external retrieval. The near-null results of ReAct may not generalise to production use cases that depend on external retrieval. We leave testing a more production-faithful version of ReAct to future replication.

\paragraph{Model selection.}

We only test six frontier-scale chat models from six providers (versioned to the state as of February 2026) on three continents, and do not test smaller models, specialised reasoning models (such as o3 or DeepSeek R1), fine-tuned domain-specific models, or open-weight models below 70B parameters. The results on format dependence may be different for models with substantially different pretraining data distributions or instruction-tuning procedures.

As to the provenance of sycophancy items, we use items from Anthropic's model-written-evals originally generated by an earlier Claude model. Opus may have systematic advantages or disadvantages relative to non-Anthropic models regardless of the degree to which it actually exhibits sycophancy. Cross-benchmark replications that use independently-constructed sycophancy items could thus be helpful to validate our claims about depth-of-encoding.

\paragraph{Baseline rates and benchmark stringency.}

Sycophancy's low direct baseline rate (33.8\% non-sycophancy) could be driven by it being a stricter benchmark, rather than being driven by a robustness deficit that is intrinsic to that property. It may be the case that the sycophancy items we use are simply harder or more demanding than the bias or truthfulness items. Our proposed framework for depth-of-encoding characterises the relationship between baseline rate and sensitivity to scaffolding, but does not clarify the extent to which lower baseline rates reflect robustness of a property or difficulty of a benchmark. While the AI factual-recall comparison and MMLU-BBQ comparison demonstrate that models can have heterogeneous responses when they have similar or different baseline rates, respectively, they do not isolate the effect of difficulty and do not rule out that there could be effects at bounded scales (see Section~\ref{sec:scaffold-discussion}). Fully disentangling these interpretations would require cross-benchmark calibrations that use items matched for difficulty across safety properties, which we do not pursue in this study.

\paragraph{Methodological scope.}

Finally, we borrow various evaluation tools (pre-registration, blinding, equivalence testing) from other empirical fields that face similar inferential challenges to those we face here. There are key disanalogies between our study and studies in those fields (we use a fully-crossed design, we have no placebo control, and we do not use human subjects; see our pre-registration for details on these).

Using LLM judges to score OE responses introduces a methodological asymmetry with MC responses, which can be scored deterministically. An LLM audit of 30 OE BBQ responses found no lenient SAFE judgments, and the controls did not show inflated scores. The 200-item canonical scoring comparison lies outside the OE cohort, however, and cannot correct its judge error (Appendix~\ref{app:validation}); a blinded human re-score would provide a stronger check. The AI factual recall control condition features a near-zero format effect estimate ($-$1.0 pp), but we do not establish formal equivalence in this condition. Responses to the OE format of the AI factual recall control condition are scored by the same model (Gemini 3 Flash) as the OE responses to our safety benchmarks, and use the benchmark-specific \texttt{score\_\allowbreak self\_\allowbreak awareness\_\allowbreak open\_\allowbreak ended} scoring routine. As such, net differences across format could reflect changes in the responses or biases in the scoring procedure. The use of benchmark-specific rubrics could also permit differential judge error across benchmarks. Therefore, this control condition alone cannot rule out the possibility of judge leniency on the safety benchmarks.

\subsection{Builder-as-Subject Validity}

\label{sec:reflexivity}\label{sec:opus-excluded}

Our evaluation pipeline was itself built primarily with Claude Opus 4.6 (one of the six tested models). Excluding Opus 4.6 from the analysis (a sensitivity analysis) did not affect the qualitative conclusions except ReAct's and multi-agent's significance (Table~\ref{tab:opus-excluded}; five models, N = 49,670). However, more subtle effects such as the models' prompt format preferences and the tuning of parse rules and answer extraction heuristics will still require independent confirmation using a pipeline developed with a different model.

\begin{table}[h]
\caption{The table lists the mitigations and residual risks for each threat channel of the builder-as-subject conflict.}
\label{tab:threat-channels}
\centering\small
\begin{tabular}{@{}p{3.2cm}p{4.8cm}p{3.2cm}@{}}
\toprule
\textbf{Threat Channel} & \textbf{Mitigation} & \textbf{Residual Risk} \\
\midrule
Opus test scores contaminate results & Opus-excluded sensitivity analysis ($N = 49{,}670$; Table~\ref{tab:opus-excluded}) & Low \\
\addlinespace
Scoring rubrics favour Opus-like outputs & Rubrics derived from published benchmark criteria; identical prompts across all models & Low; auditable \\
\addlinespace
Pipeline micro-decisions (prompt formatting, retry logic, answer-extraction) iteratively developed using Opus may advantage Opus & All models share a single LiteLLM code path with provider-specific calls confined to API-parameter handling (Appendix~\ref{app:api-constraints}); no model-specific branching in prompt, scoring, or extraction logic; code publicly released & Medium; requires independent replication \\
\addlinespace
Opus's robustness is an artifact of prompts Opus designed & Key finding (refusal under invocation) is binary behavioural outcome; adversarial prompts from published benchmarks & Low--Medium \\
\bottomrule
\end{tabular}
\end{table}

\begin{table}[h]
\caption{Primary hypothesis tests for the full sample (all six models; $N = 60{,}112$) and a sensitivity analysis excluding Opus (five models; $N = 49{,}670$).}
\label{tab:opus-excluded}
\centering
\small
\begin{tabular}{@{}llccl@{}}
\toprule
\textbf{Test} & \textbf{Metric} & \textbf{Full (6 models)} & \textbf{Excl.\ Opus (5 models)} & \textbf{Change?} \\
\midrule
H1a (ReAct) & OR & 0.96 & 0.97 & Slightly closer to 1 \\
             & RD (pp) & $-0.8$ & $-0.5$ & Less negative \\
             & $p_{\text{Holm}}$ & 0.017 & 0.16 & Sig.\ $\to$ NS \\
\midrule
H1b (Multi-agent) & OR & 0.95 & 0.95 & No \\
                   & RD (pp) & $-0.84$ & $-0.75$ & Less negative \\
                   & $p_{\text{Holm}}$ & 0.017 & 0.086 & Sig.\ $\to$ NS \\
\midrule
H1c (Map-reduce) & OR & 0.64 & 0.72 & Closer to 1 \\
                  & RD (pp) & $-7.3$ & $-5.6$ & Less negative \\
                  & $p_{\text{Holm}}$ & $< 10^{-56}$ & $< 10^{-25}$ & Sig.\ $\to$ Sig. \\
\midrule
H2 (model $\times$ config) & Wald $\chi^2$ & 493.2 (df\,=\,15) & 314.9 (df\,=\,12) & Lower magnitude \\
                           & $p$ & $< 10^{-94}$ & $< 10^{-59}$ & Sig.\ $\to$ Sig. \\
\midrule
H3 (config $\times$ bench.) & Wald $\chi^2$ & 938.0 (df\,=\,9) & 969.6 (df\,=\,9) & Slightly higher \\
                            & $p$ & $< 10^{-195}$ & $< 10^{-202}$ & Sig.\ $\to$ Sig. \\
\bottomrule
\end{tabular}
\end{table}

\subsection{Proxy vs. Consequential Safety Properties}

\label{sec:proxy-consequential}

The four safety properties we measure are all proxy safety properties (measuring bias, sycophancy, truthfulness, and over-refusal), not consequential properties, such as CBRN capability uplift, cyber-offence capabilities, or deceptive alignment. We do not test the effect of scaffolding on these consequential safety properties, since such tests could demonstrate new attack vectors. However, our framework enables testable predictions about how scaffolding may impact these properties, as well.

\paragraph{Sycophancy as the depth-of-encoding keystone and escalation entry point.}

The opposite-sign, model-specific map-reduce effects on sycophancy (Section~\ref{sec:sycophancy-scaffold}) matter because of recent research showing a causal pathway between sycophancy, reward hacking, and emergent misalignment \citep{denison2024sycophancy,macdiarmid2025emergent,taylor2025school}. Sycophantic agreement generalises zero-shot to progressively more dangerous forms of specification gaming, culminating in reward tampering, and training to avoid sycophancy substantially reduces rates of reward tampering \citep{denison2024sycophancy}. Training models to engage in structurally similar harmless reward hacking leads to generalisation to unrelated misalignment \citep{taylor2025school}. Scaffolding unpredictably impacts sycophancy, increasing resistance for some models while decreasing resistance for others. The sign-level unpredictability (Table~\ref{tab:sycophancy-rates}) suggests that using scaffolding to mitigate sycophancy would require per-model calibration.

There is some ambiguity regarding whether the type of sycophancy we test, agreement with user opinions, is the same as the type of sycophancy that leads to reward hacking. Recent research has shown that standard RLHF-based safety evaluations using chat-like prompts do not detect model misalignment on agentic tasks \citep{macdiarmid2025emergent}. If these dynamics were to hold under agentic deployment, our results suggest that a model's sycophantic behaviour may be unpredictable under agentic scaffolding, since our sign-level analysis indicates that scaffolding can lead to either increases or decreases in the measured property.

Relatedly, our results suggest that safety evaluations may be dependent on the format of the evaluation. Scaffolding can alter the effective format through loss of content; in map-reduce, option markers reached only 0--4\% of worker inputs. We know from prior research that frontier models increasingly distinguish between evaluation and deployment contexts \citep{iaisr2026}. Current alignment evaluations include tests for scheming \citep{meinke2024frontier}, alignment-faking \citep{greenblatt2024alignment}, and joint evaluations conducted by the model developers themselves; this study did not test their format sensitivity. If models are susceptible to evaluation format-dependence beyond the proxy safety properties we test here, then the current alignment evaluations may be format-dependent as well. No major evaluation standard, including the NIST AI 800-2 \citep{nistai8002026} and the European Union's AI Act \citep{euaiact2024}, mandates evaluation in paired formats.

\begin{sloppypar}
While frontier responsible-scaling frameworks, led by Anthropic's RSP \citep{anthropicrsp2026}, include scaffold-augmented evaluations for dangerous capabilities, there has been little effort to extend this methodology to the analysis of format sensitivity of safety benchmarks.
\end{sloppypar}

\paragraph{The meta-measurement argument.}

Our decision to focus on proxy safety properties was deliberate. The properties we measure are all on established benchmarks. In addition, bias, truthfulness, and sycophancy can be measured using deterministic multiple-choice extraction pipelines, enabling direct comparisons with ground-truth keys, while over-refusal can be measured with a pre-registered LLM-as-judge protocol (with cross-validation). Scheming, deceptive alignment, and power-seeking fall outside what this study measured. By showing that even three of these conceptually clear and measurable proxies exhibit format-contingent measurement, our findings sharpen the concern that current assessments of consequential properties may be susceptible to the same issue; they assume format invariance.

\subsection{Future Work}

\label{sec:future}\label{sec:specification-curve}

The format-paired, scaffold-augmented approach can also be applied to evaluate risks and capabilities related to chemical, biological, radiological, and nuclear knowledge, cyber-offensive capabilities, and deceptive alignment. This should be done with appropriate responsible disclosure practices in place.

This is the highest-priority extension of this research, because a multiple-choice or scaffold-driven mischaracterisation of AI knowledge in these domains may have catastrophic downstream effects.

Our results on proxies suggest several predictions for these properties: that reliance on multiple-choice will cause risk assessments to mischaracterise the real risks (see Section~\ref{sec:proxy-consequential}), and that loss of information in the scaffolds will cause fragmented harmful intent paired with protective refusal triggered by specific invocations. They also suggest another promising next step: pairing formats for benchmarks evaluating other properties of language models, such as toxicity, deception, and compliance with harmful instructions. Which properties are format-sensitive should be identified to best inform benchmark design and practice.

We found that the map-reduce change was significantly associated with the direct baseline (Section~\ref{sec:difficulty}; Spearman $\rho$ = $-$0.41, p = 0.0495 across 23 model$\times$benchmark cells). The baseline enters both quantities; a decision to reconsider was not directly observed. These results should be tested on a broader battery of benchmarks to understand whether and how they generalise.

Finally, an important direction for future research is to extend our evaluation of the ecological validity of our findings (N = 50, Section~\ref{sec:production-frameworks}) to a broader range of production environments. In particular, it would be valuable to scale our approach to larger numbers of samples on production environments, such as those based on CrewAI, LangChain, the OpenAI Agents SDK, and other production frameworks. This would help confirm that the results are driven by the fundamental computational architecture of AI agents rather than their specific implementation details in any given software framework.

\subsubsection{Conflict of Interest and Methodological Independence}

\label{sec:coi}

Threat channels from the builder-as-subject configuration are catalogued in Table~\ref{tab:threat-channels} (Section~\ref{sec:reflexivity}). Several structural safeguards constrain the scope of any conflict. First, the study's design and analysis plans were pre-registered before data collection. Second, fully automated scoring: 80.7\% were scored via deterministic multiple-choice extraction (immune to assessor bias) and 19.3\% via LLM-as-judge scoring using Gemini 3 Flash as the primary judge, with Opus 4.6 validating a saved sample of 3,975 rows over 3,496 responses, including 663 rows of Opus's own outputs. Third, we treat the scaffolds uniformly across models, using text-based parsing and identical prompts. Fourth, we make the entire code available.

We surfaced one confirmed apparatus--model mismatch (namely the Gemini parse failures; Section~\ref{sec:gemini-results}), but more subtle apparatus--model interactions may linger. If bias from such interactions were dominant, we would expect the improvement to be roughly uniform across configurations. Instead, we see that Opus is robust to evaluative-pressure scaffolds (multi-agent: $-$1.3 pp, ReAct: $-$2.0 pp; Table~\ref{tab:model-config}) but brittle to content-destroying ones like map-reduce (TruthfulQA: $-$24.7 pp; aggregate: $-$15.5 pp). This weakens the builder-as-subject concern, but does not fully eliminate it. Hence, we treat the robustness of Opus under evaluative pressure as hypothesis-generating until it is independently confirmed.

All qualitative conclusions except ReAct's and multi-agent's significance are preserved in a sensitivity analysis that excludes Opus entirely (Table~\ref{tab:opus-excluded}, N = 49,670).

\section{Implications for Evaluation Practice}

\label{sec:policy}

We have identified three gaps that must be closed for AI safety pre-deployment testing frameworks \citep{nistai8002026}, which focus on isolated models in one response format, to effectively prepare models for real-world, compound, and open-ended deployment.

\begin{tcolorbox}[coltitle=black,colback=orange!3, colframe=orange!40, title={%
Recommended Pre-Deployment Testing Mandates}]
\begin{enumerate}[leftmargin=*,nosep]
\item %
Mandate format-paired evaluations. Pre-deployment testing should cover the full range of deployment architectures that model developers and downstream deployers intend to use models for, not just worst-case scenarios. All safety evaluations must provide both MC and OE ratings for each benchmark; scores in one format alone are uninterpretable. Relative to OE, MC scores are 5 pp lower for bias safety on BBQ, 20 pp lower for sycophancy resistance, and 9 pp higher for capability on MMLU (Section~\ref{sec:format-dependence}). The direction of the gap differs between safety and capability, and its size is unpredictable from one format. Therefore, format-paired reporting of benchmarks is a prerequisite for meaningful comparisons of models, deployments, and times. The EU AI Act requires high-risk AI systems to achieve ``an appropriate level of accuracy, robustness and cybersecurity,'' \citep{euaiact2024} and we show that measured safety can differ by about 5--20 pp between formats.

\item %
Mandate structure-destroying scaffold testing for agentic deployments. Models intended to be used as agents must be evaluated under at least one structure-destroying delegation scaffold (e.g., a map-reduce), in addition to the direct API baseline. We have shown that while content-preserving scaffolds are surprisingly resilient (though not universally neutral), structure-destroying scaffolds can lead to safety loss that varies widely between six models (Section~\ref{sec:primary-analysis}). This is important because models evaluated only on the direct API interface cannot be certified as safe at the system level. While frontier AI responsible scaling plans, led by Anthropic's RSP \citep{anthropicrsp2026}, have pioneered using scaffolds to evaluate dangerous capabilities, our results show that the same concern should apply to proxied safety benchmarks, where vulnerability differs by an order of magnitude across models.

\item %
Mandate propagation verification for audits of agentic deployments. Audits of agentic deployments must empirically verify what percentage of safety-critical instructions make it to the terminal worker sub-calls, rather than assuming prompt adherence by the direct API. In our tracing, option markers reached only 0--4\% of map-worker inputs, while the reducer still received the original question and its options (Section~\ref{sec:option_preserving}).
\end{enumerate}
\end{tcolorbox}

These recommendations have downstream implications. E.g., the NIST AI 800-2 covers automated benchmark evaluations of models \citep{nistai8002026}. It would be natural to extend this document to cover the necessity of reporting both MC and OE results, at minimum, since format, a first-order variable, can reverse model rankings. As a basic diagnostic for all agentic evaluations, it may also be useful to employ both intention-to-treat and per-protocol scoring. The former would count all parsing failures as unsafe responses, and the latter would exclude them; paired together, they show how much of the reported degradation depends on counting parse failures (see Appendix~\ref{app:itt-pp} for an example).

We have also shown that compound systems pose additional problems for evaluations of this type because they can multiplex system prompts together in ways that can suppress the intended functionality of scaffolds (Appendix~\ref{app:sysprompt}).

First, the use of the direct API alone mischaracterises the safety of models that will be deployed in an agentic fashion, and evaluation using MC alone mischaracterises the safety of models that will be used in an open-ended fashion. Second, paired MC and OE administration estimates how far scores depend on the elicitation-and-scoring protocol (see Experiment 5 in Section~\ref{sec:format-dependence}).

\begin{tcolorbox}[coltitle=black,colback=gray!5, colframe=gray!50, title={%
Recommendations for Scaffold Safety Evaluation and Design}]
For evaluators: (1) Report and test model safety in paired MC and OE formats, with specification of the response format for all scores on safety benchmarks. (2) Test for propagation of task structure and response format cues from the scaffolding to each sub-call. (3) Test for preservation of task structure, to distinguish between format-dependent and format-independent degradation. (4) Test against direct-API baselines in paired MC and OE formats.

Frontier labs have set precedent for this approach by proposing the use of scaffolds to augment capability evaluations \citep{anthropicrsp2026}. The results here show that scores on safety benchmarks from direct API use are also non-transferable to scaffolded deployments, and thus the same approach should also be applied to standardised safety evaluations.

Thus, model card proxy scores for safety benchmarks (e.g. for biases, truthfulness, refusal calibration, etc.) should not be assumed to apply to scaffolded deployments, and should be evaluated using format-aware and configuration-specific assessment methods.

Given that standardised methods for evaluation of safety are currently being developed by multiple safety institutes \citep{aisinetwork2024}, it will be important to implement such format-based evaluation methods to ensure that comparative evaluations of models are standardised across nations in the International Network of AI Safety Institutes \citep{aisinetwork2024}.

The following recommendations are for those designing the scaffolding. (5) Consider the combination of intended model, prompt and scaffold as the target of evaluation. Our pilot experiments on neutral chains (N = 50) articulate the configurations we experimented with, but do not endorse multi-step chains as safe in general. (6) When either the architecture or review instructions are changed, test the combination of both. (7) If claiming a benefit for parallel review, serial review or adjudication, first benchmark all three on the same items. (8) When delegating, keep the MC options, and check the benefit with matched items, runs, and scoring. (9) Calibrate sycophancy mitigation on a per-model basis before deployment. Our direct baseline has a non-sycophancy rate of 33.8\%, and changes from $-$16.8 to +18.8 pp across models when using map-reduce.

Our recommendations are summarised in Table~\ref{tab:recommendations}, along with the sources of evidence supporting each one.
\end{tcolorbox}

\begin{table}[h]
\caption{Actionable recommendations and their supporting evidence are listed.}
\label{tab:recommendations}
\centering\small
\begin{tabular}{@{}p{3.2cm}p{3.8cm}p{4.2cm}c@{}}
\toprule
\textbf{Recommendation} & \textbf{Evidence Basis} & \textbf{Practice Gap} & \textbf{Cost} \\
\midrule
Format-paired reporting & about 5 to 20~pp format gaps (Sec.~\ref{sec:format-dependence}) & No benchmark requires dual format & Low \\
Structure-destroying scaffold test & NNH = 14 under map-reduce (Sec.~\ref{sec:primary-analysis}) & Proxy safety benchmarks not scaffold-tested & Medium \\
Propagation verification & Option-marker loss at map workers (Sec.~\ref{sec:propagation}) & No standard propagation audit & Low \\
NNH operational reporting & Enterprise risk communication (Sec.~\ref{sec:scaffold-discussion}) & Safety scores lack operational interpretation & Low \\
System prompt governance & Prompt competition can suppress scaffolds (Sec.~\ref{sec:prompt-competition}) & No API mechanism for safety-priority prompts & Low \\
\bottomrule
\end{tabular}
\end{table}

\label{sec:prompt-competition}

\begin{figure}[p]
  \centering
  \includegraphics[width=\textwidth]{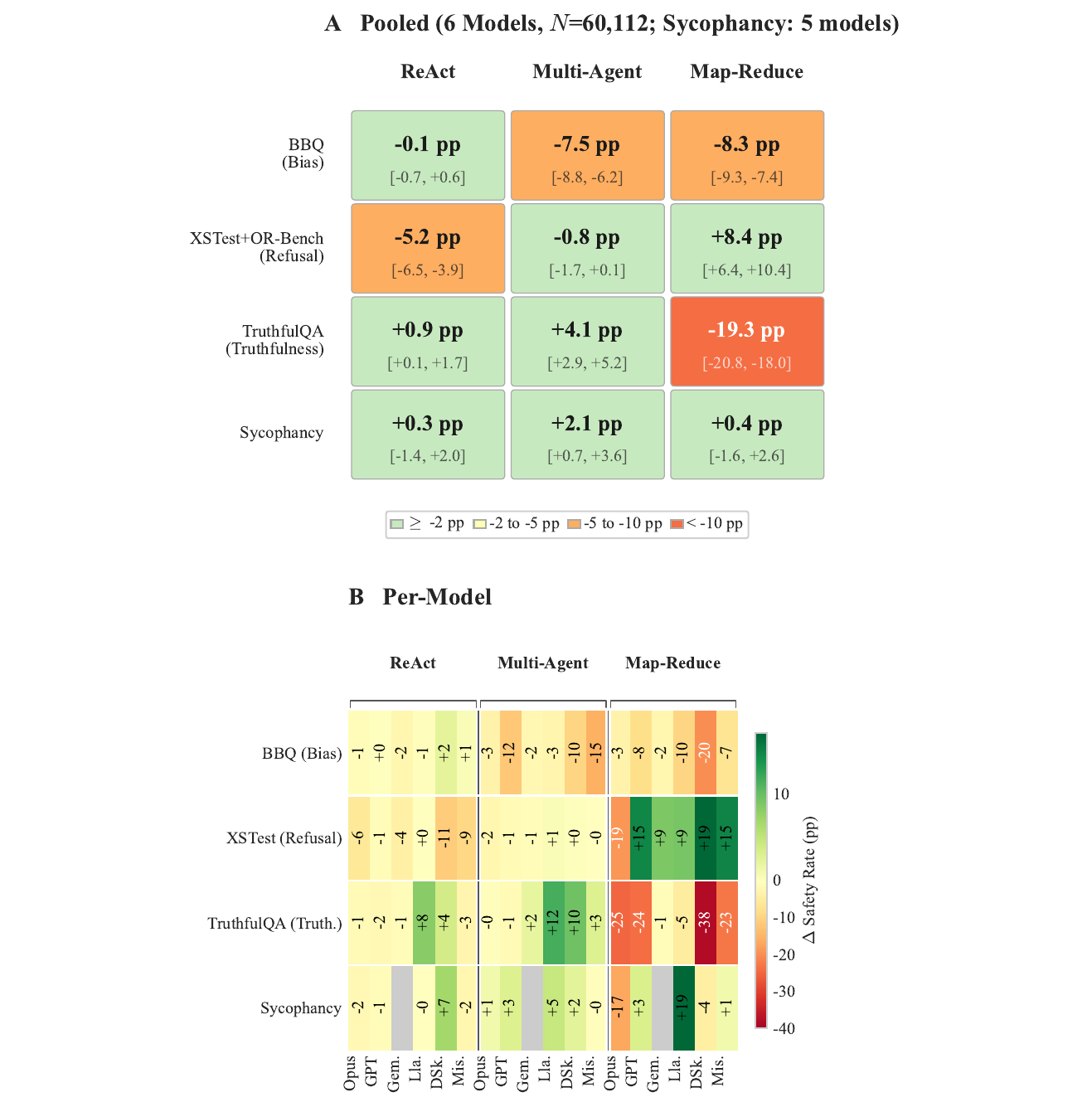}
  \caption{Configuration-Aware Safety Scorecard. \textbf{(A)}~Pooled change in safety ($\Delta$ percentage points relative to the direct baseline) for each benchmark and scaffold. Brackets indicate 95\% confidence intervals, estimated using a case-level cluster bootstrap with 2{,}000 replicates. Green indicates the point estimate for change in safety is $\geq -2$~pp (not a TOST equivalence finding), amber is a small safety degradation ($-2$~pp to $-5$~pp), red is a moderate ($-5$~pp to $-10$~pp) or large safety degradation ($< -10$~pp). \textbf{(B)}~Heatmap by model.}
  \label{fig:safety-scorecard}
\end{figure}

These results highlight that benchmark safety scores for LLMs cannot be understood without considering the configuration in which the model is deployed. An audit of academic LLM evaluations \citep{gringras2026frontierlag} found that 18.4\% of 4,766 full-text papers declare an explicit evaluation date, and suggested that scaffolding and tool access be reported. Current LLM APIs have only a single ``system prompt'' slot, which is used both for safety instructions and to configure the behaviour of scaffolds. There is no mechanism to designate instructions as safety-critical in a way that scaffolds cannot override. In practice, this can result in situations where safety instructions assume the model will be called in an MC format, but then the model is called in a different format during agentic scaffolds. The single slot can also mean that the model call contains a restrictive safety prompt from the benchmark. This suppressed the scaffold functionality of the RLM framework in 10/10 GPT-5.2 cases (Appendix~\ref{app:sysprompt}).

When interpreting LLM safety benchmarks scores, we propose that safety benchmark scores should always be presented alongside an NNH that reflects the anticipated deployment paradigm (e.g., the naive map-reduce deployment we tested in this study). For example, in our experiment, for every 14 cases in which the benchmark's instructions were routed by scaffolds, there was one additional benchmark failure. In other words, the pooled NNH for the naive map-reduce deployment is 14 relative to the direct API (for this particular mix of four benchmarks).

In addition to reporting benchmark scores and NNH, we introduce the Scaffold Safety Scorecard, which is a matrix of configurations by safety dimension (illustrated in Figure~\ref{fig:safety-scorecard} and described further in Appendix~\ref{app:scorecard}). The scorecard includes a table of NNHs for each configuration (which we recommend always being reported), as well as a stamp verifying that the researchers performed the methodology described herein.

The main effect of scaffold explains just 0.5\% of variance, whereas the benchmark explains 15.1\% of variance (33x more). The scaffold $\times$ benchmark interaction explains 1.5\% of variance, which is still less than the 1.9\% of variance explained by the model $\times$ benchmark interaction. The value of G (the generalizability coefficient) is 0.251 (95\% CI, bootstrapped: [0.000, 0.879]).

We release all code used in this evaluation as the framework ScaffoldSafety.

\FloatBarrier

\subsection*{AI Assistance Statement}

LLM tools provided substantial assistance with implementation of the evaluation pipeline, scaffold configurations and scoring infrastructure, statistical analysis, and manuscript preparation and revision. The author designed the study, specified the hypotheses and analysis plan, directed and critically reviewed the work, made the strategic and interpretive decisions, and takes responsibility for the manuscript. The builder-as-subject analysis in Section~\ref{sec:reflexivity} addresses the overlap between a pipeline-development model and the tested model set.

Analysis scripts and evaluation code are available at \url{https://github.com/davidgringras/safety-under-scaffolding}. Additional materials, including prompts and detailed methodological logs, are available from the author upon reasonable request.

\section*{Acknowledgments}
This work was conducted during the author's MPH program at Harvard T.H.\ Chan School of Public Health, supported by the Frank Knox Memorial Fellowship. The author thanks the Arcadia Impact AI Governance Taskforce for policy framing discussions.

\begin{sloppypar}The author received a BlueDot Impact Career Transition Grant during the revision of this manuscript; the funder had no role in study design, analysis, interpretation, or the decision to publish.\end{sloppypar}

\appendix
\addtocontents{toc}{\protect\bigskip\protect\noindent\textbf{Appendices}\protect\par\protect\medskip}

\section{Guide to Appendices}
\label{app:guide}

The following appendices support specific claims in the main text:

\begin{itemize}[leftmargin=*,nosep]
  \item \textbf{Appendix~\ref{app:sysprompt}}: System prompt competition traces (Section~\ref{sec:prompt-competition})
  \item \textbf{Appendix~\ref{app:dual-mechanism}}: Dual degradation mechanism details
  \item \textbf{Appendix~\ref{app:spec-curve}}: Full specification curve enumeration (Section~\ref{sec:spec-curve-results})
  \item \textbf{Appendix~\ref{app:production-frameworks}}: Production framework evaluation (Section~\ref{sec:production-frameworks})
  \item \textbf{Appendix~\ref{app:scorecard}}: ScaffoldSafety framework and scorecard (Section~\ref{sec:framework})
  \item \textbf{Appendix~\ref{app:flip-rate}}: Paired flip-rate analysis (Section~\ref{sec:self-awareness-robust})
  \item \textbf{Appendix~\ref{app:itt-pp}}: ITT vs.\ PP scorecard (Section~\ref{sec:gemini-results})
  \item \textbf{Appendix~\ref{app:ssi-nnh}}: Limits of recovery and residual NNH estimates
  \item \textbf{Appendix~\ref{app:api-constraints}}: Per-model API implementation constraints
  \item \textbf{Appendix~\ref{app:parse-artifacts}}: Measurement artifact case studies (Section~\ref{sec:scoring-sensitivity})
  \item \textbf{Appendix~\ref{app:phase1-pilot}}: Phase~1 exploratory probes: pilot data (Section~\ref{sec:semantic-invocation})
  \item \textbf{Appendix~\ref{app:exploratory}}: Exploratory analyses: full results (Section~\ref{sec:exploratory-mechanisms})
  \item \textbf{Appendix~\ref{app:validation}}: Format dependence validation inventory: 18 checks, including one canonical-data comparison outside the OE cohort (Section~\ref{sec:format-dependence})
  \item \textbf{Appendix~\ref{app:wald-cis}}: Independence-based Wald RD confidence intervals (Table~\ref{tab:confirmatory})
  \item \textbf{Appendix~\ref{app:detailed-methods}}: Detailed methods (Section~\ref{sec:methods})
  \item \textbf{Appendix~\ref{app:variance-decomposition}}: Factorial variance decomposition and generalizability analysis (Section~\ref{sec:policy})
\end{itemize}

\section{System Prompt Competition Traces}
\label{app:sysprompt}

The feasibility test behind the system prompt competition finding (Section~\ref{sec:prompt-competition}), with both conditions and their results, is in the ancillary file \texttt{prompt\_traces.md}, listed on the arXiv abstract page.

\section{Dual Degradation Mechanism Details}
\label{app:dual-mechanism}

In our pilot experiments, we found that scaffolding changes safety-relevant outputs on TruthfulQA~MC1 by operating through two different mechanisms, motivating our pre-registered experimental design. \textbf{Map-Reduce} reduces model performance via \emph{content loss}. The decomposition removes the answer choices~(A)--(F), and DeepSeek only achieves 33.3\% accuracy with short context ($n{=}57$) compared to 69.7\% with long context ($n{=}66$). By contrast, \textbf{RLM} largely preserves content: 87.9\% of DeepSeek sub-calls contain the MC options, but it changes the system prompt to a REPL instruction. RLM achieves 80.0\% on DeepSeek ($n{=}5$) and 65.0\% on GPT-5.2 ($n{=}20$). Table~\ref{tab:dual-mechanism} summarises these findings on content vs.\ safety-framing loss. These results motivated our two-mechanism framework.

\begin{table}[t]
\centering
\caption{Pilot-stage dual-mechanism comparison: Map-Reduce vs.\ RLM on TruthfulQA~MC1.}
\label{tab:dual-mechanism}
\small
\begin{tabular}{@{}lll@{}}
\toprule
\textbf{Dimension} & \textbf{Map-Reduce} & \textbf{RLM} \\
\midrule
MC choices in sub-calls    & 0\%     & 77.9--87.9\% (context var.)   \\
System prompt propagation  & Yes     & 0\%        \\
Content transformation     & Abstract sub-questions         & Full context via variable ref. \\
Sub-call count             & 3--5 (fixed pipeline)          & 3--12 (model-decided)         \\
\midrule
\multicolumn{3}{@{}l}{\textit{DeepSeek accuracy on TruthfulQA MC1}} \\
\quad Short context        & 33.3\% ($n{=}57$)              & ---                           \\
\quad Long context         & 69.7\% ($n{=}66$)              & 80.0\% ($n{=}5$, long only)   \\
\midrule
\multicolumn{3}{@{}l}{\textit{GPT-5.2 accuracy on TruthfulQA MC1}} \\
\quad RLM only             & ---                            & 65.0\% ($n{=}20$)             \\
\midrule
Primary degradation        & Content loss      & Safety-framing loss  \\
Failure mode               & Answers without MC options     & Processes correctly when MC   \\
                           &                                & preserved; fails on condens.  \\
\bottomrule
\end{tabular}
\end{table}

\begin{table}[t]
\centering
\caption{Pilot-stage analysis of DeepSeek condensation under RLM. We find that the condensed sub-calls (less than 5{,}000 characters) lose more MC choices, which motivated our use of the content-preservation metric in our full experiment.}
\label{tab:condensation}
\small
\begin{tabular}{@{}lcc@{}}
\toprule
& \textbf{Full context} & \textbf{Condensed} \\
& ($\geq$5{,}000 chars)      & ($<$5{,}000 chars) \\
\midrule
Sub-call count                  & 28 (84.8\%)         & 5 (15.2\%)          \\
MC choices preserved            & 27/28 (96.4\%)      & 2/5 (40.0\%)        \\
Mean prompt length (chars)      & $\sim$15{,}400      & $\sim$1{,}800       \\
Typical prefix category         & extraction           & refinement/summary  \\
Cases with errors               & 0/3 cases           & 1/2 cases affected  \\
\bottomrule
\end{tabular}
\end{table}

\section{Specification Curve Enumeration}
\label{app:spec-curve}

The ancillary file \texttt{spec\_curve\_paths.csv}, listed on the arXiv abstract page, shows our 29 planned analytic dimensions by category: scoring (six, from judge model to response truncation length), statistical model, data, sanitization, and configuration. Our primary specification curve has 18 specifications varying the benchmark inclusion, model inclusion, and parse-failure handling, and our separate exploratory analysis varies nine analytic choices in 384 specifications, but both cover only part of the planned decision space.

\section{Production Framework Evaluation}
\label{app:production-frameworks}

This content has been moved into the main text (Section~\ref{sec:production-frameworks}).

\section{ScaffoldSafety Framework and Scorecard}
\label{app:scorecard}

\subsection{ScaffoldSafety: An Open Evaluation Framework}
\label{sec:framework}

We release \textsc{ScaffoldSafety}, an open-source Python framework implementing the full evaluation pipeline for reproduction on new models, scaffolds, and benchmarks.

\paragraph{Design.}
The code for \textsc{ScaffoldSafety} consists of five major components: (1)~benchmark loaders, (2)~scaffold configurations for the four patterns (Direct API, ReAct, Multi-Agent, and Map-Reduce), (3)~the tiered scoring system, (4)~the protocol for blinding assessors, and (5)~the statistical analysis.

We provide abstract base classes to enable extending the framework with new scaffold configurations and benchmarks. Our code can be found at \url{https://github.com/davidgringras/safety-under-scaffolding} in the directory \texttt{scaffold\_safety/}.

\subsection{The Scaffold Safety Scorecard}
\label{sec:scorecard}

We propose the \emph{Scaffold Safety Scorecard} as a standardised reporting format for model cards. It includes three components. By design, it does not include a composite robustness index. This design choice is informed by the factorial variance decomposition (see Appendix~\ref{app:variance-decomposition}): the percentage of variance explained by scaffold effects is only 0.5\%, whereas scaffold-benchmark interaction effects are 3.2$\times$ larger (1.5\%). In a generalizability analysis ($G = 0.251$, bootstrap 95\% confidence interval $[0.000, 0.879]$), the reliability of a composite metric given our mixture of four benchmarks is not statistically distinguishable from either zero or 0.88~\citep{brennan2001generalizability}. Today, a single composite score is not a defensible basis for a deployment go/no-go decision.

\paragraph{(1) Safety Rate Matrix.}
The main contribution of the scorecard is a matrix of safety rates. This generalizes the idea of a single-number safety score to multiple dimensions of safety and deployment configurations. The rows of the matrix are the different safety dimensions; the columns represent different deployment configurations. A safety rate is the proportion of cases from a benchmark to which a model responded safely. This allows us to move from a single safety number to the full surface of safety benchmark~$\times$~deployment configuration. It can reveal interactions that would be obscured by a single, pooled number. For example, in our evaluation, the TruthfulQA safety rate was 1~pp to 38~pp lower under map-reduce, depending on the model.

\paragraph{(2) Number Needed to Harm (NNH).}
For each reported benchmark score on a safety test, the score should be accompanied by the NNH for the deployment configuration to which it pertains. NNH is computed as $1 / |p_{\text{scaffold}} - p_{\text{direct}}|$ and has a natural interpretation in the case of a degradation in safety: it is the number of cases that must be processed by the scaffold (compared to the direct API baseline) to observe one additional unsafe response. An NNH of 14 means that every fourteenth query, on average, results in an extra error. The NNH has the advantage of lacking any arbitrary scaling factor, and being comprehensible to operators without domain-specific training.

\paragraph{(3) Methodology Stamp.}
To enable users to assess the methodological rigour of the scores reported, every scorecard is accompanied by a verification stamp that records whether the evaluation was pre-registered, whether assessors were blinded (as measured by Bang's Blinding Index), whether results were cross-validated (as measured by Cohen's $\kappa$), and whether a specification curve analysis was performed.

\paragraph{Adoption pathway.}
The scorecard adds the dimension of deployment configuration, which is absent from model cards. We consider three stages in the adoption of the scorecard. The first stage, which we perform in this paper, is to demonstrate it. The second stage, which is yet to occur, is community adoption via our open-source package. The third stage, which is also yet to occur, is for AI labs and AI safety organisations such as AISI and METR to incorporate the scorecard into model-card templates.

\section{Paired Flip-Rate Analysis}
\label{app:flip-rate}

The figure presenting the paired flip rates has been incorporated into the main body of the paper as Figure~\ref{fig:flip-rate}, in Section~\ref{sec:self-awareness-robust}.

\section{ITT vs.\ PP Scorecard}
\label{app:itt-pp}

See the MC-format ITT vs.\ PP scorecard for our primary TruthfulQA, BBQ, and sycophancy cells in Section~\ref{sec:heterogeneity} (Table~\ref{tab:itt-pp}). Under ITT, parse failures are counted as unsafe, whereas under PP they are ignored and results reported are on the selected subset of responses that parsed successfully. For Gemini under ITT, the difference between map-reduce and direct is $-1.6$~pp, and the difference between ReAct and direct is $-1.6$~pp. Under PP, these differences are $-1.2$~pp and $+0.6$~pp. This shows that the difference is sensitive to whether parse failures are included in the evaluation. However, it does not isolate the effects of the underlying model alignment, and cannot be said to establish a causal mediation effect.

\section{Limits of Recovery and Residual NNH Estimates}
\label{app:ssi-nnh}

\paragraph{Recovery estimation.}
Recovery percentages are measured from the last run. There are 200 cases per model. Each case is paired with its own two primary records: a direct record and a map-reduce record. A single extractor scores the records for all cases (Table~\ref{tab:option_preserving}; Section~\ref{sec:option_preserving}). No residual NNH is given.

\paragraph{Relation to the primary NNH.}
The pooled primary NNH of 14, comparing map-reduce with direct across the primary cohort's four benchmarks, cannot be decomposed into format effects and reasoning effects because the matched option-preserving comparison (Table~\ref{tab:option_preserving}) only spans BBQ and TruthfulQA.

\section{Per-Model API Implementation Constraints}
\label{app:api-constraints}

See \texttt{api\_constraints.md} in the paper's ancillary files (accessible from the arXiv abstract page) for details on the API settings for each model, as well as a discussion of cross-model comparability and operational constraints.

\section{Measurement Artifact Case Studies}
\label{app:parse-artifacts}

During testing of the production frameworks (Section~\ref{sec:production-frameworks}), we encountered a bug in output parsing in the scoring code for the production-framework evaluator. The code used a fallback regular expression \verb|\b([A-Z])\b| to extract the model's answer. This inadvertently matched the pronoun ``I'' when the model's output began with long text such as ``I can determine that\ldots'' rather than a single letter. We found that outputs with long text triggered these false matches to ``I,'' while outputs with single characters were parsed correctly. The OpenAI Agents SDK template tended to elicit long text responses (765~characters on average), while the CrewAI template tended to elicit terser responses (average length~1), leading to an apparent 48 percentage point score gap between CrewAI and OpenAI Agents SDK on AI Factual Recall. After fixing the bug by restricting the fallback expression to the set of valid letters ($\{A, B, C, D, E\}$), the gap decreased from 48~pp to 15~pp. The remaining gap was still due to response formatting effects.

The specification curve did not detect this artifact, which we discovered by examining paired raw outputs. Our primary pipeline was protected from this bug because our answer extractor explicitly checked extracted letters against \path{valid_letters}, the set of options for each benchmark ($\{A, B\}$ for AI Factual Recall). The scorer used for production frameworks, written at a later point and under somewhat more time pressure, did not feature any such safeguard. Using only a single regex introduced errors of 32--41 percentage points, large enough to distort the ranking of the frameworks on this benchmark. This is an example of the ease with which measurement bugs can be introduced into evaluation code, even under normal practitioner conditions. First, this example demonstrates that answer extraction affects safety scores at least as much as model outputs. Second, it highlights the importance of adversarially evaluating evaluation pipelines in the same way we do the AI systems themselves. We keep answer extraction fixed to our pre-registered last-answer extraction method because any change to the answer-extraction regexes would alter the pre-registered scoring contract.

Upon examining raw outputs, we also discovered a second artifact. Across all four models and all three frameworks, the ``unknown'' option was chosen 0\% of the time, which at first looked like a dramatic amplification of bias induced by the production frameworks. However, it was actually due to a bug in data formatting. In our BBQ bias benchmark data, answer choices are stored in a field separate from the question text. Our primary pipeline formats these choices into the prompt (e.g., ``(A)~Robert (B)~Not answerable (C)~Margaret''), but the production framework dataloader did not. As a result, models were prompted with open-ended questions without multiple-choice options, meaning that they could not select the ``unknown'' option (which is how BBQ measures epistemic humility).

These examples demonstrate how measurement artifacts can give rise to findings that are incorrect but \emph{directionally compelling}. Both had plausible mechanistic explanations and would have withstood scrutiny by anyone not inspecting raw prompts and individual model responses. We only detected them by reading the exact text presented to models and comparing effective sample sizes after parse failures.

\section{Phase~1 Exploratory Probes: Pilot Data}
\label{app:phase1-pilot}

The Phase~1 pilot data behind Section~\ref{sec:exploratory-mechanisms} (DeepSeek~V3.2, Opus~4.6, and GPT-5.2; $N = 50$ per condition) are in the ancillary file \texttt{phase1\_pilot.md} and its two figure files, listed on the arXiv abstract page.

\section{Exploratory Analyses: Full Results}
\label{app:exploratory}

The ancillary file \texttt{exploratory\_results.md}, available on the arXiv abstract page, describes all eleven tables of exploratory results, their notes, and the persona-feature analysis. These results appear in the CSV files listed in that file.

All results reported are exploratory. In Phase~1 dose pilots ($N = 50$ per condition), DeepSeek accuracy on BBQ decreased monotonically from 94.0\% to 82.0\% from passthrough to aggressive bias invocation, and accuracy on TruthfulQA increased from 74.0\% to 92.0\% from passthrough to aggressive misconception checking. For aggressive doses, accuracy on BBQ changed by $+4.0$~pp for Opus, $-12.0$~pp for DeepSeek, and $-22.0$~pp for GPT-5.2. On the subset of 50 items from the BBQ dataset used for the CrewAI comparison on DeepSeek V3.2, the three-agent system attained a 48/50 (96.0\%) accuracy, compared to a 42/50 (84.0\%) accuracy for the two-agent system, corresponding to a $+12.0$ percentage point improvement. Across the full $N = 150$ sample, the three- and two-agent systems attained accuracies of 88.7\% and 80.7\%, respectively ($+8.0$~pp). The Adjudicator Agent successfully `rescued' 26 of the 29 (89.7\%) overcorrections made by the Bias Checking Agent, all of which were for disambiguated items, and did not introduce any new errors. Note, however, that we cannot be sure that the improvements in the three-agent accuracy are due to the rescue operation. For the persona analysis, we used 2{,}483 model--item pairs, corresponding to the same 500 cases under map-reduce for each of the five models (483 for GPT-5.2). ITT sycophancy was 60.3\% (924/1{,}533) when no adversarial persona features were detected, 70.8\% (494/698) when one was detected, 86.5\% (186/215) when two were detected, and 78.4\% (29/37) when three were detected (85.3\% (215/252) when ${\geq}2$ were detected). In a joint logistic regression model using non-robust maximum likelihood standard errors, we found that each additional adversarial persona feature increased the odds of ITT sycophancy (OR 1.67, $p = 1.01 \times 10^{-11}$). Neither contextual features (OR 1.06, $p = 0.546$) nor name detection (OR 1.05, $p = 0.697$) were significantly associated with ITT sycophancy in the same model. We also found that political leaning had the largest effect size in separate models (OR 3.21, $p = 2.99 \times 10^{-15}$). The ratio of sub-call refusal rate to final refusal rate, which we term the distributed safety score, was 0.661 for DeepSeek, 0.784 for Opus, and 1.285 for GPT-5.2 on clear-cut questions (B1), but 0.059--0.104 on boundary questions (B2). LLM judge rescoring reduced heuristic sub-call leakage to 4.4\% (18/407), down from 39.3\% (160/407).

\section{Format Dependence Validation Protocol}
\label{app:validation}

This appendix summarises the 18 format-dependence checks we executed (six for scoring, three for evasion, two for MMLU, and seven for the pipeline). The follow-up checks (Tests~1a--1e and~2a--2c) are exploratory in nature and their decision rules are not established as preregistered. Tests~1a--1e represent unblinded audits generated by a model (according to the saved source, by an Opus~4.6 agent), without independent human adjudication. The scoring audit asserts that the criteria were set before any samples were reviewed, but no timestamped independent evidence of this is present in the reviewed records. The saved evasion report reinterprets the results of the length and content classifications post hoc.

The checks are listed in the ancillary file \texttt{validation\_protocol.md}, available on the arXiv abstract page. The file contains, for each test, the threat model, saved evidence, limitations, and historical decision rules. It also records the historical verdict for Tests~1 and~2, and a scoped status for Tests~3 and~4, but no overall pass/fail counts for OE tests. Separately from the OE experiment, in Test~1f, GPT-5.2 rescored 200 canonical responses from the primary dataset, and achieved an agreement of 179/198 (90\%, $\kappa = 0.790$); this result does not mean that OE validation passed. Benchmark-specific scoring error in the OE experiment is not resolved.

\FloatBarrier

\section{Factorial Variance Decomposition and Generalizability Analysis}
\label{app:variance-decomposition}

We use a $6 \times 4 \times 4$ design ($N = 60{,}112$). There are empty cells in the Gemini $\times$ sycophancy condition. We now assess the contribution of each experimental factor to our results. Due to the empty cells, the cell-means decomposition is rank-deficient, so we use the Type II sums of squares from nested models. We report the effect sizes for all main effects and two-way interactions, in terms of both eta-squared ($\eta^2$) and bias-corrected omega-squared ($\omega^2$) in Table~\ref{tab:variance-decomposition}.

\begin{table}[h]
\caption{This table shows a factorial variance decomposition for binary safety outcomes ($N = 60{,}112$), with sources ordered by $\eta^2$. The table only includes main effects and two-way interactions. All effects are significant at $p < 10^{-6}$. Because the three-way interaction term (model $\times$ scaffold $\times$ benchmark) is contained in the residual term, the residual term overstates the pure within-cell variance. Adding the three-way interaction term to the decomposition shifts approximately 0.8~pp of $\eta^2_{\text{residual}}$ into a new model $\times$ scaffold $\times$ benchmark term and leaves the main effect of scaffold negligible. The qualitative conclusions are the same: the choice of benchmark is the dominant source of variance, the main effect of the scaffold is small, and the interaction between scaffold and benchmark is larger (3.2$\times$) than the main effect of the scaffold. Ratios quoted in the paper use unrounded $\eta^2$ (equivalently $F \times$ df).}
\label{tab:variance-decomposition}
\centering
\small
\begin{tabular}{@{}lrrccc@{}}
\toprule
\textbf{Source} & \textbf{df} & \textbf{$F$} & \textbf{$\eta^2$ (\%)} & \textbf{$\omega^2$ (\%)} & \textbf{Magnitude} \\
\midrule
Benchmark                   & 3 & 3{,}904 & 15.1 & 15.1 & Large \\
Model $\times$ benchmark    & 14 & 104 & 1.9 & 1.9 & Small \\
Model                       & 5 & 234 & 1.5 & 1.5 & Small \\
Scaffold $\times$ benchmark & 9 & 126 & 1.5 & 1.5 & Small \\
Model $\times$ scaffold     & 15 & 25 & 0.5 & 0.5 & Negligible \\
Scaffold                    & 3 & 119 & 0.5 & 0.5 & Negligible \\
\midrule
Residual (within-cell)      & 60{,}062 &       & 77.6 &      & \\
\bottomrule
\end{tabular}
\end{table}

\paragraph{Key findings.}
We see that the main effect of the benchmark (i.e. the safety property being tested) has an effect size ($\eta^2 = 15.1$\%) that is $\sim$33$\times$ larger than the effect of the scaffold architecture being used ($\eta^2 = 0.5$\%). The interaction between scaffold and benchmark ($\eta^2 = 1.5$\%) is roughly 3.2 times the size of the main effect of scaffold: the effect of a scaffold depends on the benchmark. Map-reduce, for example, has a negative effect on TruthfulQA of $\sim$19~pp on average, but a positive effect on XSTest of $\sim$8~pp (Figure~\ref{fig:safety-scorecard}). The interaction between model and benchmark ($\eta^2 = 1.9$\%) is the second-largest systematic effect: models' rankings by safety depend on what benchmark is being used. Thus, any single composite metric that integrates both benchmarks will also have rankings that depend on the benchmark.

\paragraph{Policy interpretation.}
While the main effect of the scaffold is small (0.5\%, in terms of $\eta^2$, the proportion of variance in the outcome explained by the type of scaffold used), this does not mean that scaffolds are safe. In a single condition (model-benchmark pair), the effect of the scaffold on safety can be as large as 47.5~pp (DeepSeek $\times$ TruthfulQA). The important quantity for a particular deployment is the conditional expectation (e.g. NNH of 14 under map-reduce). For policy audiences, the variance decomposition shows that average effects of scaffolds on safety can be substantially uninformative, because the effects of scaffolds on safety appear largely unpredictable from the average effects. This means that reporting per-model, per-benchmark results is a necessary statistical consequence of the generalizability interval (Appendix~\ref{app:gtheory}), and any responsible evaluation of model deployments should include them.

\subsection{Generalizability Analysis}
\label{app:gtheory}

We further conduct a Generalizability Theory (G-theory) analysis~\citep{cronbach1972dependability, brennan2001generalizability}. The models (the object of measurement) are a random facet. The scaffolds and benchmarks ($n_I = n_J = 4$) are also random facets because we want to know whether a model's position would remain if we changed the scaffold or the benchmark. After the exclusions, we have exactly one missing model $\times$ benchmark cell (Gemini 3 Pro $\times$ sycophancy). We excluded that model from the complete design, leaving us with $n_p = 5$. We calculate variance components using a cell-means approach based on the 80 cell-level proportions following Brennan's expected mean square equations for this design's facets~\citep{brennan2001generalizability}.

From this procedure, we calculate a generalizability coefficient of $G = 0.251$ (bootstrap 95\% CI: $[0.000, 0.879]$; $B = 10{,}000$ resamples of models with replacement). The model true score variance estimate ($\hat{\sigma}^2_p$) is positive (approximately $0.00069$), largely driven by the model $\times$ benchmark interaction term ($\hat{\sigma}^2_{pJ}$), which accounts for approximately 58.8\% of the total random variance. A decision (D-)study extrapolating to a hypothetical design with 8 scaffolds $\times$ 12 benchmarks results in $G = 0.51$. In 18.4\% of bootstrap samples, $G$ was greater than 0.70. The bootstrap upper bound of $0.879$ is specific to the particular mix of four benchmarks. In all three sensitivity analyses, the interval contains zero and the point estimate is below 0.70: $G = 0.58$ $[0.00, 0.87]$ when dropping the sycophancy benchmark; $G = 0.00$ $[0.00, 0.75]$ on the data prior to exclusions; and $G = 0.37$ $[0.00, 0.89]$ when fitting all valid cells using restricted maximum likelihood (REML), which can handle the missing cell for Gemini 3 Pro $\times$ sycophancy (6 models; 92 of 96 model $\times$ scaffold $\times$ benchmark cells).

Because the confidence interval $[0.000, 0.879]$ spans from ``of little use'' to ``very good'' reliability, we cannot distinguish the composite reliability on this mix of benchmarks from zero or 0.88. The wide interval does not support using a single composite safety index as an input into deployment decisions: while we cannot prove the reliability is zero, our data cannot adjudicate the suitability of aggregation for informing deployment. The finding that $G = 0.25$ turns our principled objection to composite indices (Appendix~\ref{sec:scorecard}) into an empirical constraint (with the upper-bound of the CI limiting the strength of this claim).

\subsection{Per-Model Scaffold Sensitivity Profiles}
\label{app:model-sensitivity}

\begin{sloppypar}
Table~\ref{tab:model-sensitivity} presents the scaffold sensitivity for each model, sorted by average safety-rate range across benchmark$\times$\allowbreak scaffold cells.
\end{sloppypar}

\begin{table}[h]
\caption{Sensitivity of models to the choice of scaffolding. Range (max $-$ min) of safety rates across the four scaffold configurations within each benchmark, averaged across benchmarks. $\eta^2_{\text{scaffold}}$ is the fraction of within-model variance explained by scaffold.}
\label{tab:model-sensitivity}
\centering
\small
\begin{tabular}{@{}lcccc@{}}
\toprule
\textbf{Model} & \textbf{$\eta^2_{\text{scaffold}}$ (\%)} & \textbf{Avg range (pp)} & \textbf{Max range (pp)} & \textbf{Overall safe (\%)} \\
\midrule
DeepSeek V3.2   & 2.1 & 27.7 & 47.5 & 66.8 \\
Mistral Large 2 & 0.2  & 17.3 & 25.6 & 69.7 \\
Opus 4.6        & 2.5  & 16.2 & 24.7 & 80.5 \\
GPT-5.2         & 0.4 & 14.0 & 23.9 & 70.5 \\
Llama 4 Maverick & 0.1 & 13.6 & 19.0 & 68.8 \\
Gemini 3 Pro    & 0.1 & 6.1 & 12.8 & 89.2 \\
\bottomrule
\end{tabular}
\end{table}

The most sensitive model, DeepSeek (27.7~pp average range), has 4.6$\times$ the range of the most robust model, Gemini (6.1~pp; three benchmarks). This heterogeneity in model-by-model results shows that robustness to scaffolds is a model$\times$scaffold interaction. It is not a fixed property of a given scaffold architecture. For this reason, the scorecard reports results for each model separately (see Section~\ref{sec:scorecard}); if we aggregated results over models, we would miss this heterogeneity.

\section{Independence-Based Wald RD Confidence Intervals}
\label{app:wald-cis}

\begin{sloppypar}
Table~\ref{tab:wald-cis} gives Wald confidence intervals for the risk difference, assuming that the two proportions are independent. This analysis is naive, because it ignores the covariance between the two proportions within a case. This is why the confidence intervals are wider than the case-cluster bootstrap confidence intervals in the primary analysis (Table~\ref{tab:confirmatory}). The standard error for the risk difference is: $\text{SE}(\text{RD}) = \sqrt{p_d(1-p_d)/n_d + p_s(1-p_s)/n_s}$.
\end{sloppypar}

\begin{table}[h]
\caption{These Wald CIs for the RD assume independence and are presented as a conservative sensitivity check.}
\label{tab:wald-cis}
\centering
\setlength{\tabcolsep}{5.8pt}
\begin{tabular}{@{}llcccc@{}}
\toprule
\textbf{Hyp.} & \textbf{Comparison} & \textbf{RD (pp)} & \textbf{Wald 95\% CI} & \textbf{Wald 90\% CI} & \textbf{Bootstrap 95\% CI} \\
\midrule
H1a & ReAct vs.\ direct       & $-0.8$ & [$-1.7$, $+0.2$] & [$-1.6$, $+0.1$] & [$-1.2$, $-0.2$] \\
H1b & Multi-agent vs.\ direct & $-0.8$ & [$-1.8$, $+0.1$] & [$-1.7$, $-0.03$] & [$-1.5$, $-0.2$] \\
H1c & Map-reduce vs.\ direct  & $-7.3$  & [$-8.3$, $-6.3$] & [$-8.2$, $-6.5$] & [$-8.1$, $-6.4$] \\
\bottomrule
\end{tabular}
\end{table}

\section{Detailed Methods}
\label{app:detailed-methods}

Material duplicating Section~\ref{sec:methods} or the OSF registration is included in an ancillary file (\texttt{detailed\_\allowbreak methods.md}, as listed on the arXiv abstract page). This includes the scoring table, the blinding steps, the registered model, secondary analyses, test families, and registered elements.

\subsection{Scoring Protocol Details}

\begin{sloppypar}
\paragraph{Last-answer extraction.}
We took the final valid answer marker in each model response and not the first one. This is because while most models simply answer ``Answer: (A)'', Llama~4 tends to provide verbose chain-of-thought reasoning before giving a final answer, even when it is not being scaffolded. We applied this ``last valid answer marker'' parsing rule to all multiple choice benchmarks, but only Llama~4 was materially affected. It prevented intermediate reasoning steps from being confused for the model's final answer, and increased Llama~4's BBQ passthrough accuracy from 76.7\% to 93.0\%.
\end{sloppypar}

\subsection{Statistical Analysis Details}

\subsubsection{Primary Model Specification}

\paragraph{Protocol adaptation (D-006).} We implement a logistic regression with configuration, model, and benchmark fixed effects and case-clustered sandwich standard errors~\citep{cameron2015practitioner} in place of our pre-registered case-random-intercept GLMM (see Table~\ref{tab:deviations}). We do not fit a matched GLMM, and thus cannot compare coefficient point estimates and standard errors. We assess configuration-by-model and configuration-by-benchmark interactions with case-cluster-robust Wald tests (registered as LRTs). Our specification curves are sensitivity analyses within the context of the logistic regression.

\subsubsection{Equivalence Testing}

We apply TOST~\cite{schuirmann1987tost, lakens2017equivalence} to every primary scaffold comparison, using an equivalence margin of $\Delta = 2$~pp. 2~pp is the smallest difference that we think could plausibly affect a deployment decision; it is also smaller than the performance spread between models for any of the benchmarks in this paper. Sensitivity margins are 1, 3, and 5~pp. Statistical significance and practical equivalence are orthogonal, and we report labels for both; e.g., ReAct is statistically significant ($p_{\text{Holm}} = 0.017$) while also being TOST-equivalent at $\pm 2$~pp. When the 90\% CI of the risk difference lies within $(-0.02, +0.02)$, we conclude equivalence. Because a fixed equivalence margin has different implications for different baseline values (2~pp is different when the baseline is 98\% vs.\ 70\%), we report equivalence results for each benchmark in Table~\ref{tab:tost-bench}. Across benchmarks, 4 out of 12 comparisons are statistically equivalent: ReAct for BBQ, TruthfulQA, and sycophancy; and multi-agent for XSTest/OR-Bench.

\begin{table}[h]
\caption{Per-benchmark TOST at $\Delta = 2$~pp: risk difference (pp) over model--case pairs present in both arms, with 90\% case-cluster bootstrap CI ($B = 2{,}000$). $^{\dagger}$CI inside $(-2, +2)$: equivalent.}
\label{tab:tost-bench}
\centering
\small
\begin{tabular}{@{}lccc@{}}
\toprule
\textbf{Benchmark} & \textbf{ReAct} & \textbf{Multi-agent} & \textbf{Map-reduce} \\
\midrule
BBQ & $-0.21$ [$-0.72$, $+0.31$]$^{\dagger}$ & $-7.55$ [$-8.58$, $-6.48$] & $-8.31$ [$-9.16$, $-7.52$] \\
Sycophancy & $+0.28$ [$-1.13$, $+1.70$]$^{\dagger}$ & $+2.06$ [$+0.85$, $+3.31$] & $+0.24$ [$-1.54$, $+2.07$] \\
TruthfulQA & $+0.63$ [$-0.06$, $+1.29$]$^{\dagger}$ & $+4.13$ [$+3.20$, $+5.12$] & $-19.35$ [$-20.54$, $-18.14$] \\
XSTest/OR-Bench & $-5.24$ [$-6.39$, $-4.13$] & $-0.48$ [$-1.24$, $+0.28$]$^{\dagger}$ & $+6.68$ [$+5.04$, $+8.35$] \\
\bottomrule
\end{tabular}
\end{table}

\subsubsection{Specification Curve Analysis}

See Appendix~\ref{app:spec-curve} for 29 planned analytic dimensions in five categories. The primary specification curve (Figure~\ref{fig:spec-curve}) consists of 18 different combinations of analytic choices: (a)~benchmark inclusion (all benchmarks, all benchmarks but the sycophancy benchmark, all benchmarks but XSTest and OR-Bench), (b)~model inclusion (all models, all models but Opus, all models but DeepSeek), and (c)~parse failure inclusion (keeping and scoring parse failures as unsafe, excluding parse failures). The specifications are logistic regressions estimating the contrast between a particular scaffold and direct prompting with model and benchmark fixed effects. Standard errors are case-cluster sandwich estimates with finite-sample correction, and confidence intervals (CIs) and two-sided Wald p-values are based on a normal reference distribution.

In the first plot of the specification curve, all 54 contrasts had non-robust (standard) covariance. Figure~\ref{fig:spec-curve} shows the plot with the corrected case-cluster covariance. It has the same point estimates and the same outcome labels and inclusion criteria as the original plot. Compared to the original plot, thirteen pointwise decisions are changed from not significant to significant, eleven of which are decreases in safety odds of ReAct or multi-agent, and two of which are increases for ReAct. No decisions are changed from significant to not significant. Here we report the primary specification curve in terms of odds ratios, with pointwise 95\% confidence intervals, along with the median, interquartile range, and number of nominally significant results. We do not apply multiplicity correction across the specification grid. There is no fitted GLMM subset in the saved analysis. We separately report the ``exploratory'' specification curve, corrected the same way, which spans nine analytic choice dimensions and five models in 384 different specifications. It does not cover all 29 analytic dimensions on the menu.

\subsubsection{Effect Size Reporting}

\paragraph{Risk difference computation.} Risk differences (RDs) are unadjusted marginal effects, while odds ratios (ORs) are adjusted conditional effects from the logistic regression in Equation~\ref{eq:primary} (with model and benchmark fixed effects). We report both: the risk difference as an unadjusted effect size relevant for practical deployment of scaffolds, and the odds ratio as an adjusted effect size for testing statistical significance. We use a case-cluster bootstrap, respecting the repeated-measures design, to construct confidence intervals for the risk differences ($B = 2{,}000$ resamples on $k = 2{,}617$ benchmark cases, random seed~42). Each bootstrap sample gives a risk difference (scaffold rate minus direct rate), and we use the 2.5th and 97.5th percentiles of the bootstrap distribution as the lower and upper bounds of the 95\% confidence intervals (and the 5th and 95th percentiles for the 90\% confidence intervals used in our TOST tests). As a robustness check, we also compute Wald confidence intervals for the risk differences under an independence assumption (reported in Appendix~\ref{app:wald-cis}). These are conservative (because they do not account for the repeated-measures design) and wider than our bootstrap intervals. In TOST equivalence testing~\cite{schuirmann1987tost}, equivalence is established if the bootstrap 90\% CI lies entirely within $(-\Delta, +\Delta)$. The NNH is computed directly from the unadjusted RDs.

\subsection{Pre-Registration Details}

\begin{sloppypar}
``Pilot'' has two distinct meanings in this study: the engineering pilot (${\sim}$5\%; ${\sim}$130 cases per configuration) was discarded before analysis and used only to validate the pipeline; none of the hypotheses (H1--H4) or analytic specifications were informed by the engineering pilot. The Phase~1 exploratory probes ($N = 50$ per condition; see Section~\ref{sec:exploratory-mechanisms} and Appendix~\ref{app:phase1-pilot}) are distinct from the engineering pilot. Because these probes informed the pre-registered Phase~2 hypotheses (DOI: \href{https://doi.org/10.17605/OSF.IO/WA9Y7}{10.17605/OSF.IO/WA9Y7}), they reduce the predictive novelty of the Phase~2 results. The protocol for Phase~2 was pre-registered separately, but all 600 items are in the primary dataset.
\end{sloppypar}

\end{document}